%% file: main.tex
\documentclass[10pt,twocolumn,letterpaper]{article}

\usepackage{cvpr}              % To produce the CAMERA-READY version
\input{preamble}

\definecolor{cvprblue}{rgb}{0.21,0.49,0.74}
\usepackage[pagebackref,breaklinks,colorlinks,allcolors=cvprblue]{hyperref}

\def\confName{CVPR}
\def\confYear{2027}

\title{LiteCalib: Fast, Model-Agnostic Single-View Camera Calibration}

\author{
Viktor Kocur$^1$ \quad Luk\'{a}\v{s} Buj\v{n}\'{a}k$^1$  \quad Luk\'{a}\v{s} Gajdo\v{s}ech$^{1}$  \quad Torsten Sattler$^2$  \quad Zuzana Kukelova$^3$ \\
$^1$ Faculty of Mathematics, Physics and Informatics, Comenius University in Bratislava\\
$^2$ Czech Institute of Informatics, Robotics and Cybernetics, Czech Technical University in Prague \\
$^3$ Visual Recognition Group, Faculty of Electrical Engineering, Czech Technical University in Prague \\
{\tt\small viktor.kocur@fmph.uniba.sk }
\and
}

\begin{document}
\maketitle
\begin{abstract}
Recent learning-based methods for single-image camera calibration rely on predicting dense per-pixel fields as an intermediate representation, which can be used to fit the parameters of a specific model. In this paper we show that predicting only a sparse intermediate representation is sufficient and makes it possible to design a significantly faster method while retaining accuracy. Beyond focusing on speed we scale up the training data by utilizing existing datasets as well as collecting a new CommonsPano dataset of 9 891 publicly available panoramas with permissive licenses. We perform an extensive evaluation in terms of direct camera calibration accuracy and performance in downstream 3D vision tasks such as relative pose estimation, SfM, and 3D reconstruction. The evaluation shows that our method is both significantly faster and more accurate compared to the previous state of the art. Code and dataset are available at: https://kocurvik.github.io/litecalib
\end{abstract}

\begin{figure}[t]
\centering
\input{fig/speed_acc}
\vspace{-6pt}
\caption{\textbf{Speed-Accuracy Tradeoff.}  Predicted ray angular error mAA($5^\circ$) averaged over ten benchmark sets against the total runtime (including model fit) per image on a Jetson Orin Nano 8\,GB
% at 15\,W with TensorRT fp16 engines (top) and on
and an A100 GPU.
AnyCalib$_\text{gen}$ is fitted with its released calibrator, AnyCalib$_\text{ours}$ with our implementation of it.
Our method outperforms previous works~\cite{tirado2025anycalib, veicht2024geocalib, piccinelli2025unik3d} in terms of accuracy and runtime.
In a fair comparison against AnyCalib~\cite{tirado2025anycalib} trained using the same data (AnyCalib$_{\text{ours}}$), our method achieves the same accuracy while being significantly faster.
}
\label{fig:speed-acc}
\end{figure}
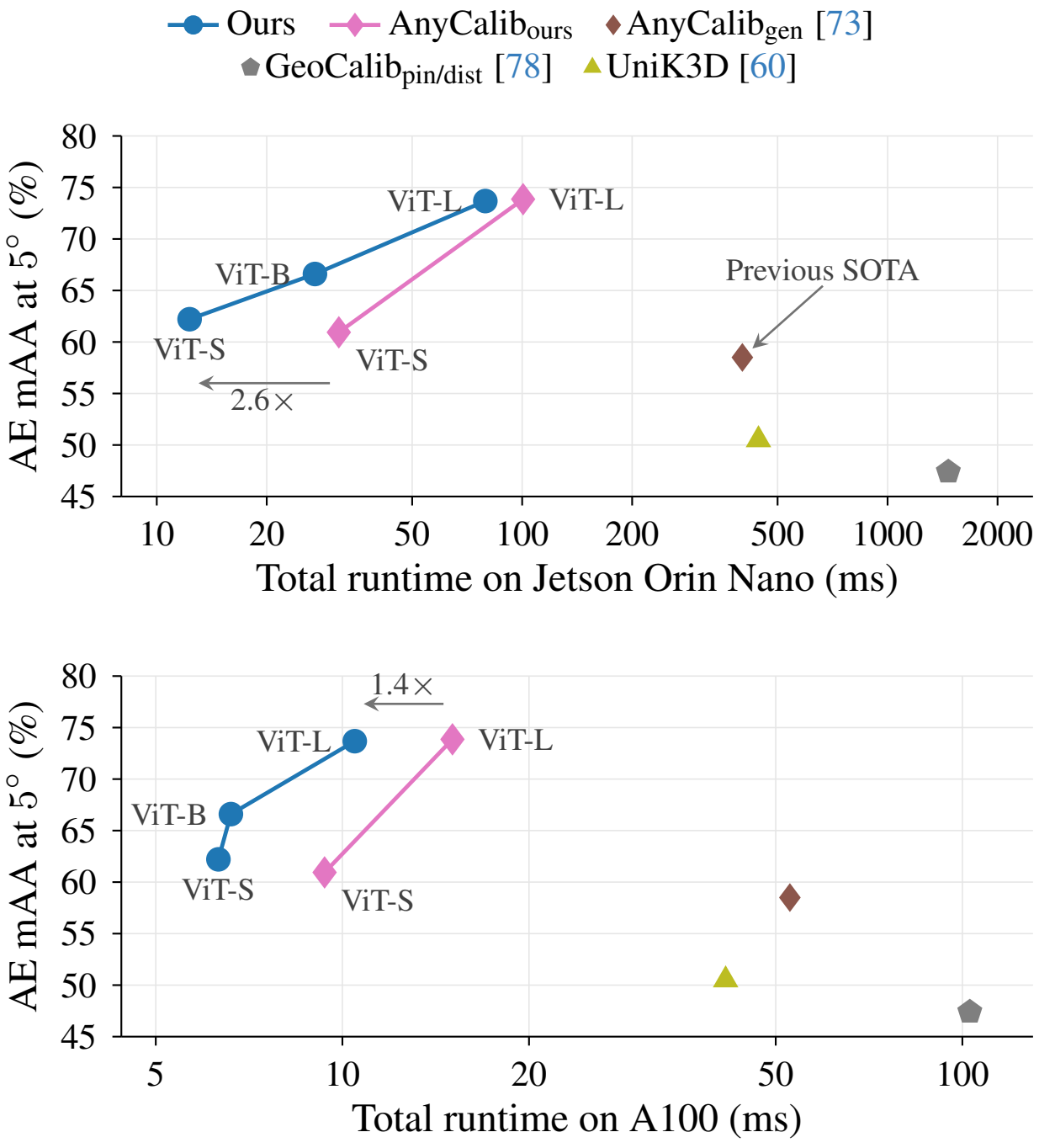

\section{Introduction}

For many 3D vision tasks using RGB images such as SfM~\cite{schoenberger2016sfm}, SLAM~\cite{mur2015orb}, object pose estimation~\cite{labbe2022megapose}, or metric measurements~\cite{sochor2017traffic}, it is important to have an accurate mathematical model of the imaging process, which describes how a camera projects a 3D scene onto the image pixels.
Camera calibration is the task of estimating such a model.
For most applications, it is sufficient to assume that the imaging process is described well by models which require only a few parameters (intrinsics) such as the pinhole camera model or models that also assume radial distortions~\cite{brown1971close,fitzgibbon2001division,kannala2006generic,khomutenko2016enhanced}.

In this paper, we address the problem of camera calibration from a single image of an arbitrary scene.
Camera calibration from a single image is a classical and well-studied problem. While classical approaches typically require some knowledge of the scene, such as a calibration pattern~\cite{zhang2000flexible} or vanishing points~\cite{dubska2013}, more recent learning-based approaches~\cite{lopez2019deep,dal2024revisiting,veicht2024geocalib,tirado2025anycalib} aim to solve this problem for arbitrary scenes.
Similar to previous works~\cite{zhu2023wildcam,jin2023perspective,veicht2024geocalib}, the current state-of-the-art method AnyCalib~\cite{tirado2025anycalib} does not regress the camera parameters directly, but instead predicts an intermediate per-pixel representation in the form of a FoV field.
The FoV field provides per-pixel constraints which can be used to fit the intrinsics of different camera models.
On several benchmarks, AnyCalib has demonstrated state-of-the-art results while being model-agnostic.
Unfortunately, dense fields come at the cost of a large optimization problems required to fit the camera model. Moreover, the inference of the dense field adds cost to the network. 
We show that the dense output ($\sim10^5$ rays) can be replaced by a much sparser representation ($25$ rays), retaining the same accuracy and the ability to fit different camera models while significantly reducing the computational cost of both inference and camera model fit.
Our experiments show that the specific structure of the network output is not an essential factor in terms of accuracy as long as it provides sufficient constraints for all considered camera models.
However, we show that during training auxiliary dense supervision is beneficial.

Based on this insight, we design a neural network that outputs a sparse representation that can still accommodate multiple camera models.
Our design is based on existing components and is intentionally simple.
To improve accuracy, we expand the training dataset by using OpenPano V2~\cite{tirado2025anycalib}, MegaScenes~\cite{tung2024megascenes}, and curating our own dataset \commonspano{} containing 9 891 panoramas licensed under Creative Commons licenses from Wikimedia Commons.
To fully use the benefits of the sparse network output, we implement an efficient parameter fit in Numba~\cite{lam2015numba} on both CPU and GPU.
Compared to the released AnyCalib models, our network achieves significantly better calibration accuracy on all evaluated metrics.
In a more fair comparison, against AnyCalib's network trained on the same dataset and using our efficient Numba implementation for fitting, our network achieves the same accuracy while being computationally more efficient (see Fig.~\ref{fig:speed-acc}). 
Beyond direct evaluation of the camera calibration accuracy, we also show the practical benefits of our network for various downstream tasks, including relative pose estimation~\cite{kocur2026minimal}, SfM~\cite{schoenberger2016sfm,pan2024glomap}, and feed-forward 3D reconstruction~\cite{wang2025pi3,keetha2026mapanything}.

% To summarize, in this paper we present the following contributions:
In summary, this work makes the following contributions: 
% \begin{itemize}
    % \item 
    (1) An experimental 
    evaluation showing that, during inference, the dense fields are not necessary to retain high accuracy.
    % with dense output only being used to provide a dense supervision signal during training. 
    % \item 
    (2) A computationally efficient neural network that outputs a sparse, model-agnostic camera representation and an efficient implementation of the final camera model fit.
    % \item 
    (3) A newly collected \commonspano{} dataset of 9 891 publicly available panoramas along with a training data mix enabling training at scale.
    % \item 
    (4) An extensive evaluation of single image camera calibration methods via direct evaluation, as well as evaluation on downstream 3D vision tasks: relative pose estimation, SfM, and, feed-forward 3D reconstruction.
    % , and, in the SM, localization. 
    (5) \commonspano{} dataset, code and the trained models are made publicly available.\footnote{\url{https://kocurvik.github.io/litecalib}}
% \end{itemize}

% We will make the code, the \commonspano{} dataset, and the trained models publicly available.

\section{Related Work}

\subsection{Camera Models}

The pinhole camera model is the most commonly used for images without distortion.
The model has several parameters: focal length, aspect ratio, principal point and skew.
Often, simplifying assumptions can be made (aspect ratio $\approx$ 1, principal point in the image center, no skew) that reduce the calibration problem to estimating only the unknown focal length.
The pinhole camera model can be supplemented by modeling polynomial radial distortion with the Brown-Conrady model~\cite{brown1971close}, which can represent limited small distortions well~\cite{leotta2022maximum}. For wide-FoV and fisheye cameras distortion can be modeled with the simple division~\cite{fitzgibbon2001division}, Kannala-Brandt~\cite{kannala2006generic}, UCM~\cite{geyer2000unifying,mei2007single}, EUCM~\cite{khomutenko2016enhanced} or double sphere~\cite{usenko2018double} models.
Which model suits a camera depends on its lens, and the parameters of one model may not transfer to another~\cite{usenko2018double}.
A calibration network that predicts the parameters of one model is thus limited to the cameras it represents well. This motivates model-agnostic intermediate representations, which can model the imaging process without an explicit model~\cite{camposeco2015non,schops2020having}.

The standard method~\cite{zhang2000flexible} for obtaining camera parameters for a given camera requires taking several images of a calibration pattern. 
However, in some situations, such as working with crowd-sourced images or zooming cameras it may be impossible to obtain such images and thus the calibration parameters must be obtained by other means.

\subsection{Monocular Camera Calibration}

\paragraph{Classical Methods} The problem of monocular camera calibration has long been studied~\cite{brown1971close}. Classical approaches may rely on a camera observing a known 3D object where 2D-3D correspondences can be established. 
The intrinsic camera parameters can then be estimated jointly with the camera pose~\cite{tsai1987versatile,kukelova2013real,larsson2017making,bartl2021landmark}. Camera self-calibration can also exploit vanishing points which provide geometric constraints on the intrinsics and the camera orientation~\cite{caprile1990using}. Vanishing points can be estimated using RANSAC \cite{bazin20123}, optimization \cite{bazin2012globally}, search \cite{qian2022reliable}, or the cascaded Hough transform \cite{dubska2013}. Curved lines~\cite{lochman2021minimal} or repeating patterns~\cite{pritts2018radially,pritts2020minimal} can also be used to estimate radial distortion. These methods require sufficient geometric cues in the image, and some also need three dominant orthogonal directions~\cite{dubska2013}, which largely restricts them to man-made environments.

\paragraph{Learning-based methods} 
Early learning-based methods regress the camera parameters directly, such as the focal length or field of view and one distortion coefficient from a single image~\cite{workman2015deepfocal,hold2018perceptual,bogdan2018deepcalib,lopez2019deep} and are tied to the camera model they were trained for.
Alternatively, networks can predict an intermediate structure such as vanishing points~\cite{chang2018deepvp,kocur2021traffic,wakai2024deep}. 
Recent methods estimate dense representations such as incidence rays~\cite{zhu2023wildcam,he2025diffcalib}, perspective fields~\cite{jin2023perspective,veicht2024geocalib,li2026calibanyview} or FoV fields used by AnyCalib~\cite{tirado2025anycalib}.
After inference, the camera parameters of multiple different models can be fitted to the resulting dense fields.
In this paper, we argue that estimating the full dense fields is unnecessary to achieve good accuracy. Instead, we estimate only a sparse representation which is sufficient to model a diverse set of camera models while being computationally efficient. A similar approach regressing sparse rays has been used for camera pose estimation~\cite{zhang2024cameras}.

A previous line of work~\cite{camposeco2015non,liao2021deep,dal2024revisiting} utilizes sparse representations for camera calibration. Work closely related to ours is VaCR~\cite{dal2024revisiting}. VaCR estimates polar angles of rays at 25 radii forming a 1D profile from which a camera model can be fit. The profile assumes a centered principal point and square pixels by construction. VaCR targets wide-angle lenses (80–190° FoV), with synthetic fisheye training data and a fit initialized by differential evolution. Our grid of 2D rays keeps no such assumption in the representation and covers a wider range of camera models.

\paragraph{3D Foundation Models} Monocular depth and 3D reconstruction models often estimate the camera pinhole intrinsics alongside the geometry~\cite{piccinelli2024unidepth,piccinelli2025unidepthv2,wang2025moge,wang2025moge2,kong2026moge3,knaebel2026surge}, and so do recent multi-view feed-forward reconstruction models~\cite{lin2025da3,keetha2026mapanything,wang2026vggtomega}. Despite being designed for multi-view tasks, they also perform well when only a single image is provided at inference. 
Among monocular depth estimators, UniK3D~\cite{piccinelli2025unik3d} and OmniPoint~\cite{ye2026omnipoint} address camera models beyond the standard pinhole model by predicting a ray map instead of the parameters of a specific model, which enables the network to generate 3D point clouds even for distorted images. 
Fisheye3R~\cite{duan2026fisheye3r} extends multi-view models with a small set of learned tokens to deal with fisheye cameras.
We evaluate our method against UniK3D and other 3D foundation models with public implementations and model weights.
The recent developments in 3D foundation models show the importance of training data scaling for 3D vision tasks.
We show that this holds also for the case of camera calibration.

\section{Do Calibration Fields Need to Be Dense?}
\label{sec:sparse}
Recent neural networks for monocular camera calibration \cite{veicht2024geocalib,tirado2025anycalib} estimate dense fields which provide constraints on the  parameters of a given camera model.
Estimating the parameters of a specific camera model can then be formulated as an optimization task.
Due to the dense per-pixel nature of the output fields, the optimization tasks are computationally expensive.
In this section, we show on AnyCalib~\cite{tirado2025anycalib}, whose field serves any of six camera models, that using the dense field is not necessary and relying only on a sparse subset of points in the field retains accuracy while significantly decreasing the computational costs.
We also reimplement the fit efficiently. In the next section, we leverage this observation to design a more efficient neural network.

\subsection{AnyCalib Camera Calibration}
\label{sec:fov_field}

For each pixel in the input image $\vec{x} = (u, v)$ AnyCalib predicts the unit ray $\vec{p} = (X, Y, Z) \in \mathcal{S}^2$. To calibrate a camera from the field of rays, AnyCalib considers a forward projection function $\pi: \mathcal{S}^2 \mapsto \mathbb{R}^2$ of the form:
\begin{equation}
\pi(\vec{p}) 
= \begin{pmatrix} u \\ v \end{pmatrix}
= f \phi(R, Z) \begin{pmatrix} X \\ aY \end{pmatrix} + \begin{pmatrix} c_x \\ c_y \end{pmatrix} \enspace,   
\label{eq:ac_proj}
\end{equation}
where $\phi(R, Z)$ is a camera-model specific function of the ray radius $R = \sqrt{X^2 + Y^2}$ and its $Z$ component. 
We denote by $a$ the pixel aspect ratio $f_y/f_x$ and $\vec{c} = (c_x, c_y)$ as the principal point. 
The intrinsics $\boldsymbol{\eta}$ of a chosen camera model (pinhole, Brown-Conrady~\cite{brown1971close}, Kannala-Brandt~\cite{kannala2006generic}, UCM~\cite{mei2007single},
EUCM~\cite{khomutenko2016enhanced}, or division~\cite{fitzgibbon2001division}) are fitted to pixel-ray pairs $(\vec{x}, \vec{p})$ in three stages.
By calculating the ratio of $u$ and $v$ from \eqref{eq:ac_proj}, $f$ and $\phi$ can be eliminated yielding:
\begin{equation}
a\,Y (u - c_x) = X (v - c_y) \enspace,
\label{eq:anycalib_c}
\end{equation}
which is linear in $a$, $a c_x$ and $c_y$. 
Thus, the principal point $\vec{c}$ and the pixel aspect ratio $a$ can be obtained using linear least squares over all pixels.

With $\vec{c}$ and $a$ known, the remaining intrinsics enter the projection linearly for a wide range of camera models (see Tab. 1 in \cite{tirado2025anycalib}) and can also be solved by least squares over all pixels. 
Since the solutions minimize only algebraic errors, a final refinement with 10 iterations of Gauss-Newton is performed to minimize the angle between the predicted rays and the rays of the
model:
\begin{equation}
E_\text{AC}(\boldsymbol{\eta}) = \sum_{\vec{x}} \arccos^2\!\left( \vec{p}^\top \pi^{-1}(\vec{x}; \boldsymbol{\eta}) \right),
\label{eq:anycalib}
\end{equation}
where $\pi^{-1}(\vec{x}; \boldsymbol{\eta}) \in \mathcal{S}^2$ is the ray through $\vec{x}$ given by the model with intrinsics $\boldsymbol{\eta}$.
% Each pixel contributes a 2D residual, the difference of the two rays expressed in the tangent plane of the model's ray, whose norm is the angle in \eqref{eq:anycalib}.

\paragraph{Intermediate Representation}
AnyCalib's network does not predict $\vec{p}$ directly. Instead it uses an intermediate representation $\vec\theta$:
\begin{equation}
    \vec{\theta} = \begin{pmatrix} 1 & 0 & 0 \\ 0 & 1 & 0 \end{pmatrix} \arccos\left(\vec{z}_1^\top \vec{p} \right)
    \frac{\vec{p}- (\vec{z}_1^\top \vec{p})      
    \vec{z}_1}{|| \vec{p}- (\vec{z}_1^\top \vec{p}) \vec{z}_1 ||},
\end{equation}
where $\vec{z}_1 = (0, 0, 1)$. The inverse mapping is:
\begin{equation}
\vec{p} = \begin{bmatrix} \frac{\sin(\theta)}{\theta} \vec{\theta} \\ \cos(\theta)\end{bmatrix},
\end{equation}
where $\theta= || \vec{\theta}||$ is the polar angle of the ray $\vec{p}$. A dense field containing $\vec{\theta}$ for each pixel is predicted by a network with a DINOv2~\cite{oquab2023dinov2} ViT-L backbone~\cite{dosovitskiy2021vit} and a light DPT decoder~\cite{ranftl2021dpt}. For more details refer to~\cite{tirado2025anycalib}.

\subsection{Efficient Sparse Optimization}
\label{sec:layouts}

\begin{table}[t]
\centering
\setlength{\tabcolsep}{3pt}
\renewcommand{\arraystretch}{1.1}
\resizebox{\linewidth}{!}{%
\begin{tabular}{llrc cc cc}
\toprule
& & & & \multicolumn{2}{c}{Mean error $\downarrow$} & \multicolumn{2}{c}{ms $\downarrow$} \\
\cmidrule(lr){5-6}\cmidrule(lr){7-8}
Fit & Points & \#pts & $\mathbf{c}$, $a$ & Pinhole & Distorted & CPU & GPU \\
\midrule
% rows sparse: begin
\multirow{2}{*}{AnyCalib~\cite{tirado2025anycalib}} & Dense & $\sim\!10^5$ & free & 2.12$^\circ$ & 2.37$^\circ$ & -- & 35.6 \\
 & Dense & $\sim\!10^5$ & fixed & $-0.3\%$ & $-0.3\%$ & -- & 34.7 \\
\midrule
\multirow{6}{*}{Ours} 
% & Dense & $\sim\!10^5$ & free & $0.0\%$ & $0.0\%$ & 33.9 & 1.73 \\
%  & $99 \times 99$ & 9801 & free & $0.0\%$ & $-0.1\%$ & 8.60 & 0.97 \\
%  & $25 \times 25$ & 625 & free & $-0.1\%$ & $-0.2\%$ & 0.79 & 0.75 \\
%  & $7 \times 7$ & 49 & free & $-0.3\%$ & $-0.8\%$ & 0.17 & 0.68 \\
%  & $5 \times 5$ & 25 & free & $-0.3\%$ & $-1.1\%$ & 0.15 & 0.77 \\
%  & $3 \times 3$ & 9 & free & $+0.1\%$ & $+510\%$ & 0.13 & 0.71 \\
% \cmidrule(lr){2-8}
 & Dense & $\sim\!10^5$ & fixed & $-0.3\%$ & $-0.4\%$ & 15.1 & 1.53 \\
 & $99 \times 99$ & 9801 & fixed & $-0.3\%$ & $-0.4\%$ & 4.81 & 1.10 \\
 & $25 \times 25$ & 625 & fixed & $-0.4\%$ & $-0.5\%$ & 0.78 & 0.76 \\
 & $7 \times 7$ & 49 & fixed & $-0.7\%$ & $-1.0\%$ & 0.17 & 0.69 \\
 & $5 \times 5$ & 25 & fixed & $-0.8\%$ & $-1.5\%$ & 0.14 & 0.70 \\
 & $3 \times 3$ & 9 & fixed & $-0.9\%$ & $+553\%$ & 0.13 & 0.73 \\
% rows sparse: end
\bottomrule
\end{tabular}}
\caption{\textbf{Sparse fits of AnyCalib's field.} AnyCalib's calibrator in its released implementation
(ten Gauss-Newton iterations) and our Numba~\cite{lam2015numba} implementation (two iterations), on all pixels and on $N \times N$ grids. First row shows the mean angular error averaged over the benchmarks. The remaining rows show the change relative to the first row. $\mathbf{c}$, $a$: principal point and aspect ratio estimated (free) or fixed to the image center and $a = 1$. ms: fit time per image.}
\label{tab:sparse}
\end{table}

To show that AnyCalib's field is redundant and the optimization problem can be simplified, we consider using only a sparse subset of points. 
More specifically, we consider only the rays at an $N \times N$ grid of pixels evenly spaced from border to border of the image, its corners included. 
We consider $N \in \{3, 5, 7, 25, 99\}$, \ie, from 9 to 9801 of the $\sim\!10^5$ pixels of the full image resolution after preprocessing.

Additionally, we consider that for standard consumer cameras, the aspect ratio $a$ is typically very close to 1, and the principal point $\vec{c}$ is close to the image center. Previous work has shown that these assumptions are sufficient for many applications \cite{sochor2017traffic,larsson2017making,camposeco2015non}. Moreover, outside of synthetically generated cropped and squashed images used to train a separate model both AnyCalib and GeoCalib use training data where this assumption holds. In the main paper we therefore operate with this assumption when fitting camera models. In SM we show that our sparse representation can also be used to fit both $\vec{c}$ and $a$ and achieve the same accuracy as AnyCalib.

The parameter fit \eqref{eq:anycalib} provided by the official PyTorch implementation does not scale well with the number of residuals. 
Due to the overhead of some operations, the parameter fit for a sparse subset of points results in a total runtime similar to that of using the full dense field.
Moreover, the parameter fit is computationally inefficient, taking almost three times as long as the network inference on an A100 GPU (35.6\,ms vs.\ 12.6\,ms). 
We reimplement the camera parameter fit in Numba for both CPU and GPU. We also find that, due to the initial linear fit, using just two iterations of Gauss-Newton optimization is sufficient without accuracy degradation. 

\subsection{Results}

\label{sec:sparse_results}

We evaluate the effect of using a sparse grid on the 7 test sets used in the AnyCalib paper. We follow the evaluation protocol from~\cite{tirado2025anycalib}. For each set the protocol fits the same camera model the set uses as ground truth. Four sets use the standard pinhole camera model, one set uses the one-parameter simple radial model~\cite{brown1971close}, one set uses the UCM model~\cite{mei2007single} and the remaining set uses the four-parameter Kannala-Brandt model~\cite{kannala2006generic}. Table~\ref{tab:sparse} shows the mean angular error averaged across the 7 test sets for different sparse point configurations (see Sec.~\ref{sec:layouts}). 
Our implementation shows a significant improvement over the official one.

Using the full dense field, our implementation of the camera model fit on an A100 GPU is over 20 times faster than the official implementation.
Due to overheads and high parallelization, the GPU implementation has only limited gains when sparser layouts are used. 
However, with the CPU implementation the gain is significant up to 0.14\,ms for the $5 \times 5$ layout.
% until reaching its minimum with a $5 \times 5$ layout, taking only 0.14 ms. 
Compared to the official implementation this is an  improvement of over $200\times$.
A significant portion of this gap is due to an inefficient implementation. When AnyCalib's dense field is used with our efficient Numba implementation on GPU the gap shrinks leaving our sparse fit $10 \times$ faster.   

In terms of accuracy fixing $\vec{c}$ in center and $a = 1$ leads to slight improvement. At the same time, we note that the evaluated metrics keep improving with fewer points used.
This may be caused by implicitly giving a larger proportion of the overall residual weight to the points in the outermost regions of the image where distortion is the most apparent.
We do not consider these improvements significant, but they provide a strong indication that subsampling the dense field and fixing $\vec{c}$ and $a$ does not lead to degradation of accuracy. 
The $+553\%$ of the $3 \times 3$ grid on the distorted benchmarks is caused by a single benchmark, ScanNet++. Its fisheye images are fitted with the four-parameter Kannala-Brandt model~\cite{kannala2006generic}, which has more degrees of freedom than independent constraints provided by the $3 \times 3$ grid and the fit thus fails.
More details including per-dataset comparisons and evaluation using 10 GN iterations and fitting with $\vec{c}$ and $a$ free are provided in SM.

\section{Method}
\label{sec:method}

\begin{figure*}[t]
\centering
\includegraphics[width=\linewidth]{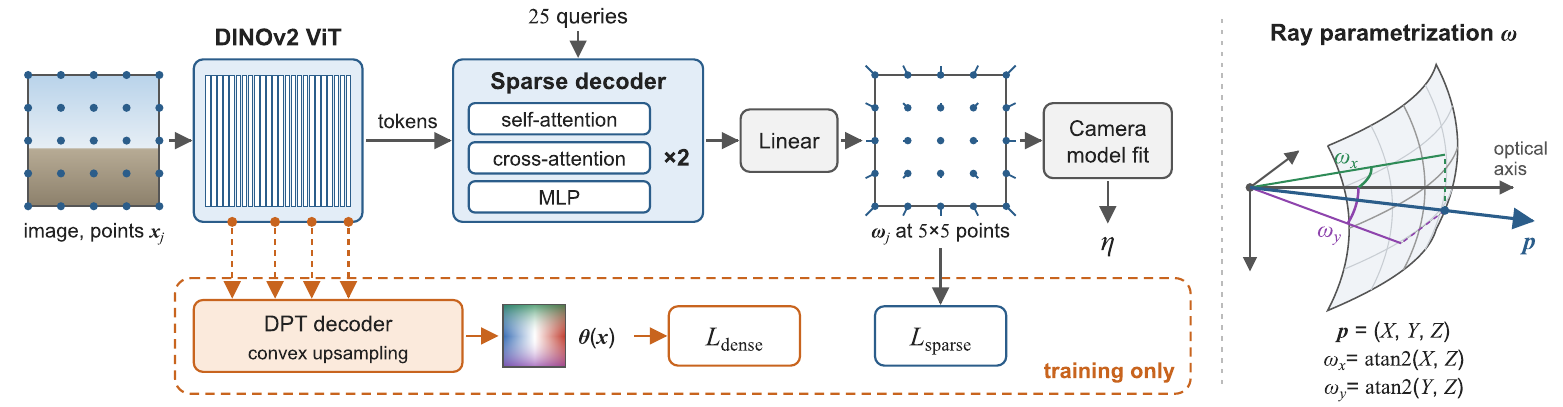}
\caption{\textbf{Network.} Top: the network at inference, predicting the signed rays $\vec\omega_j$ at the $5 \times 5$ points $\vec{x}_j$, from which we fit the intrinsics $\boldsymbol{\eta}$ (Sec.~\ref{sec:sparse}). Bottom: AnyCalib's dense decoder, used only during training. Right: the ray $\vec{p}$ is represented by its two angles from the optical axis, $\omega_x$ in the $XZ$ plane and $\omega_y$ in the $YZ$ plane~\eqref{eq:omega}. $\vec{x}$ is the point where $\vec{p}$ intersects the projection surface (plane for pinhole cameras). }
\label{fig:architecture}
\end{figure*}

In the previous section, we have shown that the full dense FoV field is redundant for fitting  camera parameters. In this section, we propose a neural network design that uses this insight by predicting the FoV field only for a grid of sparse points (see Fig.~\ref{fig:architecture}).

\subsection{Output Representation}

Instead of AnyCalib's dense per-pixel field we output a $5 \times 5$ grid of rays $\vec{p}$. We chose the $5 \times 5$ layout as it provides the best speed without degrading accuracy (Sec.~\ref{sec:sparse_results}). Each ray $\vec{p}_j$ is represented by two signed angles $\vec\omega_j$, each measured from the optical axis in the plane of one image axis:
\begin{equation}
\vec\omega = \begin{pmatrix} \omega_x \\ \omega_y \end{pmatrix} = \begin{pmatrix} \operatorname{atan2}(X, Z) \\ \operatorname{atan2}(Y, Z) \end{pmatrix} \enspace ,
\label{eq:omega}
\end{equation}
with the inverse mapping:
\begin{equation}
\vec{p} \propto \begin{pmatrix} \sin\omega_x \cos\omega_y & \cos\omega_x \sin\omega_y & \cos\omega_x \cos\omega_y \end{pmatrix}^\top.
\label{eq:omega_inv}
\end{equation}
% In 
Sec.~\ref{sec:ablations} later shows %we show
that the choice of representation is not crucial, and we choose $\vec\omega$ due to its simplicity (in contrast to $\vec\theta$ (see Sec.~\ref{sec:fov_field}), which relies on differential geometry). The rays $\vec{p}_j$ are obtained for each point on the $5 \times 5$ grid by using \eqref{eq:omega_inv}. We then fit the desired camera model using our implementation of AnyCalib's fit (Sec.~\ref{sec:layouts}).
% Unlike AnyCalib, we fix the principal point $\vec{c}$ at the image center and the aspect ratio to $a = 1$ in this fit. On unedited image benchmarks this change does not lead to decreased accuracy as the network is trained with images where these assumption holds.
% It is still possible to train our network with $\vec{c}$ and $a$ free during both training and when fitting a model (see SM for experiments using cropped and squashed images). 

\subsection{Sparse Prediction Head}
\label{sec:head}

To predict the $5 \times 5$ grid containing $\vec{\omega}_j$ we use a DINOv2~\cite{oquab2023dinov2} pre-trained vision transformer~\cite{dosovitskiy2021vit} as the backbone.
The image is resized to $H \times W \approx 10^5$ pixels with $H$ and $W$ divisible by the patch size 14.
The backbone outputs $N = HW / 14^2$ patch tokens $\vec{t}_n \in \mathbb{R}^D$ and a class token $\vec{t}_\text{cls}$.
Instead of decoding the tokens into a dense field, we predict the field only at the $5 \times 5$ grid of points $\vec{x}_j$ evenly spaced from border to border of the image (see Fig.~\ref{fig:architecture}).

Each grid point has a learned query and a positional embedding.
The $25$ queries pass through two transformer decoder layers~\cite{vaswani2017attention} that cross-attend to the image patch tokens and the special CLS token.
To obtain the output $\vec\omega_j$ from the embeddings we use a single linear layer. Since the decoder processes only $25$ queries, it adds 2.8 GMACs to the 174.2 GMACs of the ViT-L backbone at $322 \times 322$ pixels, although it has 33.8\,M parameters.
AnyCalib's DPT decoder and head have only 15.9\,M parameters, but they operate on feature maps at $1/7$ of the input resolution and thus result in 30.5 GMACs at the same input.

\subsection{Training Loss}
\label{sec:aux}

The sparse head is trained with the L1 loss on the predicted field $\vec{\omega}$ against the ground truth values $\hat{\vec{\omega}}$ computed from the GT camera parameters:
\begin{equation}
\mathcal{L}_\text{sparse} = \frac{1}{25} \sum_{j=1}^{25} \big\| \vec\omega_j - \hat{\vec\omega}_j \big\|_1.
\label{eq:loss_sparse}
\end{equation}
Images whose ground-truth camera does not define a ray at every grid point are excluded.
This loss supervises the backbone through only $25$ points per image and we found that models trained using only sparse supervision do not achieve the accuracy of AnyCalib.

To aid the supervision, we thus add AnyCalib's dense decoder to the backbone during training as an auxiliary branch with its own dense loss.
Its light DPT decoder~\cite{ranftl2021dpt} fuses the tokens of four backbone blocks into a feature map at $1/7$ of the input resolution, and its head predicts $\vec\theta$ at every input pixel by convex upsampling~\cite{teed2020raft}.
The branch is trained with AnyCalib's loss ((\ref{eq:loss_sparse}) with $\vec\theta$ instead of $\vec\omega$, summed over every pixel).
The network is trained with $\mathcal{L} = \mathcal{L}_\text{sparse} + \lambda \mathcal{L}_\text{dense}$, $\lambda = 1$.
We found that varying $\lambda$ does not yield significantly different results.
The two branches share only the backbone (see SM).
After training, we discard the dense branch so it does not contribute to the computational cost.

\subsection{Training}
\label{sec:training}

We train on the data described in Sec.~\ref{sec:train_data} 
% from pre-trained DINOv2 backbone 
using AdamW~\cite{loshchilov2017decoupled} with gradient clipping at 1.0 and a batch size of 24.
As in AnyCalib~\cite{tirado2025anycalib}, the input has $320^2$ pixels with aspect ratios sampled from $[0.5, 2]$, and we use the same augmentations.
For ViT-L we also keep AnyCalib's learning rates, $6 \times 10^{-5}$ for the decoders and heads and $6 \times 10^{-6}$ for the backbone, with a weight decay of 0.01.
For the smaller backbones we use $10^{-4}$ and $10^{-5}$ with a weight decay of 0.05.

Instead of AnyCalib's step decay, we use a warmup-stable-decay schedule~\cite{hagele2024scaling}.
We train for 24 epochs of 6000 steps, 144\,000 steps in total.
The learning rate rises linearly to its peak over the first 1000 steps, stays there until the last 5 epochs, and then decays to zero as $1 - \sqrt{t}$, where $t \in [0, 1]$ is the progress through the decay phase. The length of the stable training phase was determined by observing the validation performance.

\section{Datasets}
\label{sec:datasets}

\subsection{Evaluation Datasets}
\label{sec:eval_data}

We evaluate on ten benchmarks. Seven of them are the test sets of AnyCalib~\cite{tirado2025anycalib}: Stanford2D3D~\cite{armeni2017joint}, TartanAir~\cite{wang2020tartanair}, LaMAR~\cite{sarlin2022lamar}, ScanNet++~\cite{yeshwanth2023scannetpp}, Mono~\cite{engel2016photometrically}, and MegaDepth~\cite{li2018megadepth} in the original and undistorted versions.

With the exception of MegaDepth, most of the benchmark sets include only very few distinct cameras. We therefore further extend the evaluation with three additional benchmarks: InFlux~\cite{liang2025influx}, InFlux++ Real~\cite{liang2026influxpp} and Cathedral~\cite{kocur2026minimal}.
Both InFlux datasets consist of real videos in which focal length and focus change over time and per-frame ground truth is provided. We sample 1860 and 1484 frames from the two datasets. Cathedral is a dataset with ground truth derived from SfM. We use 1253 of its images covering 42 cameras with significant distortion.
More details are in SM.

\subsection{Training Datasets}
\label{sec:train_data}

We train on synthetic views rendered from real panoramas and on real photos with ground truth derived from SfM. We render the views from two sets of panoramas, in both cases following the recipe of AnyCalib's OP$_\text{g}$~\cite{tirado2025anycalib}: 16 views per panorama at $640 \times 640$ pixels with cameras sampled with 34\% pinhole, 33\% Brown-Conrady, and 33\% EUCM camera models with varying distortion. 
The first set is the 3 651 training panoramas from OpenPano V2~\cite{tirado2025anycalib} giving a set of 51\,719 views. It is an extended version of OpenPano V1~\cite{veicht2024geocalib} and is composed of panoramas from multiple sources including the Laval Indoor Panoramic HDR Dataset~\cite{bolduc2023beyond,gardner2017learning}.
The second is \commonspano{}, a set of 9 891 public domain, CC0, CC BY, and CC BY-SA panoramas that we collected from Wikimedia Commons, whose training split gives 148\,626 views from 9336 panoramas. We exclude panoramas where metadata suggest that they may have been taken in close proximity from any of the benchmark scene locations.
We will make the \commonspano{} dataset publicly available for download under the CC BY-SA license.

The real photos come from MegaScenes~\cite{tung2024megascenes}, whose COLMAP reconstructions of Wikimedia Commons photo collections provide a one-parameter radial camera with the principal point at the image center for each photo.
We keep the photos in their distorted form, center-cropped to keep the principal point centered, and exclude the scenes that show the landmarks of the MegaDepth benchmark~\cite{li2018megadepth}, which leaves about 1.21 million views from 14\,382 scenes.

In each epoch, we draw 72\,000 views from the panorama views, and 72\,000 views from MegaScenes, so that the real photos make up half of the training data.
Each source has its own validation split, which we use to monitor the training.

\section{Evaluation}
\label{sec:evaluation}

\begin{table*}[t]
\caption{\textbf{Camera Calibration Accuracy.} Mean and standard deviation over seeds of the median error and mAA, each averaged over the ten benchmarks. Runtimes are from an NVIDIA A100 running inference in half precision, one image at a time. Our fit is timed on the same machine using 16 CPUs. Bold: best mean. $^\dagger$ UniK3D results may be overly optimistic since it was trained on parts of ScanNet++ and MegaDepth which are also included in the test sets.}
\centering
\resizebox{\textwidth}{!}{%
\begin{tabular}{l l cc cc cc cc ccc}
\toprule
 & & \multicolumn{2}{c}{AE ($^\circ$)} & \multicolumn{2}{c}{RE (px)} & \multicolumn{2}{c}{vFoV ($^\circ$)} & \multicolumn{2}{c}{hFoV ($^\circ$)} & \multicolumn{3}{c}{Time (ms)} \\
\cmidrule(lr){3-4} \cmidrule(lr){5-6} \cmidrule(lr){7-8} \cmidrule(lr){9-10} \cmidrule(lr){11-13}
Network & Fit & Median $\downarrow$ & mAA ($5^\circ$) $\uparrow$ & Median $\downarrow$ & mAA (50\,px) $\uparrow$ & Median $\downarrow$ & mAA ($5^\circ$) $\uparrow$ & Median $\downarrow$ & mAA ($5^\circ$) $\uparrow$ & Net $\downarrow$ & Fit $\downarrow$ & Total $\downarrow$ \\
\midrule
GeoCalib$_\text{pin/dist}$ & official & 3.028 & 47.4 & 78.0 & 31.6 & 6.54 & 28.5 & 8.34 & 26.5 & 23.5 & 79.4 & 102.8 \\
UniK3D$^{\dagger}$ & ours, GPU & 3.018 & 50.5 & 80.6 & 28.2 & 6.46 & 32.4 & 8.70 & 26.9 & 39.2 & 1.7 & 40.9 \\
AnyCalib$_\text{pin/dist}$ & official & 1.930 & 59.5 & 54.2 & 39.4 & 4.23 & 34.4 & 5.08 & 33.2 & 12.6 & 38.1 & 50.7 \\
AnyCalib$_\text{gen}$ & official & 1.975 & 58.5 & 55.2 & 37.7 & 4.32 & 33.2 & 5.31 & 30.4 & 12.7 & 39.9 & 52.6 \\
AnyCalib$_\text{ours}$ & official & 1.175 $\pm$ 0.024 & \textbf{73.9} $\pm$ 0.4 & \textbf{32.8} $\pm$ 1.0 & \textbf{49.7} $\pm$ 0.4 & 2.49 $\pm$ 0.05 & 51.4 $\pm$ 0.6 & \textbf{3.00} $\pm$ 0.07 & \textbf{47.6} $\pm$ 0.6 & 12.6 & 39.4 & 52.1 \\
AnyCalib$_\text{ours}$ & ours, GPU & 1.175 $\pm$ 0.024 & \textbf{73.9} $\pm$ 0.4 & \textbf{32.8} $\pm$ 1.0 & \textbf{49.7} $\pm$ 0.4 & 2.49 $\pm$ 0.05 & 51.4 $\pm$ 0.6 & \textbf{3.00} $\pm$ 0.07 & \textbf{47.6} $\pm$ 0.6 & 12.7 & 1.6 & 14.4 \\
% Ours, $5 \times 5$ $\vec\theta$ & ours, CPU & 1.174 $\pm$ 0.014 & 73.5 $\pm$ 0.4 & 33.3 $\pm$ 0.6 & 49.7 $\pm$ 0.3 & 2.46 $\pm$ 0.03 & 51.6 $\pm$ 0.4 & 3.06 $\pm$ 0.06 & 46.9 $\pm$ 0.4 & \textbf{9.5} & \textbf{0.3} & \textbf{9.8} \\
Ours & ours, CPU & \textbf{1.165} $\pm$ 0.016 & 73.7 $\pm$ 0.3 & 33.1 $\pm$ 0.4 & \textbf{49.7} $\pm$ 0.2 & \textbf{2.44} $\pm$ 0.04 & \textbf{51.9} $\pm$ 0.3 & 3.01 $\pm$ 0.04 & 47.5 $\pm$ 0.3 & \textbf{9.4} & \textbf{0.3} & \textbf{9.7} \\
\bottomrule
\end{tabular}}
\label{tab:results}
\end{table*}

\subsection{Metrics}
\label{sec:metrics}

The ten evaluation sets use different ground truth camera models. To evaluate the accuracy of the estimated camera parameters we use model-agnostic metrics, which allows us to compare methods regardless of the camera models they fit. We use the metrics from AnyCalib~\cite{tirado2025anycalib}. The angular error (AE) of an image is the angle between the rays obtained from the ground-truth and the estimated camera model, averaged over a uniform grid of pixels.
The reprojection error (RE) is the corresponding average distance in pixels.
The vFoV and hFoV errors are the absolute differences between the ground-truth and the estimated vertical and horizontal fields of view. 
For each error we report the median over the images of a benchmark and the mean average accuracy (mAA), the area under its recall curve up to $5^\circ$, or 50 pixels for RE using AnyCalib's implementation.

All tables average these values over the ten benchmarks of Sec.~\ref{sec:eval_data}. For each evaluated variant, we train the network five times with different seeds to evaluate the statistical significance of our results.   
For networks trained by us, we report both the mean and standard deviations.

For each set and method combination we fit the camera model the set uses as ground truth or the closest one available for the given method (see SM for details). 
In practice, multiple different camera models can describe a single type of camera (\eg fisheye) well enough. 
In SM we show that for current camera calibration networks changing between similar models during fit often does not bring significant degradation in terms of the actual distortion modeled.

\subsection{Baselines}

We use GeoCalib~\cite{veicht2024geocalib}, AnyCalib~\cite{tirado2025anycalib}, and UniK3D~\cite{piccinelli2025unik3d} as baselines. We use models released by the authors. 
GeoCalib has separate models for pinhole and distorted cameras. 
We use the appropriate one based on the type of the benchmark set, denoted as GeoCalib$_\text{pin/dist}$.
UniK3D predicts ray maps and the implementation of the camera model fit is not public. We therefore use our implementation of AnyCalib's dense fitting strategy to obtain the camera parameters.
For AnyCalib, we evaluate its general version, which was trained on both pinhole and distorted cameras (denoted as AnyCalib$_\text{gen}$), and a combination of specialized versions AnyCalib$_\text{pin/dist}$.
Additionally, we train AnyCalib using our training schedule, full dataset, and implementation of the dense fit on the GPU, which we denote as AnyCalib$_\text{ours}$. 
In SM, we also evaluate monocular depth estimators~\cite{wang2025moge,wang2025moge2,kong2026moge3,knaebel2026surge,piccinelli2024unidepth,piccinelli2025unidepthv2} and feed-forward reconstruction models used with a single image~\cite{lin2025da3,keetha2026mapanything,wang2026vggtomega} as calibration networks for pinhole images. The results show that dedicated calibration methods outperform them.

\subsection{Results}
\label{sec:eval_results}

Table~\ref{tab:results} compares our network to the baselines.
Training on our data lowers AnyCalib's error by 40\% compared to the released model. Moreover, our model is over 5 times faster.
With the same data and efficient fit implementation, our network and AnyCalib's are on par in terms of accuracy, while our network takes only 9.7\,ms for inference and model fitting against 14.4\,ms for AnyCalib$_\text{ours}$. 
Results for each benchmark individually and with additional baselines are in SM.
We also perform an evaluation using AnyCalib's benchmark for images with principal points off-center (cropped images) and $a \neq 1$ (squashed images). The results are provided in SM and show that when trained on the same data our results are on par with those of AnyCalib.

\paragraph{Speed-accuracy Tradeoff} Fig.~\ref{fig:speed-acc} shows the angular error mAA against the total runtime on an A100 GPU and a Jetson Orin Nano 8\,GB with TensorRT for models with different backbones.
Our models occupy the Pareto-optimal frontier making them the fastest choice for a given accuracy level.
Compared to AnyCalib$_\text{ours}$, our method takes 33\% less time on the A100 for the ViT-L model and is 2.6 times faster on the Jetson for the ViT-S backbone.

\subsection{Ablation Experiments}
\label{sec:ablations}

\begin{table}[t]
\caption{\textbf{Representation Ablation.} Mean and standard deviation over seeds for the median error and the mAA (\%), each averaged over the ten benchmarks. Aux: the representation predicted by the auxiliary dense branch; \xmark: none. $^{*}$~significant against the first row with $p < 0.05$. RE and vFoV are provided in SM.}
\newcommand{\mk}[1]{\makebox[0.55em][l]{$^{#1}$}}
\centering
\setlength{\tabcolsep}{3pt}
\resizebox{\columnwidth}{!}{%
\begin{tabular}{lc cc cc}
\toprule
 & & \multicolumn{2}{c}{AE ($^\circ$)} & \multicolumn{2}{c}{hFoV ($^\circ$)} \\
\cmidrule(lr){3-4} \cmidrule(lr){5-6}
Variant & Aux & Median $\downarrow$ & mAA ($5^\circ$) $\uparrow$ & Median $\downarrow$ & mAA ($5^\circ$) $\uparrow$ \\
\midrule
$5 \times 5$ $\vec\omega$ & $\vec\theta$ & 1.165 $\pm$ 0.016\mk{} & 73.7 $\pm$ 0.3\mk{} & 3.01 $\pm$ 0.04\mk{} & 47.5 $\pm$ 0.3\mk{} \\
$5 \times 5$ $\vec\omega$, free $\vec{c}, a$ & $\vec\theta$ & 1.173 $\pm$ 0.015\mk{} & 73.6 $\pm$ 0.2\mk{} & 3.01 $\pm$ 0.04\mk{} & 47.4 $\pm$ 0.3\mk{} \\
$5 \times 5$ $\vec\omega$ & \xmark & 1.204 $\pm$ 0.021\mk{*} & 72.7 $\pm$ 0.5\mk{*} & 3.08 $\pm$ 0.05\mk{*} & 46.3 $\pm$ 0.3\mk{*} \\
$5 \times 5$ $\vec\omega$ & $\vec\omega$ & 1.183 $\pm$ 0.008\mk{} & 73.5 $\pm$ 0.1\mk{} & 3.05 $\pm$ 0.03\mk{} & 47.3 $\pm$ 0.2\mk{} \\
$7 \times 7$ $\vec\omega$ & $\vec\theta$ & 1.186 $\pm$ 0.012\mk{} & 73.3 $\pm$ 0.3\mk{*} & 3.08 $\pm$ 0.04\mk{*} & 47.1 $\pm$ 0.2\mk{} \\
$5 \times 5$ $\vec\theta$ & $\vec\theta$ & 1.174 $\pm$ 0.014\mk{} & 73.5 $\pm$ 0.4\mk{} & 3.06 $\pm$ 0.06\mk{} & 46.9 $\pm$ 0.4\mk{*} \\
$5 \times 5$ $\vec\theta$ & \xmark & 1.194 $\pm$ 0.024\mk{} & 73.1 $\pm$ 0.2\mk{*} & 3.06 $\pm$ 0.08\mk{} & 46.5 $\pm$ 0.2\mk{*} \\
Radial profile & $\vec\theta$ & 1.184 $\pm$ 0.015\mk{} & 73.4 $\pm$ 0.2\mk{} & 3.06 $\pm$ 0.05\mk{} & 47.0 $\pm$ 0.3\mk{*} \\
Direct regression & $\vec\theta$ & 1.204 $\pm$ 0.022\mk{*} & 73.3 $\pm$ 0.5\mk{} & 3.12 $\pm$ 0.07\mk{*} & 46.8 $\pm$ 0.5\mk{*} \\
\bottomrule
\end{tabular}}
\label{tab:ablations}
\end{table}

\begin{table}[t]
\caption{\textbf{Data Ablation.} Median error and mAA, each averaged over the ten benchmarks. OP$_\text{g}$: OpenPano V2~\cite{tirado2025anycalib}, Commons: our \commonspano{} dataset, MS: MegaScenes~\cite{tung2024megascenes}. WSD schedule was tuned for the full data pipeline.}
\centering
\resizebox{\columnwidth}{!}{%
\begin{tabular}{ll ccc cc}
\toprule
 & & \multicolumn{3}{c}{Data} & \multicolumn{2}{c}{AE ($^\circ$)} \\
\cmidrule(lr){3-5} \cmidrule(lr){6-7}
Schedule & Network & OP$_\text{g}$ & Commons & MS & Median $\downarrow$ & mAA ($5^\circ$) $\uparrow$ \\
\midrule
\multirow{2}{*}{Step decay~\cite{tirado2025anycalib}} & AnyCalib$_\text{gen}$ & $\checkmark$ & & & 1.975 & 58.5 \\
 & Ours & $\checkmark$ & & & 1.867 & 60.5 \\
\midrule
\multirow{4}{*}{WSD~\cite{hagele2024scaling}} & \multirow{4}{*}{Ours} & $\checkmark$ & & & 2.065 & 56.8 \\
 & & $\checkmark$ & $\checkmark$ & & 1.715 & 63.8 \\
 & & $\checkmark$ & & $\checkmark$ & 1.317 & 70.0 \\
 & & $\checkmark$ & $\checkmark$ & $\checkmark$ & \textbf{1.165} & \textbf{73.7} \\
\bottomrule
\end{tabular}}
\label{tab:data}
\end{table}

Table~\ref{tab:ablations} varies the output representations of the evaluated networks trained with the ViT-L backbone, schedule and data.
We consider variants without auxiliary dense supervision, auxiliary dense supervision using $\vec{\omega}$ instead of AnyCalib's $\vec{\theta}$, and, a variant using $7\times7$ sparse points instead of $5 \times 5$.
We also evaluate our network with a fit that estimates $\vec{c}$ and $a$ freely. 
As alternative output representations, we include a variant that outputs AnyCalib's $\vec\theta$ instead of $\vec\omega$ (see Sec.~\ref{sec:fov_field}), 
a variant inspired by~\cite{camposeco2015non,liao2021deep,dal2024revisiting} that outputs a 1D profile representing the ray angle at 12 points along the image half-diagonal, and a network that directly regresses the logarithm of focal length and the single parameter of the division model~\cite{fitzgibbon2001division}.

We train each model with five different seeds and show statistical differences.
The differences among the evaluated representations are small and mostly insignificant, with the exception of the direct regression network and the auxiliary dense supervision.
% Our $5 \times 5$ $\vec\omega$ has the lowest mean error over the ten benchmarks, but it is not the best variant on every benchmark: on three of the four pinhole benchmarks the $\vec\theta$ variant is slightly more accurate (see SM).
Most of the variants are statistically indistinguishable.
The direct regression of the division model is on par with the other variants on the pinhole benchmarks, but falls behind on the distorted benchmarks (see SM).
Estimating $\vec{c}$ and $a$ in the fit only slightly decreases the accuracy.
This decrease comes almost entirely from ScanNet++, which fits a four-parameter Kannala-Brandt model~\cite{kannala2006generic}.
These results suggest that, given our scale of training data, the specific output representation is not as impactful as long as it can sufficiently represent various models.

Table~\ref{tab:data} shows how the angular error changes with added training data. 
Both \commonspano{} and MegaScenes~\cite{tung2024megascenes} datasets significantly boost performance over OpenPano V2~\cite{tirado2025anycalib}. 
Unsurprisingly, data plays the most important role in terms of accuracy improvement.

\subsection{Downstream Applications}

Beyond direct evaluation of the estimated camera parameters, we also evaluate the performance of calibration networks in downstream tasks: relative pose estimation~\cite{kocur2026minimal}, SfM~\cite{schoenberger2016sfm}, and feed-forward 3D reconstruction~\cite{keetha2026mapanything,wang2025pi3}. Extended results, including localization~\cite{sattler2018benchmarking} and details for all evaluation setups are provided in the SM.

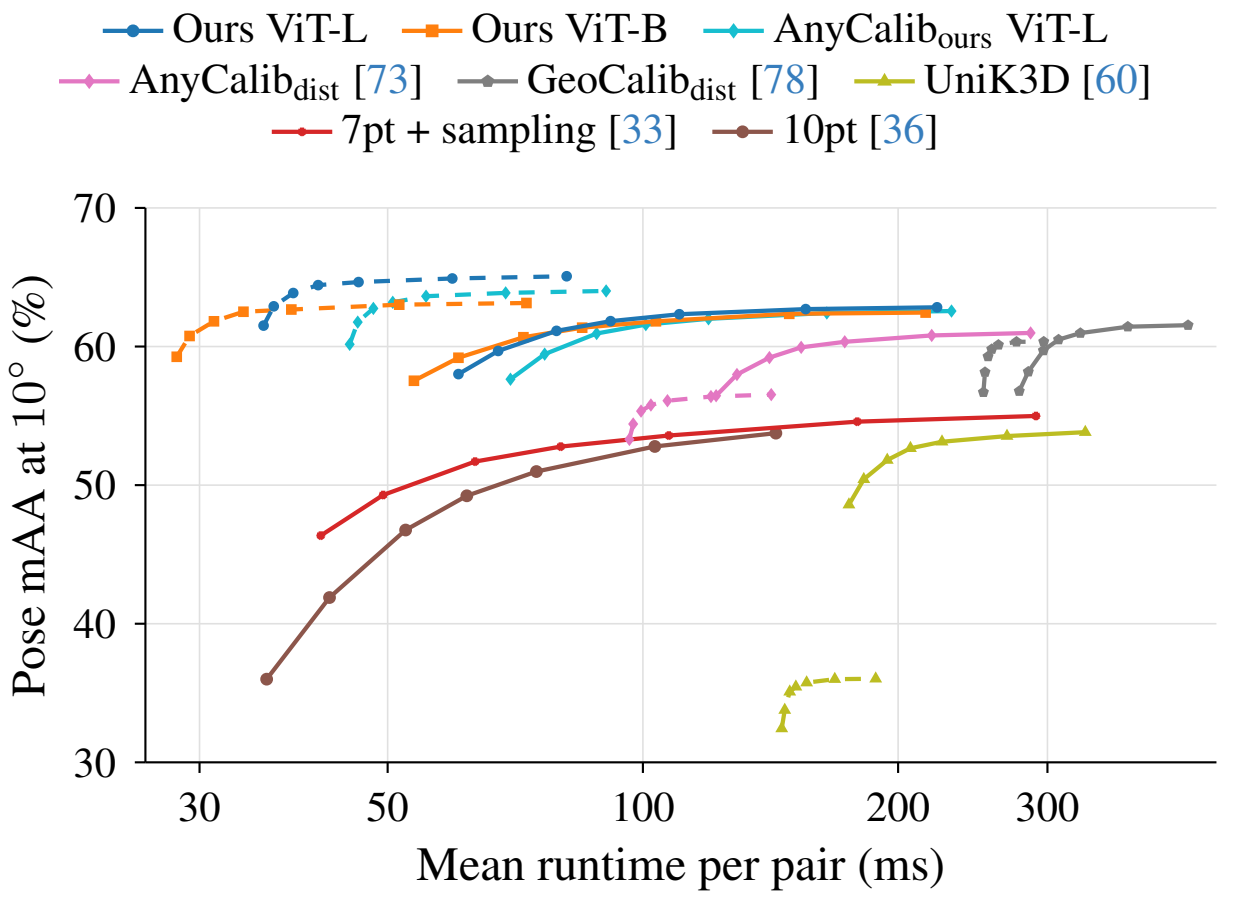
\begin{figure}[t]
\centering
\input{fig/pose_different}
\caption{\textbf{Relative Pose Estimation of Radially Distorted Cameras.} Pose mAA at $10^\circ$ over 10\,000 image pairs of Cathedral~\cite{kocur2026minimal} against the mean runtime per pair, one point per RANSAC budget from 10 to 1000 PoseLib~\cite{PoseLib} iterations using SuperPoint~\cite{detone2018superpoint} and LightGlue~\cite{lindenberger2023lightglue} matches. 
A network adds its forward pass and fit on an A100, twice per pair to RANSAC runtime on the CPU.
The intrinsics provided by networks were combined with the 5-point solver~\cite{nister2004efficient} to obtain initial models with local optimization either only for pose (dashed lines) or also for the intrinsics (solid lines).
}
\label{fig:pose_different}
\end{figure}

\paragraph{Relative Pose Estimation} 
We evaluate the compared networks for the task of two-view relative pose estimation with radially distorted cameras.
We use the Cathedral scene, evaluation setup and baselines from \cite{kocur2026minimal}.
Fig.~\ref{fig:pose_different} shows the speed-accuracy tradeoff of different methods including the standard non-minimal solver~\cite{kukelova2015efficient}, sampling-based solver~\cite{kocur2026minimal}, and the 5-point solver~\cite{nister2004efficient} used together with the priors for focal length and radial distortion provided by camera calibration networks.
Our network in combination with the 5-point solver beats all other methods both in terms of speed and accuracy.
Poses obtained using our camera priors are harmed when the intrinsics are optimized within local optimization, showing that it provides a stronger prior than the outlier-contaminated two-view geometry.
% This suggests that the estimated intrinsics are better than what can be estimated from two views using the provided matches~\cite{detone2018superpoint, lindenberger2023lightglue}.

\begin{table}[t]
\centering
\caption{\textbf{Structure-from-Motion.} mAA of the relative pose error over all image pairs and the share of registered images (\%), averaged over 100 subsets of 10 images from each of five PhotoTourism~\cite{jin2021image} scenes and
% five EuRoC~\cite{burri2016euroc} sequences, 
the Rotunda scene~\cite{kocur2026minimal}; mean $\pm$ standard deviation over 5 runs of COLMAP.}
\setlength{\tabcolsep}{4pt}
\resizebox{\columnwidth}{!}{%
\begin{tabular}{c l cc cc}
\toprule
 &  & \multicolumn{2}{c}{Incremental~\cite{schoenberger2016sfm}} & \multicolumn{2}{c}{Global~\cite{pan2024glomap}} \\
\cmidrule(lr){3-4} \cmidrule(lr){5-6}
 & Prior & mAA($10^\circ$) $\uparrow$ & Reg. $\uparrow$ & mAA($10^\circ$) $\uparrow$ & Reg. $\uparrow$ \\
\midrule
\multirow{5}{*}{\rotatebox[origin=c]{90}{\textit{\begin{tabular}{@{}c@{}}Photo-\\Tourism\end{tabular}}}} & None & 66.2 $\pm$ 0.5 & \textbf{99.8} $\pm$ 0.1 & 63.6 $\pm$ 0.5 & 98.9 $\pm$ 0.2 \\
 & GeoCalib$_\text{pinhole}$~\cite{veicht2024geocalib} & 68.3 $\pm$ 0.5 & \textbf{99.8} $\pm$ 0.1 & 63.3 $\pm$ 0.3 & 99.3 $\pm$ 0.2 \\
 & AnyCalib$_\text{pinhole}$~\cite{tirado2025anycalib} & 69.2 $\pm$ 0.6 & \textbf{99.8} $\pm$ 0.1 & 68.4 $\pm$ 0.4 & 99.4 $\pm$ 0.3 \\
 & AnyCalib$_\text{ours}$ & \textbf{70.8} $\pm$ 0.1 & \textbf{99.8} $\pm$ 0.1 & 71.4 $\pm$ 0.2 & \textbf{99.9} $\pm$ 0.1 \\
 & Ours & \textbf{70.8} $\pm$ 0.3 & \textbf{99.8} $\pm$ 0.1 & \textbf{72.3} $\pm$ 0.1 & 99.6 $\pm$ 0.2 \\
\midrule
% \multirow{5}{*}{\rotatebox[origin=c]{90}{\textit{EuRoC}}} & None & 72.6 $\pm$ 0.6 & 95.4 $\pm$ 0.1 & 79.5 $\pm$ 0.3 & 98.6 $\pm$ 0.2 \\
%  & GeoCalib$_\text{dist}$~\cite{veicht2024geocalib} & 75.5 $\pm$ 0.5 & 95.6 $\pm$ 0.3 & 81.2 $\pm$ 0.2 & \textbf{99.5} $\pm$ 0.0 \\
%  & AnyCalib$_\text{dist}$~\cite{tirado2025anycalib} & 75.9 $\pm$ 0.2 & 95.3 $\pm$ 0.2 & \textbf{81.5} $\pm$ 0.1 & \textbf{99.5} $\pm$ 0.0 \\
%  & AnyCalib$_\text{ours}$ & 76.0 $\pm$ 0.4 & 95.6 $\pm$ 0.2 & 81.4 $\pm$ 0.1 & 99.4 $\pm$ 0.1 \\
%  & Ours & \textbf{76.3} $\pm$ 0.5 & \textbf{95.8} $\pm$ 0.1 & 81.4 $\pm$ 0.1 & \textbf{99.5} $\pm$ 0.1 \\
% \midrule
\multirow{5}{*}{\rotatebox[origin=c]{90}{\textit{Rotunda}}} & None & 32.6 $\pm$ 0.9 & 67.0 $\pm$ 0.5 & 32.1 $\pm$ 1.2 & 92.7 $\pm$ 1.2 \\
 % & EXIF & 36.4 $\pm$ 0.6 & 68.4 $\pm$ 0.4 & 35.1 $\pm$ 0.8 & 91.4 $\pm$ 0.8 \\
 & GeoCalib$_\text{dist}$~\cite{veicht2024geocalib} & 38.9 $\pm$ 0.4 & 71.1 $\pm$ 0.3 & 40.0 $\pm$ 0.8 & \textbf{98.2} $\pm$ 0.1 \\
 & AnyCalib$_\text{dist}$~\cite{tirado2025anycalib} & \textbf{40.8} $\pm$ 0.3 & 71.6 $\pm$ 0.1 & 45.4 $\pm$ 0.3 & 97.2 $\pm$ 0.2 \\
 & AnyCalib$_\text{ours}$ & 40.7 $\pm$ 0.2 & 71.7 $\pm$ 0.2 & 45.3 $\pm$ 0.4 & 96.6 $\pm$ 0.5 \\
 & Ours & 40.6 $\pm$ 0.4 & \textbf{71.8} $\pm$ 0.2 & \textbf{47.3} $\pm$ 0.3 & 96.8 $\pm$ 0.0 \\
\bottomrule
\end{tabular}}
\label{tab:sfm}
\end{table}

\paragraph{Structure-from-Motion and 3D Reconstruction} 
\label{sec:sfm}%
We evaluate whether a calibration prior improves SfM and feed-forward 3D reconstruction on internet photos, for which EXIF metadata is often missing or does not provide distortion priors. We evaluate COLMAP (both incremental~\cite{schoenberger2016sfm} and global~\cite{pan2024glomap}) on five PhotoTourism~\cite{jin2021image} test scenes that do not appear in our training data and the Rotunda scene from~\cite{kocur2026minimal}. We also evaluate four feed-forward multi-view reconstruction methods~\cite{wang2025pi3,wang2026vggtomega,keetha2026mapanything,lin2025da3} on the Cathedral scene~\cite{kocur2026minimal} and the Zip-NeRF dataset~\cite{barron2023zipnerf} (using images before undistortion). For each scene we sample 100 subsets containing 10 images. We compare the methods with priors provided by camera calibration networks against no prior.

The results are shown in Tab.~\ref{tab:sfm} for COLMAP and Tab.~\ref{tab:recon} for the feed-forward models. For COLMAP using the priors from camera calibration networks leads to improvements with AnyCalib and our method performing best. Without camera priors the feed-forward methods perform poorly. With priors added the accuracy rises significantly with our method performing best or close to best.

\begin{table}[t]
\caption{\textbf{Feed-Forward 3D Reconstruction of Distorted Images.} mAA($5^\circ$) $\uparrow$ of the relative pose error over all image pairs, averaged over 100 subsets of 10 frames from each of four Zip-NeRF~\cite{barron2023zipnerf} scenes and over 100 subsets of 10 covisible images of Cathedral~\cite{kocur2026minimal}. $^\dagger$ Takes no intrinsics: with a prior it gets the undistorted frames only.}
\centering
\setlength{\tabcolsep}{4pt}
\resizebox{\columnwidth}{!}{%
\begin{tabular}{c l cccc}
\toprule
 & Prior & Pi3X~\cite{wang2025pi3} & MapAnything~\cite{keetha2026mapanything} & DA3$^\dagger$~\cite{lin2025da3} & VGGT-$\Omega^\dagger$~\cite{wang2026vggtomega} \\
\midrule
\multirow{6}{*}{\rotatebox[origin=c]{90}{\textit{Zip-NeRF}}} & None & 4.5 & 4.0 & 2.9 & 3.5 \\
 & GeoCalib$_\text{dist}$~\cite{veicht2024geocalib} & 31.9 & 32.1 & 48.9 & 44.2 \\
 & UniK3D~\cite{piccinelli2025unik3d} & 23.3 & 23.9 & 31.4 & 27.6 \\ 
 & AnyCalib$_\text{dist}$~\cite{tirado2025anycalib} & 57.9 & 40.7 & 66.7 & 61.2 \\
 & AnyCalib$_\text{ours}$ & \textbf{68.9} & \textbf{41.1} & 69.3 & 64.1 \\
 & Ours & 67.6 & \textbf{41.1} & \textbf{69.5} & \textbf{64.4} \\
\midrule
\multirow{6}{*}{\rotatebox[origin=c]{90}{\textit{Cathedral}}} & None & 14.4 & 10.4 & 13.8 & 10.7 \\
 & GeoCalib$_\text{dist}$~\cite{veicht2024geocalib} & 33.9 & 26.6 & 31.6 & 36.7 \\
 & UniK3D~\cite{piccinelli2025unik3d} & 8.7 & 7.9 & 12.2 & 10.4 \\ 
 & AnyCalib$_\text{dist}$~\cite{tirado2025anycalib} & 29.7 & 25.0 & 32.3 & \textbf{37.6} \\
 & AnyCalib$_\text{ours}$ & 38.0 & 30.5 & 31.9 & 37.5 \\
 & Ours & \textbf{40.1} & \textbf{31.9} & \textbf{32.5} & \textbf{37.6} \\
\bottomrule
\end{tabular}}
\label{tab:recon}
\end{table}

\section{Conclusion}

We have shown that using intermediate dense representations is not necessary for camera calibration networks and using a sparse representation is more computationally efficient while achieving the same accuracy.
Our method is intentionally simple and builds on existing components. 
We show that this is sufficient as the network accuracy can be boosted the most by improving the training data, which we help by contributing our own dataset.
We also show how camera calibration networks can aid performance in multiple downstream 3D vision tasks.
As a limitation, outside of one experiment with cropped and squashed images, our method assumes radially symmetric distortion with the principal point at the image center and square pixels; currently it does not model small realistic deviations from these assumptions (\eg, a slightly off-center principal point) or tangential distortion, which we leave for future work.
 
% we see this as a strength rather than a shortcoming, since our results show that a careful reduction of the output representation combined with large-scale training data is sufficient to match the state of the art at a fraction of its cost, without the need for more elaborate mechanisms.

% \clearpage
\section*{LLM Usage Statement}

We have used Claude Code with Opus 5.5 to implement the method, evaluation and data scrapers. We have also used it to assist with writing, generate figures and tables.

\section*{Acknowledgments}

This research was supported by the EU NextGenerationEU through the Recovery and Resilience Plan for Slovakia under the project No. 09I02-03-V01-00029 and the Czech Science Foundation (GACR) through JUNIOR STAR Grant No. 22-23183M. 
Part of research results were obtained using computational resources procured within the National Competence Centre for High-Performance Computing project (Project Code: 311070AKF2), funded by the European Regional Development Fund under the Operational Programme Integrated Infrastructure 2014–2020. Part of the research results were obtained using computational resources funded by the European Union – NextGenerationEU through the Recovery and Resilience Plan for Slovakia under Project No. 17I03-04-P02-00001: Development and Construction of a Supercomputer for the National Supercomputing Centre.

{
    \small
    \bibliographystyle{ieeenat_fullname}
    \bibliography{main}
}

\input{X_suppl}

\end{document}

%% file: preamble.tex
\usepackage{bm}

\usepackage{booktabs}
\usepackage{pifont}
\usepackage{multirow}
\usepackage{placeins}
\usepackage{afterpage}

\newcommand{\xmark}{\ding{55}}

\usepackage{tikz}
\usepackage{pgfplots}
\pgfplotsset{compat=1.18}
\tikzset{ptsicon/.style={scale=0.25, baseline={([yshift=-0.6ex]current bounding box.center)}}}

\renewcommand{\vec}{\bm}
\newcommand{\commonspano}[0]{\textit{CommonsPano}}
\usepackage{microtype}

\renewcommand{\paragraph}[1]{\vspace{.5em}\noindent\textbf{#1.}}

%% file: fig/speed_acc.tex
% (a) Jetson Orin Nano 8 GB, 15 W, batch 1: TensorRT fp16 engine + fit, docs/results.md "Latency on a Jetson Orin Nano" (eval/paper_speed_acc.py JETSON), the accuracy of panel (b)
% (b) total = fp16 net + fit, results/timing/eval_timing_sa_acsq.json (Devana job 1912345, one A100), GeoCalib from eval_geocalib_*_ae.json as tab:results
% mAA@5 over the ten benchmarks, the mean over five seeds where they exist (no whiskers), otherwise seed 0 (ours ViT-L, AnyCalib ViT-S and ViT-L; results/ac_vits_seeds/)
\begin{tikzpicture}
\definecolor{cOursL}{HTML}{1F77B4}
\definecolor{cOursC}{HTML}{FF7F0E}
\definecolor{cAny}{HTML}{E377C2}
\definecolor{cAnyG}{HTML}{8C564B}
\definecolor{cGeo}{HTML}{7F7F7F}
\definecolor{cUni}{HTML}{BCBD22}
\pgfplotsset{speedacc/.style={
  scale only axis, width=0.86\linewidth, height=0.34\linewidth,
  ymin=45, ymax=80, minor xtick={},
  ytick={45,50,55,60,65,70,75,80}, yticklabels={45,50,55,60,65,70,75,80},
  label style={font=\small}, tick label style={font=\footnotesize},
  xlabel style={yshift=3pt}, ylabel={AE mAA at $5^\circ$ (\%)}, ylabel style={yshift=-3pt},
  axis lines=left, axis line style={-, line width=0.4pt}, tick style={black, line width=0.4pt}, tick align=outside,
  major tick length=2pt, grid=major, grid style={black!10, line width=0.3pt},
  clip mode=individual}}
\begin{axis}[xmode=log, log basis x=10, speedacc, name=axJ, xmin=8, xmax=2500,
  xtick={10,20,50,100,200,500,1000,2000}, xticklabels={10,20,50,100,200,500,1000,2000},
  xlabel={Total runtime on Jetson Orin Nano (ms)}]
\addplot[cGeo, only marks, mark=pentagon*, mark size=2.7pt] coordinates {(1466.5,47.4)};
\addplot[cUni, only marks, mark=triangle*, mark size=2.9pt] coordinates {(443.28,50.5)};
\addplot[cAnyG, only marks, mark=diamond*, mark size=2.9pt] coordinates {(400.50,58.5)};
\addplot[cAny, line width=0.9pt, mark=diamond*, mark size=2.9pt] coordinates {(31.50,60.94) (100.60,73.86)};
\addplot[cOursL, line width=0.9pt, mark=*, mark size=2.3pt] coordinates {(12.30,62.21) (27.10,66.6) (79.30,73.68)};
\begin{scope}[every node/.style={font=\scriptsize, text=black!75, inner sep=1.5pt}]
\node[anchor=north, yshift=-2.5pt] at (axis cs:12.30,62.21) {ViT-S};
\node[anchor=east, xshift=-3.5pt] at (axis cs:27.10,66.62) {ViT-B};
\node[anchor=east, xshift=-3.5pt] at (axis cs:79.30,73.68) {ViT-L};
\node[anchor=north west, xshift=2pt, yshift=-2pt] at (axis cs:31.50,60.94) {ViT-S};
\node[anchor=west, xshift=4pt] at (axis cs:100.60,73.86) {ViT-L};
\draw[-stealth, black!55, line width=0.5pt, shorten <=2pt, shorten >=2pt] (axis cs:31.50,56.0) -- (axis cs:12.30,56.0) node[midway, below, inner sep=1pt] {2.6$\times$};
\draw[stealth-, black!55, line width=0.5pt, shorten <=3pt] (axis cs:400.50,58.5) -- ++(18pt,16pt) node[anchor=south, inner sep=1pt] {Previous SOTA};
\end{scope}
\end{axis}
\begin{axis}[xmode=log, log basis x=10, speedacc, name=axA, at={(axJ.outer south west)}, anchor=outer north west, yshift=-8pt,
  xmin=4.4, xmax=130, xtick={5,10,20,50,100}, xticklabels={5,10,20,50,100},
  xlabel={Total runtime on A100 (ms)}]
\addplot[cGeo, only marks, mark=pentagon*, mark size=2.7pt] coordinates {(102.8,47.4)};
\addplot[cUni, only marks, mark=triangle*, mark size=2.9pt] coordinates {(41.51,50.5)};
\addplot[cAnyG, only marks, mark=diamond*, mark size=2.9pt] coordinates {(52.70,58.5)};
\addplot[cAny, line width=0.9pt, mark=diamond*, mark size=2.9pt] coordinates {(9.36,60.94) (15.05,73.86)};
\addplot[cOursL, line width=0.9pt, mark=*, mark size=2.3pt] coordinates {(6.31,62.21) (6.61,66.6) (10.47,73.68)};
\begin{scope}[every node/.style={font=\scriptsize, text=black!75, inner sep=1.5pt}]
\node[anchor=north, yshift=-2.5pt] at (axis cs:6.31,62.21) {ViT-S};
\node[anchor=east, xshift=-3.5pt] at (axis cs:6.61,66.62) {ViT-B};
\node[anchor=east, xshift=-3.5pt] at (axis cs:10.47,73.68) {ViT-L};
\node[anchor=north west, xshift=2pt, yshift=-2pt] at (axis cs:9.36,60.94) {ViT-S};
\node[anchor=west, xshift=4pt] at (axis cs:15.05,73.86) {ViT-L};
\draw[-stealth, black!55, line width=0.5pt, shorten <=2pt, shorten >=2pt] (axis cs:15.05,77.3) -- (axis cs:10.47,77.3) node[midway, above, inner sep=1pt] {1.4$\times$};
\end{scope}
\end{axis}
\newcommand{\lgd}[1]{\tikz[baseline=-0.6ex]\draw[#1, mark indices={2}] plot coordinates {(0,0) (0.2,0) (0.4,0)};}
\node[anchor=south, inner sep=0pt, yshift=4pt] at (axJ.outer north) {\footnotesize\begin{tabular}{@{}c@{}}
 \mbox{\lgd{cOursL, line width=0.9pt, mark=*, mark size=1.7pt}\,Ours}\hspace{7pt}\mbox{\lgd{cAny, line width=0.9pt, mark=diamond*, mark size=2.0pt}\,AnyCalib$_\text{ours}$}\hspace{7pt}\mbox{\lgd{cAnyG, only marks, mark=diamond*, mark size=2.0pt}\,AnyCalib$_\text{gen}$~\cite{tirado2025anycalib}} \\
 \mbox{\lgd{cGeo, only marks, mark=pentagon*, mark size=2.0pt}\,GeoCalib$_\text{pin/dist}$~\cite{veicht2024geocalib}}\hspace{7pt}\mbox{\lgd{cUni, only marks, mark=triangle*, mark size=2.0pt}\,UniK3D~\cite{piccinelli2025unik3d}}
\end{tabular}};
\end{tikzpicture}

%% file: fig/pose_different.tex
% generated by rdnet/plot_calib_net_tikz.py from calib_net_cathedral_ray.csv
\begin{tikzpicture}
\definecolor{cOursL}{HTML}{1F77B4}
\definecolor{cOursC}{HTML}{FF7F0E}
\definecolor{cAny}{HTML}{E377C2}
\definecolor{cAnyO}{HTML}{17BECF}
\definecolor{cGeo}{HTML}{7F7F7F}
\definecolor{cUni}{HTML}{BCBD22}
\definecolor{cSix}{HTML}{2CA02C}
\definecolor{cSeven}{HTML}{D62728}
\definecolor{cNine}{HTML}{9467BD}
\definecolor{cTen}{HTML}{8C564B}
\begin{axis}[
  scale only axis, width=0.85\linewidth, height=0.44\linewidth,
  xmode=log, log basis x=10,
  xmin=25.9, xmax=474.7, ymin=30, ymax=70,
  xtick={30,50,100,200,300,500}, xticklabels={30,50,100,200,300,500}, minor xtick={},
  ytick={30,40,50,60,70}, yticklabels={30,40,50,60,70},
  xlabel={Mean runtime per pair (ms)}, ylabel={Pose mAA at $10^\circ$ (\%)},
  label style={font=\small}, tick label style={font=\footnotesize},
  xlabel style={yshift=2pt}, ylabel style={yshift=-3pt},
  axis lines=left, axis line style={-, line width=0.4pt}, tick style={black, line width=0.4pt}, tick align=outside,
  major tick length=2pt, grid=major, grid style={black!12, line width=0.3pt},
  clip mode=individual, name=ax,
]
\addplot[cTen, line width=0.8pt, mark=*, mark size=0.8pt] coordinates {(36.0,36.00) (42.7,41.89) (52.5,46.76) (62.0,49.23) (74.9,50.98) (103.3,52.79) (143.5,53.75)};
\addplot[cSeven, line width=0.8pt, mark=star, mark size=0.8pt] coordinates {(41.7,46.36) (49.4,49.29) (63.4,51.70) (80.0,52.78) (107.3,53.58) (179.0,54.58) (290.8,54.99)};
\addplot[cUni, dashed, line width=0.8pt, mark=triangle*, mark size=0.6pt, mark options={solid}] coordinates {(145.8,32.44) (147.0,33.77) (149.0,35.09) (151.5,35.45) (156.0,35.75) (168.4,36.00) (188.2,36.03)};
\addplot[cUni, solid, line width=0.8pt, mark=triangle*, mark size=0.6pt, mark options={solid}] coordinates {(175.0,48.59) (182.2,50.42) (194.2,51.81) (206.8,52.66) (225.3,53.13) (268.8,53.54) (332.5,53.82)};
\addplot[cGeo, dashed, line width=0.8pt, mark=pentagon*, mark size=0.6pt, mark options={solid}] coordinates {(252.1,56.70) (253.2,58.15) (255.3,59.28) (257.9,59.81) (262.6,60.12) (275.8,60.34) (297.0,60.35)};
\addplot[cGeo, solid, line width=0.8pt, mark=pentagon*, mark size=0.6pt, mark options={solid}] coordinates {(277.9,56.78) (284.8,58.18) (296.7,59.71) (309.2,60.50) (328.1,60.97) (373.2,61.43) (439.5,61.54)};
\addplot[cAny, dashed, line width=0.8pt, mark=diamond*, mark size=0.6pt, mark options={solid}] coordinates {(96.4,53.30) (97.4,54.41) (99.5,55.34) (102.2,55.78) (106.9,56.09) (120.3,56.39) (141.7,56.53)};
\addplot[cAny, solid, line width=0.8pt, mark=diamond*, mark size=0.6pt, mark options={solid}] coordinates {(122.0,56.45) (129.1,57.98) (141.0,59.20) (153.7,59.94) (173.0,60.34) (219.1,60.80) (286.6,60.98)};
\addplot[cAnyO, dashed, line width=0.8pt, mark=diamond, mark size=0.6pt, mark options={solid}] coordinates {(45.1,60.16) (46.1,61.76) (48.1,62.75) (50.7,63.19) (55.5,63.63) (68.9,63.87) (90.5,64.01)};
\addplot[cAnyO, solid, line width=0.8pt, mark=diamond, mark size=0.6pt, mark options={solid}] coordinates {(69.8,57.64) (76.6,59.45) (88.2,60.94) (100.8,61.58) (119.5,62.00) (164.8,62.42) (231.2,62.56)};
\addplot[cOursC, dashed, line width=0.8pt, mark=square*, mark size=0.6pt, mark options={solid}] coordinates {(28.2,59.26) (29.2,60.76) (31.2,61.82) (33.8,62.51) (38.5,62.67) (51.6,63.02) (72.9,63.14)};
\addplot[cOursC, solid, line width=0.8pt, mark=square*, mark size=0.6pt, mark options={solid}] coordinates {(53.7,57.53) (60.6,59.19) (72.3,60.67) (84.8,61.36) (103.6,61.82) (148.8,62.36) (215.5,62.45)};
\addplot[cOursL, dashed, line width=0.8pt, mark=*, mark size=0.6pt, mark options={solid}] coordinates {(35.7,61.51) (36.7,62.90) (38.7,63.86) (41.4,64.43) (46.2,64.65) (59.6,64.91) (81.3,65.07)};
\addplot[cOursL, solid, line width=0.8pt, mark=*, mark size=0.6pt, mark options={solid}] coordinates {(60.6,58.01) (67.5,59.68) (79.1,61.13) (91.6,61.83) (110.4,62.33) (155.6,62.70) (222.3,62.83)};
\end{axis}
\node[anchor=south, inner sep=0pt, yshift=3pt] at (ax.outer north) {\footnotesize\begin{tabular}{@{}c@{}}
 \mbox{\tikz[baseline=-0.6ex]\draw[cOursL, line width=0.8pt, mark=*, mark size=0.8pt, mark indices={2}] plot coordinates {(0,0) (0.2,0) (0.4,0)};\,Ours ViT-L}\hspace{6pt}\mbox{\tikz[baseline=-0.6ex]\draw[cOursC, line width=0.8pt, mark=square*, mark size=0.8pt, mark indices={2}] plot coordinates {(0,0) (0.2,0) (0.4,0)};\,Ours ViT-B}\hspace{6pt}\mbox{\tikz[baseline=-0.6ex]\draw[cAnyO, line width=0.8pt, mark=diamond, mark size=0.8pt, mark indices={2}] plot coordinates {(0,0) (0.2,0) (0.4,0)};\,AnyCalib$_\text{ours}$ ViT-L} \\
 \mbox{\tikz[baseline=-0.6ex]\draw[cAny, line width=0.8pt, mark=diamond*, mark size=0.8pt, mark indices={2}] plot coordinates {(0,0) (0.2,0) (0.4,0)};\,AnyCalib$_\text{dist}$~\cite{tirado2025anycalib}}\hspace{6pt}\mbox{\tikz[baseline=-0.6ex]\draw[cGeo, line width=0.8pt, mark=pentagon*, mark size=0.8pt, mark indices={2}] plot coordinates {(0,0) (0.2,0) (0.4,0)};\,GeoCalib$_\text{dist}$~\cite{veicht2024geocalib}}\hspace{6pt}\mbox{\tikz[baseline=-0.6ex]\draw[cUni, line width=0.8pt, mark=triangle*, mark size=0.8pt, mark indices={2}] plot coordinates {(0,0) (0.2,0) (0.4,0)};\,UniK3D~\cite{piccinelli2025unik3d}} \\
 \mbox{\tikz[baseline=-0.6ex]\draw[cSeven, line width=0.8pt, mark=star, mark size=0.8pt, mark indices={2}] plot coordinates {(0,0) (0.2,0) (0.4,0)};\,7pt + sampling~\cite{kocur2026minimal}}\hspace{6pt}\mbox{\tikz[baseline=-0.6ex]\draw[cTen, line width=0.8pt, mark=*, mark size=0.8pt, mark indices={2}] plot coordinates {(0,0) (0.2,0) (0.4,0)};\,10pt~\cite{kukelova2015efficient}}
\end{tabular}};
\end{tikzpicture}

%% file: X_suppl.tex
\clearpage
\setcounter{page}{1}
\maketitlesupplementary

\section{Overview}

In this Supplementary Material we provide additional information related to the main paper including extended results and experiments. 
For clarity, all table and section labels of SM start with S.
In Sec.~\ref{sec:suppl_cams} we provide more details on the camera models used for evaluation (Sec.~\ref{sec:metrics} the main paper).
Section~\ref{sec:suppl_sparse} provides extended results for Sec.~\ref{sec:sparse_results} and Table~\ref{tab:sparse} in the main paper which show that using only a sparse subset of points during camera model fitting is sufficient.
In Sec.~\ref{sec:suppl_net} we provide more details on the neural network architecture and training.
Sec.~\ref{sec:suppl_data} provides more information on the data used for both training and evaluation.
Extended results for direct evaluation of the estimated camera models and more details about the evaluation details are provided in Sec.~\ref{sec:suppl_eval}. 
The extended results show comparison for individual benchmarks, more metrics and comparison against MDEs~\cite{wang2025moge,wang2025moge2,kong2026moge3,piccinelli2024unidepth,piccinelli2025unidepthv2,knaebel2026surge} and feed-forward reconstruction methods~\cite{wang2025pi3,wang2026vggtomega,lin2025da3,keetha2026mapanything}. 
We also show (Sec.~\ref{sec:suppl_edit}) how our method performs when trained and evaluated on edited images where the assumptions of the principal point $\vec{c}$ in center and aspect ratio $a = 1$ do not hold.
Sec.~\ref{sec:suppl_downstream} provides more details on the downstream evaluation as well as additional experiments including localization (Sec.~\ref{sec:suppl_loc}).

\section{Camera Models}
\label{sec:suppl_cams}
Tab.~\ref{tab:cams} lists the camera models whose intrinsics are fitted in our experiments and the acronyms used for them in the following tables, where BC-$n$ and KB-$n$ have $n$ distortion coefficients. Besides its distortion parameters, every model has the focal length $f$, the principal point $\vec{c}$ and the aspect ratio $a$ of \eqref{eq:ac_proj}. Our network can fit the same camera models as AnyCalib~\cite{tirado2025anycalib}. We therefore follow its evaluation strategy: each benchmark is fitted with one model by default. GeoCalib's distorted weights support only BC-1 and the division model, so GeoCalib fits BC-1 on InFlux and InFlux++ and the division model on ScanNet++ and Mono. EUCM appears only among the cameras of our training views.

\begin{table}[t]
\caption{\textbf{Camera models.} The fitted camera models, their acronyms, their number of distortion parameters (Dist.) and the benchmarks fitted with them by default.}
\centering
\setlength{\tabcolsep}{4pt}
\resizebox{\linewidth}{!}{%
\begin{tabular}{l l c l}
\toprule
Camera model & Acronym & Dist. & Benchmarks \\
\midrule
Pinhole & Pinhole & 0 & S2D3D, TartanAir, MegaDepth, LaMAR \\
Brown-Conrady~\cite{brown1971close} & BC-$n$ & $n$ & MD-radial (BC-1), InFlux(++) (BC-2) \\
Kannala-Brandt~\cite{kannala2006generic} & KB-$n$ & $n$ & ScanNet++ (KB-4) \\
Unified camera model~\cite{geyer2000unifying,mei2007single} & UCM & 1 & Mono \\
Enhanced UCM~\cite{khomutenko2016enhanced} & EUCM & 2 & -- \\
Division~\cite{fitzgibbon2001division} & Division & 1 & Cathedral \\
\bottomrule
\end{tabular}}
\label{tab:cams}
\end{table}

\input{tab/sparse_full}

\section{Sparse Fields: All Results}
\label{sec:suppl_sparse}
Tab.~\ref{tab:sparse_full} gives AnyCalib's fit on every point layout, per benchmark, in its released
PyTorch implementation and compiled with Numba: the mean angular error of the released calibrator on the
dense field, the change of it for each fit, and the fit time for each fitted camera model on the CPU and the
GPU. 
Similar to the results shown in the main paper, the extended evaluation shows that using only a sparse subset of points is sufficient with the exception of the $3 \times 3$ configuration on ScanNet++~\cite{yeshwanth2023scannetpp} where the four parameter Kannala-Brandt model~\cite{kannala2006generic} is used as ground truth. 
The results also show that using 10 instead of 2 iterations does not bring improvements in terms of accuracy while increasing the runtime.
Moreover, fixing $\vec{c}$ to image center and $a=1$ brings slight improvement in terms of accuracy.
This is expected as the evaluated model was trained using only synthetically rendered views where these assumptions were true.
Fixing $\vec{c}$ and $a$ during optimization also brings small improvements in terms of runtime.

\begin{table}[t]
\caption{\textbf{Sparse fits of AnyCalib's field on edited images.} AnyCalib's fit in its released
implementation (PyTorch, GPU) and our implementation of AnyCalib's fit compiled with Numba, on the same points. Median angular error: in degrees
for the released implementation on the dense field, as the change relative to it otherwise. Iter.:
Gauss-Newton iterations.}
\centering
\setlength{\tabcolsep}{3pt}
\renewcommand{\arraystretch}{1.1}
\resizebox{\linewidth}{!}{%
\begin{tabular}{llrc ccc}
\toprule
Fit & Points & \#pts & Iter. & S2D3D $\downarrow$ & TartanAir $\downarrow$ & LaMAR $\downarrow$ \\
\midrule
% rows sparse_edit: begin
AnyCalib~\cite{tirado2025anycalib} & Dense & $\sim\!10^5$ & 10 & 7.44$^\circ$ & 9.86$^\circ$ & 9.46$^\circ$ \\
\midrule
\multirow{12}{*}{\shortstack{AnyCalib\\(Numba)}} & \multirow{2}{*}{Dense} & \multirow{2}{*}{$\sim\!10^5$} & 2 & $0.0\%$ & $0.0\%$ & $0.0\%$ \\
 &  &  & 10 & $0.0\%$ & $0.0\%$ & $0.0\%$ \\
\cmidrule(lr){2-7}
 & \multirow{2}{*}{$99 \times 99$} & \multirow{2}{*}{9801} & 2 & $0.0\%$ & $0.0\%$ & $0.0\%$ \\
 &  &  & 10 & $0.0\%$ & $0.0\%$ & $0.0\%$ \\
\cmidrule(lr){2-7}
 & \multirow{2}{*}{$25 \times 25$} & \multirow{2}{*}{625} & 2 & $+0.1\%$ & $+0.1\%$ & $-0.1\%$ \\
 &  &  & 10 & $+0.1\%$ & $+0.1\%$ & $-0.1\%$ \\
\cmidrule(lr){2-7}
 & \multirow{2}{*}{$7 \times 7$} & \multirow{2}{*}{49} & 2 & $+0.3\%$ & $+0.2\%$ & $0.0\%$ \\
 &  &  & 10 & $+0.3\%$ & $+0.2\%$ & $0.0\%$ \\
\cmidrule(lr){2-7}
 & \multirow{2}{*}{$5 \times 5$} & \multirow{2}{*}{25} & 2 & $+0.4\%$ & $+0.4\%$ & $+0.3\%$ \\
 &  &  & 10 & $+0.4\%$ & $+0.4\%$ & $+0.3\%$ \\
\cmidrule(lr){2-7}
 & \multirow{2}{*}{$3 \times 3$} & \multirow{2}{*}{9} & 2 & $+1.6\%$ & $+1.6\%$ & $+1.2\%$ \\
 &  &  & 10 & $+1.6\%$ & $+1.6\%$ & $+1.2\%$ \\
% rows sparse_edit: end
\bottomrule
\end{tabular}}
\label{tab:sparse_edit}
\end{table}
Tab.~\ref{tab:sparse_edit} repeats the experiment on the edited (stretched and cropped) images of
Stanford2D3D, TartanAir and LaMAR as used by AnyCalib~\cite{tirado2025anycalib}, with the officially released AnyCalib model trained on edited
images and the pinhole model fitted with $\mathbf{c}$ and $a$ estimated. Even for edited images using a sparse output leads to negligible decrease in accuracy.

\begin{table}[t]
\caption{\textbf{Fitted camera model on ScanNet++.} Mean and 95-th percentile angular error ($^\circ$) $\downarrow$ with each camera
model fitted to the dense field or to $5 \times 5$ points, with $\mathbf{c}$, $a$ free or fixed. We use the official AnyCalib$_\text{gen}$ model~\cite{tirado2025anycalib}.}
\centering
\setlength{\tabcolsep}{4pt}
\resizebox{\columnwidth}{!}{%
\begin{tabular}{l l l ccccc}
\toprule
\multicolumn{8}{c}{Mean Error} \\
\midrule
Network & Points & $\mathbf{c}$, $a$ & KB-4 & KB-2 & KB-1 & UCM & Division \\
\midrule
AnyCalib$_\text{gen}$ & Dense & free & 2.139 & 2.138 & 2.140 & 2.137 & 2.169 \\
 & Dense & fixed & 2.134 & 2.134 & 2.135 & 2.131 & 2.165 \\
 & $5 \times 5$ & free & 2.110 & 2.119 & 2.113 & 2.109 & 2.154 \\
 & $5 \times 5$ & fixed & 2.113 & 2.110 & 2.101 & 2.097 & 2.142 \\
\midrule
AnyCalib$_\text{ours}$ & Dense & free & 1.243 & 1.243 & 1.238 & 1.234 & 1.268 \\
 & Dense & fixed & 1.223 & 1.222 & 1.217 & 1.213 & 1.247 \\
 & $5 \times 5$ & free & 1.276 & 1.276 & 1.269 & 1.264 & 1.371 \\
 & $5 \times 5$ & fixed & 1.235 & 1.233 & 1.226 & 1.220 & 1.326 \\
\midrule
Ours & $5 \times 5$ & free & 1.507 & 1.395 & 1.332 & 1.334 & 1.369 \\
 & $5 \times 5$ & fixed & 1.488 & 1.316 & 1.233 & 1.193 & 1.273 \\
\bottomrule
\multicolumn{8}{c}{95th Percentile Error} \\
\midrule
Network & Points & $\mathbf{c}$, $a$ & KB-4 & KB-2 & KB-1 & UCM & Division \\
\midrule
 AnyCalib$_\text{gen}$ & Dense & free & 4.527 & 4.528 & 4.537 & 4.542 & 4.548 \\
 & Dense & fixed & 4.529 & 4.532 & 4.539 & 4.543 & 4.551 \\
 & $5 \times 5$ & free & 4.514 & 4.526 & 4.508 & 4.504 & 4.550 \\
 & $5 \times 5$ & fixed & 4.529 & 4.519 & 4.511 & 4.504 & 4.540 \\
\midrule
AnyCalib$_\text{ours}$ & Dense & free & 2.862 & 2.862 & 2.868 & 2.861 & 2.869 \\
 & Dense & fixed & 2.860 & 2.859 & 2.858 & 2.854 & 2.861 \\
 & $5 \times 5$ & free & 2.878 & 2.877 & 2.869 & 2.858 & 2.908 \\
 & $5 \times 5$ & fixed & 2.857 & 2.858 & 2.863 & 2.852 & 2.890 \\
\midrule
Ours & $5 \times 5$ & free & 3.050 & 2.765 & 2.779 & 2.780 & 2.795 \\
 & $5 \times 5$ & fixed & 3.138 & 2.740 & 2.732 & 2.723 & 2.755 \\
\bottomrule
\end{tabular}}
\label{tab:sparse_snpp_models}
\end{table}
Tab.~\ref{tab:sparse_snpp_models} fits the networks' outputs on ScanNet++ with camera models other than the KB-4 that is used for its ground truth.
For AnyCalib$_\text{gen}$ and AnyCalib$_\text{ours}$, KB-1 to KB-4 and UCM are within $0.02^\circ$ of each other in every row, and only the division model
is less accurate. 
This suggests that the errors in the intermediate representation $\vec{\theta}$ and the rays $\vec{p}$ provided by AnyCalib models are high enough that after fitting, all models provide roughly the same camera models despite using different model formulations and numbers of free parameters.
Our $5 \times 5$ output favors the models with fewer parameters: KB-4 is the least accurate and UCM the most, ahead of AnyCalib$_\text{ours}$'s dense fit. This is evidenced by the significant gap (0.3$^\circ$ difference in terms of the 95-th percentile and mean error). 
These results suggest that for downstream applications fitting models with fewer parameters may be beneficial when using our network while providing the same accuracy in terms of modeling the distortion.

% For every model, the $5 \times 5$ points are slightly more accurate than the dense field. 

\section{Network Details}
\label{sec:suppl_net}
This section completes the description of the sparse head and the auxiliary dense branch of Sec.~\ref{sec:method}.

\paragraph{Positional encoding}
We describe the position of each grid point and of each patch center by its coordinates normalized to the image, $\vec{s} = (2u/W - 1, 2v/H - 1) \in [-1, 1]^2$.
We encode it as $\gamma(\vec{s}) \in \mathbb{R}^{4F}$ consisting of $\sin(2^i \pi s)$ and $\cos(2^i \pi s)$, $i = 0, \dots, F-1$, of both coordinates, with $F = 16$.
The patch tokens receive their positions again at the output of the backbone, $\vec{m}_n = \vec{t}_n + \mathbf{W}_m \gamma(\vec{s}_n)$, and together with $\vec{t}_\text{cls}$ they form the memory $\mathcal{M}$ of $N + 1$ tokens.
Each grid point has a query $\vec{q}_j = \vec{e}_j + \mathbf{W}_q \gamma(\vec{s}_j)$ with a learned embedding $\vec{e}_j \in \mathbb{R}^D$.

\paragraph{Decoder}
The $K^2$ queries pass through two transformer decoder layers~\cite{vaswani2017attention} of width $D$ with 8 heads.
Each layer consists of self-attention among the queries, cross-attention from the queries to $\mathcal{M}$ and an MLP with a hidden layer of width $4D$ and GELU activation, each preceded by a LayerNorm and followed by a residual connection.
After a final LayerNorm a linear layer shared by all queries maps each query to two values $\vec\omega_j =\mathbf{W}_o \vec{q}_j + \vec{b}_j$, where $\vec{b}_j \in \mathbb{R}^2$ is a learned bias of the $j$-th point.
We initialize $\mathbf{W}_o$ from $\mathcal{N}(0, 10^{-8})$ and set $\vec{b}_j$ to $\vec\omega$ of the ray $\vec{p} \propto (s_{j,x} \tan 30^\circ, s_{j,y} \tan 30^\circ, 1)$, \ie at the start of the training the network predicts the field of a pinhole camera with a $60^\circ$ field of view and the principal point in the center of a square image.

\paragraph{Dense branch}
The light DPT decoder~\cite{ranftl2021dpt} of the auxiliary branch takes the tokens after blocks 5, 12 and 18 and after the final block 24 of ViT-L (blocks 3, 6, 9 and 12 of ViT-B and ViT-S, which both have 12 blocks), each normalized by the backbone's final LayerNorm, and fuses them into a feature map at $1/7$ of the input resolution.
Its head predicts $\vec\theta$ at this resolution and upsamples it to every input pixel by convex upsampling~\cite{teed2020raft} with factor 7.
The two branches share only the backbone: the dense branch reads the tokens before the positional encoding above and the sparse head does not read the dense field.
The dense branch is initialized randomly, its width is 256 for ViT-L, 128 for ViT-B and 64 for ViT-S.

\begin{table*}[t]
\caption{\textbf{Pinhole benchmarks at ViT-L.} Median hFoV error ($^\circ$) and its mAA at a $5^\circ$ threshold (\%) per benchmark, as mean and standard deviation over seeds where there are several. Total time per image, averaged over the benchmarks on A100 GPU. Gray: the benchmark's dataset is in the model's training data. Bold: best mean among the other cells.}
\centering
\resizebox{\textwidth}{!}{%
\begin{tabular}{l cc cc cc cc cc cc c}
\toprule
 & \multicolumn{2}{c}{Stanford2D3D} & \multicolumn{2}{c}{TartanAir} & \multicolumn{2}{c}{MegaDepth} & \multicolumn{2}{c}{LaMAR} & \multicolumn{2}{c}{InFlux} & \multicolumn{2}{c}{InFlux++} & \\
\cmidrule(lr){2-3} \cmidrule(lr){4-5} \cmidrule(lr){6-7} \cmidrule(lr){8-9} \cmidrule(lr){10-11} \cmidrule(lr){12-13}
Method & Median ($^\circ$) $\downarrow$ & mAA ($5^\circ$) $\uparrow$ & Median ($^\circ$) $\downarrow$ & mAA ($5^\circ$) $\uparrow$ & Median ($^\circ$) $\downarrow$ & mAA ($5^\circ$) $\uparrow$ & Median ($^\circ$) $\downarrow$ & mAA ($5^\circ$) $\uparrow$ & Median ($^\circ$) $\downarrow$ & mAA ($5^\circ$) $\uparrow$ & Median ($^\circ$) $\downarrow$ & mAA ($5^\circ$) $\uparrow$ & Time (ms) $\downarrow$ \\
\midrule
GeoCalib$_\text{pinhole}$ & 3.56 & 37.2 & 5.12 & 29.6 & 4.91 & 30.0 & 2.61 & 45.8 & 21.15 & 4.2 & 15.76 & 9.7 & 85.0 \\
MoGe & 4.98 & 30.3 & \textcolor{gray}{1.69} & \textcolor{gray}{65.5} & \textcolor{gray}{2.31} & \textcolor{gray}{50.9} & 1.69 & 60.6 & 15.76 & 6.7 & 15.24 & 8.4 & 83.3 \\
MoGe-2 & 4.64 & 29.3 & \textcolor{gray}{6.69} & \textcolor{gray}{9.5} & \textcolor{gray}{3.32} & \textcolor{gray}{39.1} & 1.98 & 56.9 & 26.20 & 4.6 & 21.23 & 8.4 & 58.5 \\
MoGe-3 & 6.12 & 18.4 & \textcolor{gray}{7.63} & \textcolor{gray}{5.2} & \textcolor{gray}{3.50} & \textcolor{gray}{37.8} & 2.23 & 54.1 & 26.96 & 5.1 & 21.03 & 8.0 & 340.0 \\
SurGe & 4.68 & 30.9 & \textcolor{gray}{3.21} & \textcolor{gray}{36.9} & \textcolor{gray}{2.78} & \textcolor{gray}{45.0} & 1.40 & 65.8 & 18.67 & 7.4 & 15.82 & 10.8 & 215.2 \\
UniDepth & 16.72 & 7.7 & 35.10 & 1.5 & 11.10 & 14.9 & 2.89 & 32.4 & 32.11 & 1.8 & 22.89 & 4.9 & 81.0 \\
UniDepthV2 & 7.11 & 19.2 & \textcolor{gray}{3.00} & \textcolor{gray}{44.5} & \textcolor{gray}{2.45} & \textcolor{gray}{49.6} & 2.04 & 57.1 & 14.91 & 4.3 & 19.64 & 6.1 & 47.0 \\
Depth Anything 3 & 5.35 & 31.2 & \textcolor{gray}{0.24} & \textcolor{gray}{99.1} & \textcolor{gray}{1.11} & \textcolor{gray}{72.2} & \textbf{0.60} & \textbf{82.0} & 17.95 & 5.9 & 11.07 & 15.3 & 92.3 \\
MapAnything & 5.51 & 25.0 & \textcolor{gray}{3.08} & \textcolor{gray}{40.6} & \textcolor{gray}{1.31} & \textcolor{gray}{68.1} & 2.72 & 48.6 & 19.89 & 5.5 & 12.48 & 10.9 & 64.3 \\
VGGT-$\Omega$ & 4.50 & 33.2 & \textcolor{gray}{3.32} & \textcolor{gray}{34.4} & \textcolor{gray}{1.21} & \textcolor{gray}{71.4} & 2.25 & 52.5 & 8.33 & 13.7 & 7.46 & 19.3 & 57.8 \\
AnyCalib$_\text{gen}$ & 3.24 & 40.4 & 4.09 & 33.4 & 3.84 & 34.5 & 2.49 & 48.3 & 12.02 & 10.1 & 9.89 & 15.2 & 35.9 \\
AnyCalib$_\text{pinhole}$ & 2.88 & 43.3 & 3.93 & 34.7 & 3.36 & 39.3 & 1.92 & 57.1 & 12.59 & 7.7 & 9.60 & 13.2 & 34.4 \\
AnyCalib, our data & \textbf{2.40} $\pm$ 0.04 & \textbf{48.9} $\pm$ 0.5 & \textbf{3.22} $\pm$ 0.22 & \textbf{40.6} $\pm$ 1.3 & \textbf{1.62} $\pm$ 0.04 & \textbf{62.1} $\pm$ 0.8 & 1.60 $\pm$ 0.06 & 62.5 $\pm$ 0.8 & 6.39 $\pm$ 0.58 & 21.1 $\pm$ 2.9 & \textbf{6.01} $\pm$ 0.15 & \textbf{24.9} $\pm$ 1.0 & 14.1 \\
Ours, $5 \times 5$ $\vec\omega$ & 2.55 $\pm$ 0.04 & 47.1 $\pm$ 0.2 & 3.26 $\pm$ 0.16 & 40.3 $\pm$ 1.0 & 1.64 $\pm$ 0.03 & 61.8 $\pm$ 0.9 & 1.61 $\pm$ 0.08 & 62.4 $\pm$ 0.9 & \textbf{6.19} $\pm$ 0.32 & \textbf{21.2} $\pm$ 0.8 & 6.32 $\pm$ 0.17 & 24.0 $\pm$ 0.8 & \textbf{9.7} \\
\bottomrule
\end{tabular}}
\label{tab:pinhole}
\end{table*}

\section{Datasets}
\label{sec:suppl_data}
This section gives the details of the evaluation and training data of Sec.~\ref{sec:datasets}.

\paragraph{ScanNet++} Seven of the images used by~\cite{tirado2025anycalib} are no longer part of the official release of ScanNet++~\cite{yeshwanth2023scannetpp}, so we score every method on the remaining 1993 images.

\paragraph{InFlux and InFlux++ Real} Since only the validation splits of InFlux~\cite{liang2025influx} and InFlux++ Real~\cite{liang2026influxpp} have public ground truth, we use these and keep them as two separate benchmarks.
We take 30 evenly spaced frames per video at half of the original resolution, $1712 \times 1101$ pixels.
This gives 1860 images from the 62 InFlux videos that have ground truth and 1484 images from the 50 InFlux++ videos, one of which has only 14 frames.
The ground truth is the Brown-Conrady model. 
The model includes very small tangential distortion. 
Since none of the evaluated models, including ours, estimate tangential distortion we as the reference camera parameters we keep only the two radial coefficients with focal length, aspect ratio and the principal point and drop the tangential coefficients.
In terms of mean angular error the choice to exclude the tangential coefficients during evaluation would change the reported values by at most $0.01^\circ$ which is negligible.
% Both sets are telephoto, with the median half-diagonal field of view of $11.9^\circ$ for InFlux and $24.1^\circ$ for InFlux++.
% Their principal points differ: it is centered in InFlux and off-center by 2\% of the half-diagonal in InFlux++ (median).

\paragraph{Cathedral} The Cathedral scene of~\cite{kocur2026minimal} was calibrated in RealityCapture,\footnote{\url{https://www.realityscan.com}} and its images come from GoPro and Samsung cameras and from Flickr.\footnote{\url{https://www.flickr.com}}
Most of its images are video frames, so we group them into the 42 distinct cameras and take up to 50 evenly spaced images per camera, 1 253 images in total out of the 2 734 images in the original dataset.
The images are downscaled by exact integer factors to a long side of about 1000 pixels.
% The principal point is off-center by a median of 0.9\% of the half-diagonal and up to 28\%.

\begin{table}[t]
\caption{\textbf{CommonsPano licenses.} Number of panoramas per license and split.}
\centering
\setlength{\tabcolsep}{4pt}
\begin{tabular}{l rrr}
\toprule
License & Train & Val & Total \\
\midrule
Public domain & 60 & 3 & 63 \\
CC0 & 1815 & 28 & 1843 \\
\midrule
CC BY 2.0 & 275 & 10 & 285 \\
CC BY 2.5 & 1 & 0 & 1 \\
CC BY 3.0 & 376 & 0 & 376 \\
CC BY 4.0 & 1857 & 260 & 2117 \\
\midrule
CC BY-SA 2.0 & 79 & 7 & 86 \\
CC BY-SA 3.0 & 170 & 19 & 189 \\
CC BY-SA 3.0 AU & 0 & 38 & 38 \\
CC BY-SA 3.0 DE & 20 & 0 & 20 \\
CC BY-SA 4.0 & 4683 & 190 & 4873 \\
\midrule
Total & 9336 & 555 & 9891 \\
\bottomrule
\end{tabular}
\label{tab:commons_licenses}
\end{table}

\begin{table}[t]
\caption{\textbf{CommonsPano panorama sizes.} Number of panoramas per width and the smallest vertical field of view of
a view of height $h$ pixels that does not sample the panorama beyond its own resolution.}
\centering
\setlength{\tabcolsep}{4pt}
\begin{tabular}{l r cc}
\toprule
 & & \multicolumn{2}{c}{Min.\ vFoV ($^\circ$)} \\
\cmidrule(lr){3-4}
Width (px) & Panoramas & $h = 640$ & $h = 320$ \\
\midrule
4096 & 167 & 56.2 & 28.1 \\
4097--5999 & 4656 & 38.4--55.0 & 19.2--27.5 \\
6000--7999 & 1695 & 28.9--38.4 & 14.4--19.2 \\
8000--8192 & 3373 & 28.1--28.8 & 14.1--14.4 \\
\bottomrule
\end{tabular}
\label{tab:commons_sizes}
\end{table}

\paragraph{CommonsPano} We collected the 360$^\circ$ panoramas from Wikimedia Commons\footnote{\url{https://commons.wikimedia.org/}} in two rounds, the first under public domain, CC0 and CC BY licenses and the second adding CC BY-SA.
We keep panoramas at least 4096 pixels wide, downscale wider ones to 8192 pixels, reject partial spheres, and limit the number of panoramas to 500 per uploader.
The train and validation splits are separated by uploader.
We removed the 465 panoramas that duplicate another panorama of OpenPano V1~\cite{veicht2024geocalib} or V2~\cite{tirado2025anycalib}, as Commons mirrors some of the sources of OpenPano.
Tab.~\ref{tab:commons_licenses} lists the licenses of the remaining 9891 panoramas.
All are equirectangular, with a 2:1 aspect ratio up to 1\% in four of them, and range from $4096 \times 2048$ to $8192 \times 4096$ pixels, with a median of $6080 \times 3040$ (Tab.~\ref{tab:commons_sizes}).
The height $H$ of a panorama spans $180^\circ$, so a view of height $h$ samples it no more densely than its own resolution when its vertical field of view is at least $180^\circ \cdot h / H$, the scale used by the view generator of~\cite{tirado2025anycalib}.
For the rendered $640 \times 640$ views this is $37.9^\circ$ for the median panorama and between $28.1^\circ$ and $56.2^\circ$ over all of them.
At the network's input of $322 \times 322$ every panorama can provide views with the smallest FoV of $28.1^\circ$ without upsampling, with 68\% of the panoramas having high enough resolution to sample $20^\circ$ FoV views.
During training we use sampled views using the OP$_\text{g}$ recipe from AnyCalib~\cite{tirado2025anycalib}.
In very few instances, the sampler produces physically implausible views such as one pixel of the panorama projected to two or more parts of the same image due to high distortion coefficients. 
We drop such views from the training set.

\paragraph{MegaScenes} We use the reconstructions of MegaScenes~\cite{tung2024megascenes} with at least 20 photos.
Each photo is center-cropped to the closest of the aspect ratios 1:1, 4:3 and 3:2, which keeps the principal point centered, and resized to a short side of 600 pixels.
We drop photos whose crop has a short side below 480 pixels, whose distortion profile folds before the image corner, and failed self-calibrations, \ie cameras with a half-diagonal field of view below $1^\circ$ or a radial shift over 50\% at the image corner.
To keep the MegaDepth benchmark held out, we exclude the scenes that show its eight landmarks. We also exclude scenes that show the landmarks from the Cathedral and Rotunda datasets~\cite{kocur2026minimal}.

\paragraph{Validation splits} Each source has its own validation split with separate panoramas or scenes: 2872 views from the 202 validation panoramas of OpenPano, 8851 views from 555 Commons panoramas and 10 779 views from 160 MegaScenes scenes, about 1\% of the scenes with views.

\section{Direct Evaluation}
\label{sec:suppl_eval}

\subsection{Pinhole Benchmarks}
\label{sec:suppl_pinhole}
Tab.~\ref{tab:pinhole} reports the four pinhole benchmarks of AnyCalib~\cite{tirado2025anycalib} alone: Stanford2D3D, TartanAir, MegaDepth and LaMAR, all fitted with the pinhole model. It adds InFlux~\cite{liang2025influx} and InFlux++~\cite{liang2026influxpp}: their ground truth has two radial distortion coefficients, but every method is fitted with the pinhole model there too, so that the calibration networks are compared with the pinhole intrinsics of the foundation models on equal terms. The metrics, the hardware and the seeds are those of Sec.~\ref{sec:eval_results}. AnyCalib trained on our data uses our GPU implementation of its fit. Many of the evaluated methods were trained on portions of the same datasets and data leakage is not easily verifiable. Moreover, even if individual images are different in the test and training sets the cameras that have taken them might be the same. We gray out those datasets.

The results show that compared to MDEs~\cite{wang2025moge, wang2025moge2,kong2026moge3,knaebel2026surge,piccinelli2024unidepth,piccinelli2025unidepthv2} and feed-forward multi-view 3D reconstruction methods~\cite{wang2025pi3,wang2026vggtomega,keetha2026mapanything,lin2025da3} ours and AnyCalib$_\text{ours}$ perform much better (with the exception of DepthAnything v3 and SurGe on LaMAR) on the benchmarks while being significantly faster.

\begin{table}[t]
\caption{\textbf{Edited Images} Median relative errors of the focal length $e_f$ and the principal point $e_c$, and median angular error AE ($^\circ$), of the pinhole model with four degrees of freedom.}
\centering
\setlength{\tabcolsep}{3pt}
\resizebox{\linewidth}{!}{%
\begin{tabular}{l ccc ccc ccc}
\toprule
 & \multicolumn{3}{c}{Stanford2D3D} & \multicolumn{3}{c}{TartanAir} & \multicolumn{3}{c}{LaMAR} \\
\cmidrule(lr){2-4} \cmidrule(lr){5-7} \cmidrule(lr){8-10}
Method & $e_f \downarrow$ & $e_c \downarrow$ & AE $\downarrow$ & $e_f \downarrow$ & $e_c \downarrow$ & AE $\downarrow$ & $e_f \downarrow$ & $e_c \downarrow$ & AE $\downarrow$ \\
\midrule
AnyCalib$_\text{edit}$ & \textbf{0.15} & \textbf{0.26} & \textbf{7.44} & \textbf{0.19} & 0.28 & 9.86 & 0.22 & 0.41 & 9.46 \\
Ours$_\text{edit}$ & 0.17 & 0.29 & 7.96 & \textbf{0.19} & \textbf{0.24} & \textbf{8.96} & \textbf{0.18} & \textbf{0.31} & \textbf{7.40} \\
\bottomrule
\end{tabular}}
\label{tab:edit}
\end{table}

\subsection{Edited Images}
\label{sec:suppl_edit}
Tab.~\ref{tab:edit} evaluates our network on the edited (stretched and cropped) images of Stanford2D3D, TartanAir and LaMAR from~\cite{tirado2025anycalib}, on which the principal point is not at the image center and the pixels are not square. We train our network with AnyCalib's edits: with probability $0.5$ a training view is cropped by up to half of its height and width, each from one side, and stretched to the training shape. The fit estimates $\mathbf{c}$ and $a$. We train it on OP$_\text{p}$ with the step-decay schedule of AnyCalib$_\text{edit}$~\cite{tirado2025anycalib}, which was trained on the same data. Following~\cite{tirado2025anycalib}, we evaluate median $e_f = |f_\text{GT} - f| / f_\text{GT}$ and $e_c = 2 |\mathbf{c}_\text{GT} - \mathbf{c}| / \text{max}(W, H)$. Our sparse output learns to estimate the full pinhole camera from edited images: it is more accurate than AnyCalib$_\text{edit}$ on TartanAir and LaMAR, with an equal $e_f$ on TartanAir, and less accurate on Stanford2D3D.
This shows that our sparse representation is sufficient to represent pinhole camera models with $\vec{c}$ off-center and $a \neq 1$ with similar or better accuracy as AnyCalib.

\begin{table*}[t]
\caption{\textbf{Representation Ablation, all metrics} Tab.~\ref{tab:ablations} of the main paper with the reprojection and vFoV errors. Mean and standard deviation over seeds of the median error and the mAA (\%), each averaged over the ten benchmarks. $^{*}$~significant against the first row with $p < 0.05$.}
\newcommand{\mk}[1]{\makebox[0.55em][l]{$^{#1}$}}
\centering
\resizebox{\textwidth}{!}{%
\begin{tabular}{lc cc cc cc cc}
\toprule
 & & \multicolumn{2}{c}{AE ($^\circ$)} & \multicolumn{2}{c}{RE (px)} & \multicolumn{2}{c}{vFoV ($^\circ$)} & \multicolumn{2}{c}{hFoV ($^\circ$)} \\
\cmidrule(lr){3-4} \cmidrule(lr){5-6} \cmidrule(lr){7-8} \cmidrule(lr){9-10}
Variant & Aux & Median $\downarrow$ & mAA ($5^\circ$) $\uparrow$ & Median $\downarrow$ & mAA (50\,px) $\uparrow$ & Median $\downarrow$ & mAA ($5^\circ$) $\uparrow$ & Median $\downarrow$ & mAA ($5^\circ$) $\uparrow$ \\
\midrule
$5 \times 5$ $\vec\omega$ & $\vec\theta$ & 1.165 $\pm$ 0.016\mk{} & 73.7 $\pm$ 0.3\mk{} & 33.1 $\pm$ 0.4\mk{} & 49.7 $\pm$ 0.2\mk{} & 2.44 $\pm$ 0.04\mk{} & 51.9 $\pm$ 0.3\mk{} & 3.01 $\pm$ 0.04\mk{} & 47.5 $\pm$ 0.3\mk{} \\
$5 \times 5$ $\vec\omega$, free $\vec{c}, a$ & $\vec\theta$ & 1.173 $\pm$ 0.015\mk{} & 73.6 $\pm$ 0.2\mk{} & 33.3 $\pm$ 0.4\mk{} & 49.5 $\pm$ 0.2\mk{} & 2.44 $\pm$ 0.04\mk{} & 51.9 $\pm$ 0.3\mk{} & 3.01 $\pm$ 0.04\mk{} & 47.4 $\pm$ 0.3\mk{} \\
$5 \times 5$ $\vec\omega$ & \xmark & 1.204 $\pm$ 0.021\mk{*} & 72.7 $\pm$ 0.5\mk{*} & 33.6 $\pm$ 0.5\mk{} & 48.6 $\pm$ 0.4\mk{*} & 2.54 $\pm$ 0.04\mk{*} & 50.4 $\pm$ 0.5\mk{*} & 3.08 $\pm$ 0.05\mk{*} & 46.3 $\pm$ 0.3\mk{*} \\
$5 \times 5$ $\vec\omega$ & $\vec\omega$ & 1.183 $\pm$ 0.008\mk{} & 73.5 $\pm$ 0.1\mk{} & 33.4 $\pm$ 0.6\mk{} & 49.6 $\pm$ 0.2\mk{} & 2.48 $\pm$ 0.02\mk{} & 51.6 $\pm$ 0.2\mk{} & 3.05 $\pm$ 0.03\mk{} & 47.3 $\pm$ 0.2\mk{} \\
$7 \times 7$ $\vec\omega$ & $\vec\theta$ & 1.186 $\pm$ 0.012\mk{} & 73.3 $\pm$ 0.3\mk{*} & 33.8 $\pm$ 0.5\mk{} & 49.5 $\pm$ 0.1\mk{} & 2.49 $\pm$ 0.03\mk{} & 51.4 $\pm$ 0.3\mk{} & 3.08 $\pm$ 0.04\mk{*} & 47.1 $\pm$ 0.2\mk{} \\
$5 \times 5$ $\vec\theta$ & $\vec\theta$ & 1.174 $\pm$ 0.014\mk{} & 73.5 $\pm$ 0.4\mk{} & 33.3 $\pm$ 0.6\mk{} & 49.7 $\pm$ 0.3\mk{} & 2.46 $\pm$ 0.03\mk{} & 51.6 $\pm$ 0.4\mk{} & 3.06 $\pm$ 0.06\mk{} & 46.9 $\pm$ 0.4\mk{*} \\
$5 \times 5$ $\vec\theta$ & \xmark & 1.194 $\pm$ 0.024\mk{} & 73.1 $\pm$ 0.2\mk{*} & 33.7 $\pm$ 1.2\mk{} & 49.0 $\pm$ 0.2\mk{*} & 2.53 $\pm$ 0.05\mk{*} & 50.7 $\pm$ 0.2\mk{*} & 3.06 $\pm$ 0.08\mk{} & 46.5 $\pm$ 0.2\mk{*} \\
Radial profile & $\vec\theta$ & 1.184 $\pm$ 0.015\mk{} & 73.4 $\pm$ 0.2\mk{} & 33.5 $\pm$ 0.4\mk{} & 49.6 $\pm$ 0.2\mk{} & 2.49 $\pm$ 0.03\mk{} & 51.3 $\pm$ 0.3\mk{*} & 3.06 $\pm$ 0.05\mk{} & 47.0 $\pm$ 0.3\mk{*} \\
Direct regression & $\vec\theta$ & 1.204 $\pm$ 0.022\mk{*} & 73.3 $\pm$ 0.5\mk{} & 34.1 $\pm$ 0.8\mk{*} & 49.5 $\pm$ 0.3\mk{} & 2.53 $\pm$ 0.04\mk{*} & 51.1 $\pm$ 0.6\mk{*} & 3.12 $\pm$ 0.07\mk{*} & 46.8 $\pm$ 0.5\mk{*} \\
\bottomrule
\end{tabular}}
\label{tab:ablations_full}
\end{table*}

\subsection{Output Representation Variants}
\label{sec:suppl_repr}
Tab.~\ref{tab:ablations} compares four output representations. All of them use the same backbone, decoder layers and training recipe, and differ in what the queries predict and how the intrinsics are obtained from it.

\paragraph{Ray map $\vec\omega$} Our network: $K^2$ queries predict the two signed angles of \eqref{eq:omega} at the $K \times K$ grid points, trained with the L1 loss \eqref{eq:loss_sparse}. The rays follow from \eqref{eq:omega_inv} and the intrinsics are fitted to the pixel-ray pairs. We evaluate $K = 5$ and $K = 7$.

\paragraph{Log map $\vec\theta$} The same network and points, predicting AnyCalib's $\vec\theta$ of Sec.~\ref{sec:fov_field} with the same loss. Its rays follow from the inverse mapping of Sec.~\ref{sec:fov_field} and are fitted in the same way, so only the parameterization of the ray differs.

\paragraph{Radial profile} Since the code and weights for VaCR~\cite{dal2024revisiting} are not public we implement our own variant with a 1D radial profile. The network predicts the polar angle $\theta$ of the ray as a function of the distance $\rho$ from the image center alone, which assumes that the principal point is at the center and the pixels are square; the positional encoding therefore encodes only $\rho$. It predicts $\theta_k$ at 12 radii $\rho_k = \rho_\text{max} \sqrt{k / 12}$, $k = 1, \dots, 12$, up to the half-diagonal $\rho_\text{max}$, so that each radius represents about the same image area.
One query per radius and one global query are decoded into logits $z_k$ and $s$, and
\begin{equation}
\Theta = \pi \operatorname{sigmoid}(s), \qquad \theta_k = \Theta \sum_{i \le k} \operatorname{softmax}(\vec{z})_i,
\end{equation}
which makes the profile monotone with $\theta_{12} = \Theta$ at the image corner. The training loss is the L1 error of $\theta_k$ weighted by the share of the image pixels closest to each radius. The intrinsics $\boldsymbol\eta$ are fitted by minimizing $\sum_k (\theta_k - \psi(\rho_k; \boldsymbol\eta))^2$ with the principal point held at the center, where $\psi$ is the polar angle given by the camera model at radius $\rho$, using the linear solution and two iterations.

\paragraph{Direct regression} Two queries predict the focal length and the distortion of the one-parameter division model~\cite{fitzgibbon2001division} directly: $\log(f / \rho_\text{max})$ and $\kappa = 1 - \operatorname{softplus}(x)$, the division coefficient measured in units of the half-diagonal, for which every output is a valid camera. The network is trained on the 12 angles of the radial profile rendered from its output:
\begin{equation}
\theta(\rho) = \operatorname{atan2}\!\left(\frac{\rho}{f}, 1 + \kappa \frac{\rho^2}{\rho_\text{max}^2}\right),
\end{equation}
with the same loss. The network needs no fit as it outputs this camera on every benchmark, whichever model the other variants fit there.

\paragraph{Significance Test} 
Table~\ref{tab:ablations} uses an exact two-sided permutation test on the difference of means. 
Each cell is tested against the table's first row.

\paragraph{Extended Results}
Tab.~\ref{tab:ablations_full} repeats Tab.~\ref{tab:ablations} with all four metrics, and Tab.~\ref{tab:ablation_pinhole} and Tab.~\ref{tab:ablation_distorted} report the variants for each of the ten benchmarks.

\input{tab/ablation_pinhole}
\input{tab/ablation_distorted}

\subsection{Per-benchmark Results}
\label{sec:suppl_extended}
Tab.~\ref{tab:extended_pinhole} and Tab.~\ref{tab:extended_distorted} report the results of Sec.~\ref{sec:eval_results} for each of the ten benchmarks, with additional baselines where applicable.

\subsection{Embedded runtime}
We measure runtime on an NVIDIA Jetson Orin Nano 8 GB (JetPack 6.2.1, CUDA 12.6, TensorRT 10.3) in its 15 W power mode and one model per process. 
Each network is exported to ONNX (opset 17) at a fixed input size with about 102k pixels: 280×364 for the ViTs and 320×384 for GeoCalib. 
UniK3D is exported after its own resize, at 658×854. 
Each graph is compiled into an fp16 TensorRT engine with trtexec and timed at batch size 1. 
Two export details were needed for the fp16 engines to match the fp32 networks. 
First, the head's Fourier positional encoding is constant for a fixed input size, so we evaluate it in fp32 at export time and store it as a constant, because its arguments overflow fp16. 
Second, we export attention as explicit matrix products rather than as SDPA, which TensorRT 10.3 fuses into a kernel that is inaccurate on real images. With both, the engines reproduce the fp32 networks to within 0.01° median error on every benchmark.
Total runtime is the engine latency plus the camera fit, measured on the board. 
Our fit (AnyCalib's fit on our 25 rays, with the principal point and aspect ratio held) runs on the CPU, compiled with Numba. 
The dense $\sim$100k-ray fit of AnyCalib and UniK3D runs on the GPU through our Numba CUDA port of AnyCalib's fit, which is its fastest option on the board. 
GeoCalib uses its own GPU optimizer. 
Fit times are averaged over the six camera models.

\input{tab/extended_pinhole}
\input{tab/extended_distorted}

\clearpage
\clearpage

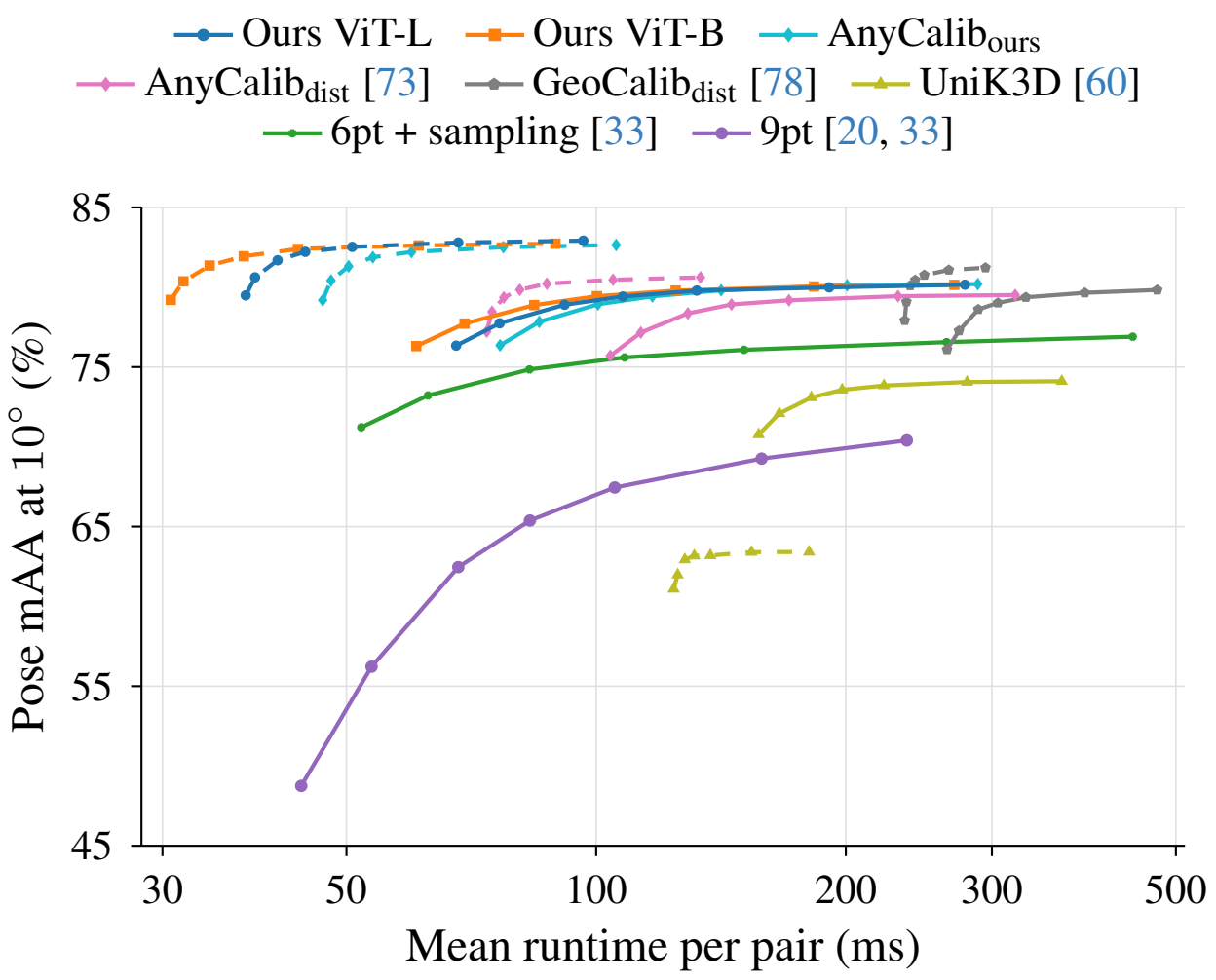
\begin{figure}[b]
\centering
\input{fig/pose_equal}
\caption{\textbf{Relative Pose of Radially Distorted Cameras with Equal Intrinsics.} Pose mAA at $10^\circ$ over 10 000 image pairs of Cathedral~\cite{kocur2026minimal} against the mean runtime per pair, one point per RANSAC budget from 10 to 1000 PoseLib~\cite{PoseLib} iterations using SuperPoint~\cite{detone2018superpoint} and LightGlue~\cite{lindenberger2023lightglue} matches.
A network adds its forward pass on both images and one joint fit on an A100 to the RANSAC runtime on the CPU.
The shared intrinsics provided by networks were combined with the 5-point solver~\cite{nister2004efficient} to obtain initial models with local optimization either only for pose (dashed lines) or also for the shared intrinsics (solid lines).
}
\label{fig:pose_equal}
\end{figure}

\begin{table}[b]
\caption{\textbf{Structure-from-Motion on EuRoC.} mAA of the relative pose error over all image pairs and the share of registered images (\%), averaged over 100 subsets of 10 keyframes from each of five EuRoC~\cite{burri2016euroc} sequences; mean $\pm$ standard deviation over 5 runs of COLMAP.}
\centering
\setlength{\tabcolsep}{4pt}
\resizebox{\columnwidth}{!}{%
\begin{tabular}{l cc cc}
\toprule
 & \multicolumn{2}{c}{Incremental~\cite{schoenberger2016sfm}} & \multicolumn{2}{c}{Global~\cite{pan2024glomap}} \\
\cmidrule(lr){2-3} \cmidrule(lr){4-5}
Prior & mAA($10^\circ$) $\uparrow$ & Reg. $\uparrow$ & mAA($10^\circ$) $\uparrow$ & Reg. $\uparrow$ \\
\midrule
None & 72.6 $\pm$ 0.6 & 95.4 $\pm$ 0.1 & 79.5 $\pm$ 0.3 & 98.6 $\pm$ 0.2 \\
GeoCalib$_\text{dist}$~\cite{veicht2024geocalib} & 75.5 $\pm$ 0.5 & 95.6 $\pm$ 0.3 & 81.2 $\pm$ 0.2 & \textbf{99.5} $\pm$ 0.0 \\
AnyCalib$_\text{dist}$~\cite{tirado2025anycalib} & 75.9 $\pm$ 0.2 & 95.3 $\pm$ 0.2 & \textbf{81.5} $\pm$ 0.1 & \textbf{99.5} $\pm$ 0.0 \\
AnyCalib$_\text{ours}$ & 76.0 $\pm$ 0.4 & 95.6 $\pm$ 0.2 & 81.4 $\pm$ 0.1 & 99.4 $\pm$ 0.1 \\
Ours & \textbf{76.3} $\pm$ 0.5 & \textbf{95.8} $\pm$ 0.1 & 81.4 $\pm$ 0.1 & \textbf{99.5} $\pm$ 0.1 \\
\bottomrule
\end{tabular}}
\label{tab:sfm_euroc}
\end{table}

\section{Downstream Applications}
\label{sec:suppl_downstream}

\subsection{Relative Pose Estimation}
\label{sec:suppl_pose_equal}
Fig.~\ref{fig:pose_different} in the main paper evaluates the calibration inside the relative camera pose estimation task on Cathedral~\cite{kocur2026minimal}, which has 2734 images from 42 cameras with ground truth poses and intrinsics from its reconstruction. We use the image pairs of~\cite{kocur2026minimal}, sampled among the pairs whose shared reconstructed points cover more than 10\% of both images, with their SuperPoint~\cite{detone2018superpoint} and LightGlue~\cite{lindenberger2023lightglue} matches. That figure uses all 10 000 pairs of the general set, no pair is discarded, and they cover 2 527 images from 40 cameras. Of these pairs, 8 320 have two different cameras and 1 680 the same camera. We run PoseLib~\cite{PoseLib} with a fixed number of RANSAC iterations, 10, 20, 50, 100, 200, 500 or 1 000, and an inlier threshold of 3 pixels. We compute mAA following~\cite{jin2021image}. The baselines are the non-minimal 10-point solver~\cite{kukelova2015efficient} and the 7-point solver with $\lambda_1$ and $\lambda_2$ sampled from $\{0, -0.6, -1.2\}$~\cite{kocur2026minimal}. In terms of runtime, each network adds two forward passes and two fits, 19\,ms per pair for our ViT-L, 12\,ms for our ViT-B, 79\,ms for AnyCalib$_\text{dist}$, 132\,ms for UniK3D and 236\,ms for GeoCalib$_\text{dist}$.

Fig.~\ref{fig:pose_equal} repeats this experiment for two views taken by the same camera. We use the second set of 10 000 pairs of~\cite{kocur2026minimal}, sampled in the same way among images sharing a calibration group in the reconstruction. All 10 000 pairs are used and all of them have the same camera, covering 2 673 images from 28 cameras. Every network fits the one-parameter division model~\cite{fitzgibbon2001division} with the principal point at the image center, as the solvers assume, and sees each image upright.

Since the two views share their intrinsics, we estimate one camera per pair from the two images together. Each network runs on both images separately, and only the fit is joint: it estimates a single $(f, \lambda)$ from the outputs of both. For our network, AnyCalib and UniK3D this is the same fit as for one image, on the rays of the two images stacked. For GeoCalib we use its own optimizer with shared intrinsics on the two perspective fields, with its default number of iterations. A pair therefore costs two forward passes and one joint fit, measured on one A100, which gives 19\,ms for our ViT-L, 12\,ms for our ViT-B, 28\,ms for AnyCalib$_\text{ours}$, 106\,ms for UniK3D and 216\,ms for GeoCalib$_\text{dist}$.

The joint camera is then used by the 5-point solver~\cite{nister2004efficient} in PoseLib~\cite{PoseLib}, as in Fig.~\ref{fig:pose_different}, and the equality is kept during the refinement: one $(f, \lambda)$ is optimized for both views. With a dashed line, local optimization refines only the pose and the final refinement also refines the shared intrinsics, of which only the pose is kept. With a solid line, local optimization and the final refinement both refine the pose and the shared intrinsics. The baselines estimate the same shared camera: the non-minimal 9-point solver~\cite{fitzgibbon2001division,kocur2026minimal} and the 6-point solver with $\lambda$ sampled from $\{0, -0.6, -1.2\}$~\cite{kocur2026minimal}.

Similarly to the results for the case of two different cameras in the main paper (Fig.~\ref{fig:pose_different}) our network provides the best accuracy while being the fastest.

\subsection{Structure from Motion}
\label{sec:suppl_sfm}
The top part of Tab.~\ref{tab:sfm} in the main paper uses the test scenes of the PhotoTourism benchmark of the Image Matching Challenge~\cite{jin2021image}, keeping the five scenes whose landmarks do not appear in our training or validation data: Lincoln Memorial Statue, Mount Rushmore, Milan Cathedral, British Museum, and Sagrada Fam\'ilia. Their images are undistorted, have no EXIF metadata and were taken with many different cameras and focal lengths. From each scene we sample 100 random subsets of 10 images, the same subsets for every prior, and reconstruct each subset with the default pipeline of COLMAP~\cite{schoenberger2016sfm}: SIFT~\cite{lowe2004distinctive} features, exhaustive matching and either the incremental mapper or the global mapper of GLOMAP~\cite{pan2024glomap}. Each image has its own pinhole camera with a single focal length. Since the input images are already undistorted we do not estimate distortion. Without a prior, COLMAP initializes the focal length to $1.2 \max(w, h)$ and the principal point to the image center. With a prior, each network calibrates each image with the pinhole model, and its focal length and principal point initialize that image's camera, which COLMAP then treats as a known prior and refines in bundle adjustment. We report the mAA of the relative pose error~\cite{jin2021image} of all image pairs in a subset, where the error of a pair is the larger of its rotation and translation direction errors and a pair with an unregistered image counts as failed, averaged over the 500 subsets.

The bottom part of Tab.~\ref{tab:sfm} in the main paper uses the Rotunda scene from \cite{kocur2026minimal}. It includes 157 photos with ground truth poses and one-parameter division cameras from its reconstruction. The images come from four cameras: two smartphone cameras with almost no distortion (69 and 26 photos) and two wide-angle action cameras with strong barrel distortion (46 and 16 photos). We sample 100 random subsets of 10 images, the same for every prior, and run the same pipeline with COLMAP's one-parameter division camera. The images of a subset are grouped into cameras as COLMAP groups them from their EXIF metadata, which gives one camera per physical camera, and this grouping is used for every prior. From the available EXIF metadata, COLMAP finds no usable focal length for the smartphone cameras and uses its default, and its focal lengths for the action cameras are 37\% and 84\% too long, since they assume a pinhole camera. Without EXIF, every camera starts from the default. With a prior, each network calibrates each image with the one-parameter division model, with the principal point held at the image center and the image upright, and one camera per group is fitted jointly to the outputs of the group's images in the subset. For GeoCalib we use the GeoCalib$_\text{dist}$ model. For AnyCalib we used AnyCalib$_\text{dist}$. The registered share is the fraction of a subset's images registered in the largest reconstruction.

Tab.~\ref{tab:sfm_euroc} uses the left camera of five EuRoC~\cite{burri2016euroc} sequences, V1\_02\_medium, V1\_03\_difficult, V2\_01\_easy, V2\_02\_medium, and V2\_03\_difficult, with $752 \times 480$ grayscale frames and ground truth poses from each sequence's ground truth trajectory and the camera's extrinsics. Each subset consists of 10 frames. To sample them, we start at a random frame of the $i$-th of 100 equal parts of the sequence. As the next frame in the subset we add the first one in the sequence that is either at least 0.2\,m from the previous one or rotated at least $10^\circ$ from the previous one. We run the same pipeline with COLMAP's one-parameter division camera, shared by all frames of a subset. 
Without a prior, it starts at a focal length of $1.2 \max(w, h)$ and no distortion. With a prior, each network calibrates each frame with the one-parameter division model, with the principal point held at the image center and the image upright, and the shared camera is fitted jointly to the outputs of the subset's 10 frames. GeoCalib and the released AnyCalib use their distorted models. With COLMAP's default one-parameter polynomial camera instead of the division camera, the reconstructions without a prior are far less accurate.

Every network prior improves both mappers, but only by few points. This is mostly due to initialization as  bundle adjustment brings all reconstructions, COLMAP's included, to within 2\% of the true focal length. 
Due to the same bundle adjustment strategy the networks are also within 0.8 points of each other.
The networks thus provide a small, but clear benefit with our network being the most efficient choice.

\subsection{3D Reconstruction}

Tab.~\ref{tab:recon} in Sec.~\ref{sec:sfm} uses four Zip-NeRF~\cite{barron2023zipnerf} scenes, Alameda, Berlin, London and NYC, with $876 \times 584$ frames and ground truth poses from the COLMAP reconstructions released with the dataset. Each subset consists of 10 consecutive frames from a random start. Each network calibrates each frame as on EuRoC. 
For each subset one camera is fitted to all images in the subset with the one-parameter division model.
UniK3D's rays are fitted with our implementation of AnyCalib's fit. Pi3X~\cite{wang2025pi3}, MapAnything~\cite{keetha2026mapanything}, DA3~\cite{lin2025da3} (DA3-GIANT-1.1) and VGGT-$\Omega$~\cite{wang2026vggtomega} (its $512$-pixel model) reconstruct each subset with their default inference. Without a prior, they get the raw frames and no intrinsics. 
With a prior, each frame is undistorted with the subset's camera to a $696 \times 396$ pinhole image at the fitted focal length with the principal point at the center.
Pi3X and MapAnything also use the estimated pinhole intrinsics at input.
DA3 accepts a camera only together with its pose, and VGGT-$\Omega$ takes images only, so these two get the undistorted images alone. 
With the ground truth camera intrinsics in place of a prior, the mAA at $5^\circ$ is 75.1 for Pi3X, 41.5 for MapAnything, 72.8 for DA3 and 67.4 for VGGT-$\Omega$.

Tab.~\ref{tab:recon_rotunda} repeats the experiment on Rotunda, with the subsets and the division cameras of Tab.~\ref{tab:sfm}: one camera model fitted jointly per camera group of each subset. Each image is undistorted with its group's camera model to a pinhole image of its own aspect ratio, at most 1024 pixels on the long side, with the principal point at the center. The smartphone photos are stored sideways with an EXIF orientation tag; every model sees each image upright, and its predicted poses are rotated back about the optical axis before scoring. A good prior helps every model, while GeoCalib's inaccurate focal length given as intrinsics makes Pi3X and MapAnything worse than no prior; the priors of ours and AnyCalib$_\text{ours}$ give the most accurate poses.

The bottom part of Tab.~\ref{tab:recon} uses Cathedral~\cite{kocur2026minimal}, all 2734 of its images with ground truth poses and one-parameter division cameras from its reconstruction. Most of them are video frames, 42 in total. The scene spans about 45 units of the reconstruction, and 10 images drawn from it at random rarely see the same part of it: in such subsets the median angle between the optical axes of two images is $78^\circ$. We therefore sample subsets of covisible images from the ground truth poses. A subset starts at a random anchor image, never used as an anchor again, and its other nine images are drawn from the images whose camera center lies within 8 units of the anchor's and whose optical axis lies within $45^\circ$ of the anchor's. To keep several cameras in every subset, one image each of two cameras other than the anchor's is drawn first and the rest at random. In the resulting 100 subsets the median angle between the optical axes of two images is $18.5^\circ$ and a subset holds images of a median of six cameras. The rest of the setup is that of Rotunda: the same subsets for every prior, one division camera fitted jointly per camera of each subset, each image undistorted with its camera to at most 1024 pixels on the long side with the principal point at the center, and the models seeing every image upright. With the ground truth camera, the mAA at $5^\circ$ is 50.6 for Pi3X, 40.6 for MapAnything, 35.9 for DA3 and 40.0 for VGGT-$\Omega$.
With Pi3X and MapAnything, which take the intrinsics, our prior gives the most accurate poses, 10.4 and 6.9 points ahead of AnyCalib$_\text{dist}$, whose focal lengths are off by a median of 7\% against 2\% for ours. DA3 and VGGT-$\Omega$ estimate the focal length themselves and get only the undistorted images, so every prior but UniK3D's gives results within one point of the others. UniK3D's rays are fitted as on Zip-NeRF; its cameras miss the barrel distortion of Cathedral's cameras, so its prior is worse than none for every model.

Similar to the results presented in the main paper, the results in Table~\ref{tab:recon_rotunda} show that undistorting images before using Feed-Forward networks significantly improves performance with our method performing the best.

\begin{table}[t]
\caption{\textbf{Feed-Forward 3D Reconstruction on Rotunda.} mAA($5^\circ$) $\uparrow$ of the relative pose error over all image pairs, averaged over 100 subsets of 10 images of Rotunda~\cite{kocur2026minimal}. $^\dagger$ Takes no intrinsics: with a prior it gets the undistorted images only. Bold: the best in each column, the GT camera aside.}
\centering
\setlength{\tabcolsep}{4pt}
\resizebox{\columnwidth}{!}{%
\begin{tabular}{l cccc}
\toprule
Prior & Pi3X~\cite{wang2025pi3} & MapAnything~\cite{keetha2026mapanything} & DA3$^\dagger$~\cite{lin2025da3} & VGGT-$\Omega^\dagger$~\cite{wang2026vggtomega} \\
\midrule
GT camera & 73.1 & 47.2 & 31.8 & 48.4 \\
\midrule
None & 33.6 & 23.0 & 22.5 & 15.3 \\
GeoCalib$_\text{dist}$~\cite{veicht2024geocalib} & 23.9 & 19.9 & 24.6 & 33.4 \\
AnyCalib$_\text{dist}$~\cite{tirado2025anycalib} & 51.4 & 33.8 & 25.0 & 34.6 \\
AnyCalib$_\text{ours}$ & 51.3 & 34.6 & 26.4 & \textbf{36.5} \\
Ours & \textbf{52.4} & \textbf{35.1} & \textbf{26.5} & 36.1 \\
\bottomrule
\end{tabular}}
\label{tab:recon_rotunda}
\end{table}

\begin{table}[t]
\caption{\textbf{Localization.} Share of the Aachen Day-Night v1.1~\cite{sattler2018benchmarking,zhang2021reference} test queries (\%) localized within each position and rotation threshold, by day and by night. Bold: the best in each column, GT intrinsics aside.}
\centering
\setlength{\tabcolsep}{4pt}
\resizebox{\columnwidth}{!}{%
\begin{tabular}{l l ccc ccc}
\toprule
 & & \multicolumn{3}{c}{Day} & \multicolumn{3}{c}{Night} \\
\cmidrule(lr){3-5} \cmidrule(lr){6-8}
Prior & Solver & 0.25m, $2^\circ$ & 0.5m, $5^\circ$ & 5m, $10^\circ$ & 0.25m, $2^\circ$ & 0.5m, $5^\circ$ & 5m, $10^\circ$ \\
\midrule
GT intrinsics & P3P & 90.3 & 96.5 & 99.2 & 75.9 & 91.1 & 99.5 \\
\midrule
None & P3.5Pf & 60.6 & 82.5 & \textbf{99.2} & \textbf{74.3} & \textbf{90.6} & \textbf{99.5} \\
None & P5Pfr & \textbf{64.7} & \textbf{85.4} & \textbf{99.2} & 70.2 & 90.1 & \textbf{99.5} \\
UniK3D~\cite{piccinelli2025unik3d} & P3P & 62.6 & 83.4 & \textbf{99.2} & 71.7 & 89.5 & \textbf{99.5} \\
GeoCalib~\cite{veicht2024geocalib} & P3P & 63.6 & 84.7 & 98.9 & 70.2 & \textbf{90.6} & 99.0 \\
AnyCalib$_\text{gen}$~\cite{tirado2025anycalib} & P3P & 63.3 & 84.2 & \textbf{99.2} & 69.1 & \textbf{90.6} & \textbf{99.5} \\
AnyCalib$_\text{ours}$ & P3P & 64.1 & 84.8 & \textbf{99.2} & 69.1 & 90.1 & 99.0 \\
Ours & P3P & 64.1 & 85.0 & \textbf{99.2} & 69.6 & 90.1 & 99.0 \\
\bottomrule
\end{tabular}}
\label{tab:loc}
\end{table}

\subsection{Localization}
\label{sec:suppl_loc}
We evaluate whether a calibration prior helps to localize query images with unknown intrinsics against a 3D map on the Aachen Day-Night v1.1 benchmark~\cite{sattler2018benchmarking,zhang2021reference}: 824 day and 191 night queries, scored by its online evaluation server. The map is triangulated from all 6 697 database images with their reference poses, using SuperPoint~\cite{detone2018superpoint} features matched by LightGlue~\cite{lindenberger2023lightglue} to the 50 database images retrieved by NetVLAD~\cite{arandjelovic2016netvlad}, as in hloc~\cite{sarlin2019coarse}. Each query's pose is estimated from its 2D-3D matches with LO-RANSAC in PoseLib~\cite{PoseLib}. Without a prior, the minimal solver also estimates the intrinsics: P3.5Pf~\cite{larsson2017making} the focal length, P5Pfr~\cite{kukelova2013real} also one radial distortion coefficient. With a prior, the query's camera is the network's estimated one-parameter Brown-Conrady~\cite{brown1971close} radial camera and the pose is estimated with P3P~\cite{persson2018lambda}; the final refinement on the inliers also refines the focal length and the distortion, and the estimation is repeated once from the refined camera. Tab.~\ref{tab:loc} shows that with any network prior P3P localizes as many queries as P5Pfr by day, more than P3.5Pf at the two tighter thresholds, while at night P3.5Pf is the most accurate. The calibration networks are within about one point of each other. UniK3D localizes about one point fewer queries than the rest by day and two more at night at the tightest threshold.
These results show that for localization the existing solvers with refinement already estimate the intrinsics well enough and networks do not bring significant improvement.

%% file: tab/sparse_full.tex
% Sparse fits of AnyCalib's field: its fit, released (torch) and compiled (Numba), and our angle fit; every benchmark (supplementary).
% Rows: radcalib's `eval/paper_sparse_table.py --paper <this repo>` (sources: radcalib docs/paper.md).
\begin{table*}[ht]
\caption{\textbf{Sparse fits of AnyCalib's field, per benchmark.} AnyCalib's fit in its
released implementation (PyTorch, GPU) and our implementation of it compiled with Numba (CPU / GPU). Mean angular error: in
degrees for the official implementation~\cite{tirado2025anycalib} on the dense field, as the change relative to it otherwise.
Iter.: Gauss-Newton iterations. Fit time per image for each fitted camera model, on the CPU / the GPU.}
\centering
\setlength{\tabcolsep}{3.5pt}
\renewcommand{\arraystretch}{1.1}
\resizebox{\linewidth}{!}{%
\begin{tabular}{llrc ccccccc cccc}
\toprule
& & & & \multicolumn{4}{c}{Pinhole $\downarrow$} & \multicolumn{3}{c}{Distorted $\downarrow$}
& \multicolumn{4}{c}{Fit time [ms] $\downarrow$, CPU / GPU} \\
\cmidrule(lr){5-8}\cmidrule(lr){9-11}\cmidrule(lr){12-15}
Fit & Points & \#pts & Iter.
& S2D3D & TartanAir & MegaDepth & LaMAR
& MD & SN++ & Mono
& Pinhole & BC-1 & KB-4 & UCM \\
\midrule
% rows sparse_full: begin
\multirow{7}{*}{AnyCalib~\cite{tirado2025anycalib}} & Dense & $\sim\!10^5$ & 10 & 2.10$^\circ$ & 2.51$^\circ$ & 2.37$^\circ$ & 1.52$^\circ$ & 2.42$^\circ$ & 2.14$^\circ$ & 2.55$^\circ$ & -- / 22.8 & -- / 31.3 & -- / 98.9 & -- / 27.7 \\
 & Dense, $\mathbf{c}$, $a$ fixed & $\sim\!10^5$ & 10 & $-0.3\%$ & $-0.3\%$ & $-0.4\%$ & $-0.4\%$ & $-0.3\%$ & $-0.2\%$ & $-0.5\%$ & -- / 22.8 & -- / 31.2 & -- / 92.5 & -- / 27.9 \\
 & $99 \times 99$ & 9801 & 10 & $0.0\%$ & $0.0\%$ & $0.0\%$ & $0.0\%$ & $0.0\%$ & $0.0\%$ & $-0.1\%$ & -- / 22.5 & -- / 30.5 & -- / 91.8 & -- / 27.3 \\
 & $25 \times 25$ & 625 & 10 & $-0.2\%$ & $-0.1\%$ & $-0.1\%$ & $+0.1\%$ & $-0.1\%$ & $0.0\%$ & $-0.5\%$ & -- / 22.1 & -- / 29.6 & -- / 89.0 & -- / 26.7 \\
 & $7 \times 7$ & 49 & 10 & $-1.3\%$ & $-0.5\%$ & $+0.3\%$ & $+0.6\%$ & $0.0\%$ & $-0.2\%$ & $-1.9\%$ & -- / 21.5 & -- / 28.9 & -- / 88.3 & -- / 25.9 \\
 & $5 \times 5$ & 25 & 10 & $-1.8\%$ & $-0.6\%$ & $+0.4\%$ & $+1.2\%$ & $+0.2\%$ & $+5.4\%$ & $-1.7\%$ & -- / 21.3 & -- / 28.9 & -- / 87.7 & -- / 25.9 \\
 & $3 \times 3$ & 9 & 10 & $-3.2\%$ & $+0.1\%$ & $+1.3\%$ & $+3.0\%$ & $-0.4\%$ & $+1746\%$ & $+3.0\%$ & -- / 21.3 & -- / 28.7 & -- / 87.6 & -- / 25.7 \\
\midrule
\multirow{12}{*}{\shortstack{Ours\\(Numba)\\$\mathbf{c}$, $a$ free}} & \multirow{2}{*}{Dense} & \multirow{2}{*}{$\sim\!10^5$} & 2 & $0.0\%$ & $0.0\%$ & $0.0\%$ & $0.0\%$ & $0.0\%$ & $0.0\%$ & $-0.1\%$ & 25.4 / 1.43 & 40.0 / 1.90 & 69.0 / 3.19 & 26.4 / 1.33 \\
 &  &  & 10 & $0.0\%$ & $0.0\%$ & $0.0\%$ & $0.0\%$ & $0.0\%$ & $0.0\%$ & $0.0\%$ & 74.0 / 2.47 & 103.8 / 3.22 & 84.5 / 7.02 & 85.4 / 2.77 \\
\cmidrule(lr){2-15}
 & \multirow{2}{*}{$99 \times 99$} & \multirow{2}{*}{9801} & 2 & $0.0\%$ & $0.0\%$ & $0.0\%$ & $0.0\%$ & $0.0\%$ & $0.0\%$ & $-0.2\%$ & 9.53 / 0.85 & 5.58 / 0.96 & 6.75 / 1.61 & 9.73 / 0.81 \\
 &  &  & 10 & $0.0\%$ & $0.0\%$ & $0.0\%$ & $0.0\%$ & $0.0\%$ & $0.0\%$ & $-0.1\%$ & 20.3 / 1.33 & 17.3 / 1.70 & 20.1 / 3.52 & 16.6 / 1.42 \\
\cmidrule(lr){2-15}
 & \multirow{2}{*}{$25 \times 25$} & \multirow{2}{*}{625} & 2 & $-0.2\%$ & $-0.1\%$ & $-0.1\%$ & $+0.1\%$ & $-0.1\%$ & $0.0\%$ & $-0.6\%$ & 0.55 / 0.69 & 0.91 / 0.76 & 1.85 / 1.06 & 0.55 / 0.68 \\
 &  &  & 10 & $-0.2\%$ & $-0.1\%$ & $-0.1\%$ & $+0.1\%$ & $-0.1\%$ & $0.0\%$ & $-0.5\%$ & 1.54 / 0.93 & 2.73 / 1.11 & 5.59 / 1.96 & 1.53 / 0.98 \\
\cmidrule(lr){2-15}
 & \multirow{2}{*}{$7 \times 7$} & \multirow{2}{*}{49} & 2 & $-1.3\%$ & $-0.5\%$ & $+0.3\%$ & $+0.6\%$ & $0.0\%$ & $-0.2\%$ & $-2.0\%$ & 0.15 / 0.65 & 0.18 / 0.69 & 0.26 / 0.86 & 0.15 / 0.64 \\
 &  &  & 10 & $-1.3\%$ & $-0.5\%$ & $+0.3\%$ & $+0.6\%$ & $0.0\%$ & $-0.2\%$ & $-1.9\%$ & 0.23 / 0.82 & 0.33 / 0.94 & 0.55 / 1.46 & 0.23 / 0.91 \\
\cmidrule(lr){2-15}
 & \multirow{2}{*}{$5 \times 5$} & \multirow{2}{*}{25} & 2 & $-1.8\%$ & $-0.6\%$ & $+0.4\%$ & $+1.2\%$ & $-0.1\%$ & $-1.3\%$ & $-1.8\%$ & 0.14 / 0.71 & 0.16 / 0.85 & 0.20 / 1.01 & 0.13 / 0.71 \\
 &  &  & 10 & $-1.8\%$ & $-0.6\%$ & $+0.4\%$ & $+1.2\%$ & $+0.2\%$ & $-1.2\%$ & $-1.7\%$ & 0.17 / 0.82 & 0.22 / 0.94 & 0.34 / 1.46 & 0.17 / 0.87 \\
\cmidrule(lr){2-15}
 & \multirow{2}{*}{$3 \times 3$} & \multirow{2}{*}{9} & 2 & $-3.2\%$ & $+0.1\%$ & $+1.3\%$ & $+3.0\%$ & $-0.7\%$ & $+1695\%$ & $+2.7\%$ & 0.12 / 0.66 & 0.14 / 0.73 & 0.16 / 0.91 & 0.12 / 0.65 \\
 &  &  & 10 & $-3.2\%$ & $+0.1\%$ & $+1.3\%$ & $+3.0\%$ & $-0.4\%$ & $+1723\%$ & $+3.0\%$ & 0.13 / 0.82 & 0.15 / 0.94 & 0.18 / 1.34 & 0.13 / 0.87 \\
\midrule
\multirow{12}{*}{\shortstack{Ours\\(Numba)\\$\mathbf{c}$, $a$ fixed}} & \multirow{2}{*}{Dense} & \multirow{2}{*}{$\sim\!10^5$} & 2 & $-0.3\%$ & $-0.3\%$ & $-0.4\%$ & $-0.4\%$ & $-0.3\%$ & $-0.2\%$ & $-0.5\%$ & 11.9 / 1.28 & 16.7 / 1.54 & 31.2 / 2.81 & 10.0 / 1.23 \\
 &  &  & 10 & $-0.3\%$ & $-0.3\%$ & $-0.4\%$ & $-0.4\%$ & $-0.3\%$ & $-0.2\%$ & $-0.5\%$ & 29.5 / 2.39 & 43.0 / 3.12 & 79.3 / 6.73 & 26.9 / 2.70 \\
\cmidrule(lr){2-15}
 & \multirow{2}{*}{$99 \times 99$} & \multirow{2}{*}{9801} & 2 & $-0.4\%$ & $-0.3\%$ & $-0.4\%$ & $-0.4\%$ & $-0.3\%$ & $-0.2\%$ & $-0.6\%$ & 4.40 / 1.10 & 5.22 / 0.96 & 6.63 / 1.57 & 4.23 / 0.80 \\
 &  &  & 10 & $-0.4\%$ & $-0.3\%$ & $-0.4\%$ & $-0.4\%$ & $-0.3\%$ & $-0.2\%$ & $-0.5\%$ & 14.0 / 1.30 & 16.0 / 1.62 & 20.0 / 3.45 & 13.7 / 1.40 \\
\cmidrule(lr){2-15}
 & \multirow{2}{*}{$25 \times 25$} & \multirow{2}{*}{625} & 2 & $-0.5\%$ & $-0.4\%$ & $-0.5\%$ & $-0.2\%$ & $-0.5\%$ & $-0.2\%$ & $-0.9\%$ & 0.55 / 0.70 & 0.89 / 0.77 & 1.82 / 1.03 & 0.54 / 0.70 \\
 &  &  & 10 & $-0.5\%$ & $-0.4\%$ & $-0.5\%$ & $-0.2\%$ & $-0.5\%$ & $-0.2\%$ & $-0.9\%$ & 1.54 / 0.91 & 2.72 / 1.07 & 5.53 / 1.85 & 1.53 / 0.95 \\
\cmidrule(lr){2-15}
 & \multirow{2}{*}{$7 \times 7$} & \multirow{2}{*}{49} & 2 & $-1.7\%$ & $-0.8\%$ & $-0.2\%$ & $+0.2\%$ & $-0.4\%$ & $-0.4\%$ & $-2.1\%$ & 0.15 / 0.66 & 0.18 / 0.70 & 0.26 / 0.88 & 0.15 / 0.64 \\
 &  &  & 10 & $-1.7\%$ & $-0.8\%$ & $-0.2\%$ & $+0.2\%$ & $-0.4\%$ & $-0.4\%$ & $-1.3\%$ & 0.23 / 0.80 & 0.33 / 0.91 & 0.55 / 1.38 & 0.22 / 0.83 \\
\cmidrule(lr){2-15}
 & \multirow{2}{*}{$5 \times 5$} & \multirow{2}{*}{25} & 2 & $-2.2\%$ & $-1.0\%$ & $-0.1\%$ & $+0.5\%$ & $-0.6\%$ & $-1.2\%$ & $-2.6\%$ & 0.13 / 0.66 & 0.16 / 0.74 & 0.19 / 0.85 & 0.13 / 0.65 \\
 &  &  & 10 & $-2.2\%$ & $-1.0\%$ & $-0.1\%$ & $+0.5\%$ & $-0.3\%$ & $-1.2\%$ & $-2.5\%$ & 0.17 / 0.80 & 0.22 / 0.91 & 0.34 / 1.38 & 0.17 / 0.83 \\
\cmidrule(lr){2-15}
 & \multirow{2}{*}{$3 \times 3$} & \multirow{2}{*}{9} & 2 & $-4.2\%$ & $-0.8\%$ & $+0.2\%$ & $+1.5\%$ & $-0.4\%$ & $+1844\%$ & $-3.1\%$ & 0.12 / 0.72 & 0.13 / 0.70 & 0.15 / 0.86 & 0.12 / 0.69 \\
 &  &  & 10 & $-4.2\%$ & $-0.8\%$ & $+0.2\%$ & $+1.5\%$ & $-1.1\%$ & $+1813\%$ & $-1.9\%$ & 0.13 / 0.80 & 0.15 / 0.91 & 0.21 / 1.40 & 0.13 / 0.83 \\
% rows sparse_full: end
\bottomrule
\end{tabular}}
\label{tab:sparse_full}
\end{table*}

%% file: tab/ablation_pinhole.tex
\begin{table*}[t]
\caption{\textbf{Representation ablation on pinhole benchmarks.} Median error and mAA per benchmark of the variants of Tab.~\ref{tab:ablations}, as mean and standard deviation over five seeds. Bold: best mean. Each variant fits the pinhole model, except the direct regression, which has no fit.}
\centering
\resizebox{\textwidth}{!}{%
\begin{tabular}{l cc cc cc cc}
\toprule
 & \multicolumn{2}{c}{AE ($^\circ$)} & \multicolumn{2}{c}{RE (px)} & \multicolumn{2}{c}{vFoV ($^\circ$)} & \multicolumn{2}{c}{hFoV ($^\circ$)} \\
\cmidrule(lr){2-3} \cmidrule(lr){4-5} \cmidrule(lr){6-7} \cmidrule(lr){8-9}
Method & Median $\downarrow$ & mAA ($5^\circ$) $\uparrow$ & Median $\downarrow$ & mAA (50\,px) $\uparrow$ & Median $\downarrow$ & mAA ($5^\circ$) $\uparrow$ & Median $\downarrow$ & mAA ($5^\circ$) $\uparrow$ \\
\midrule
\multicolumn{9}{l}{\textit{Stanford2D3D}~\cite{armeni2017joint}} \\
\midrule
$5 \times 5$ $\vec\omega$ & \textbf{0.997} $\pm$ 0.009 & \textbf{72.8} $\pm$ 0.3 & 10.9 $\pm$ 0.1 & \textbf{66.5} $\pm$ 0.4 & \textbf{2.24} $\pm$ 0.02 & 50.5 $\pm$ 0.2 & 2.55 $\pm$ 0.04 & \textbf{47.1} $\pm$ 0.2 \\
$5 \times 5$ $\vec\omega$, free $\vec{c}, a$ & 1.000 $\pm$ 0.008 & \textbf{72.8} $\pm$ 0.3 & 11.0 $\pm$ 0.1 & 66.4 $\pm$ 0.4 & \textbf{2.24} $\pm$ 0.03 & \textbf{50.6} $\pm$ 0.3 & 2.56 $\pm$ 0.03 & 46.9 $\pm$ 0.3 \\
$5 \times 5$ $\vec\omega$, no dense loss & 1.068 $\pm$ 0.019 & 71.3 $\pm$ 0.2 & 11.4 $\pm$ 0.3 & 65.2 $\pm$ 0.3 & 2.41 $\pm$ 0.04 & 48.7 $\pm$ 0.4 & 2.71 $\pm$ 0.05 & 45.2 $\pm$ 0.5 \\
$5 \times 5$ $\vec\omega$, $\vec\omega$ aux & 1.004 $\pm$ 0.017 & 72.6 $\pm$ 0.4 & \textbf{10.8} $\pm$ 0.1 & 66.3 $\pm$ 0.3 & 2.26 $\pm$ 0.04 & 50.1 $\pm$ 0.5 & \textbf{2.54} $\pm$ 0.04 & 46.7 $\pm$ 0.5 \\
$7 \times 7$ $\vec\omega$ & 1.015 $\pm$ 0.031 & 72.3 $\pm$ 0.2 & 11.0 $\pm$ 0.3 & 66.0 $\pm$ 0.2 & 2.29 $\pm$ 0.07 & 49.9 $\pm$ 0.4 & 2.58 $\pm$ 0.07 & 46.5 $\pm$ 0.4 \\
$5 \times 5$ $\vec\theta$ & 1.001 $\pm$ 0.028 & 72.6 $\pm$ 0.3 & 11.0 $\pm$ 0.3 & 66.3 $\pm$ 0.5 & 2.25 $\pm$ 0.06 & 50.1 $\pm$ 0.6 & 2.56 $\pm$ 0.06 & 46.7 $\pm$ 0.6 \\
$5 \times 5$ $\vec\theta$, no dense loss & 1.082 $\pm$ 0.015 & 71.1 $\pm$ 0.2 & 11.8 $\pm$ 0.2 & 65.0 $\pm$ 0.2 & 2.44 $\pm$ 0.03 & 48.0 $\pm$ 0.2 & 2.75 $\pm$ 0.03 & 44.6 $\pm$ 0.3 \\
Radial profile & 1.014 $\pm$ 0.027 & 72.7 $\pm$ 0.4 & 10.9 $\pm$ 0.3 & \textbf{66.5} $\pm$ 0.4 & 2.29 $\pm$ 0.06 & 49.9 $\pm$ 0.7 & 2.58 $\pm$ 0.06 & 46.5 $\pm$ 0.7 \\
Direct regression & 1.006 $\pm$ 0.018 & 72.7 $\pm$ 0.5 & 10.9 $\pm$ 0.2 & 66.4 $\pm$ 0.6 & 2.26 $\pm$ 0.04 & 50.2 $\pm$ 0.3 & 2.56 $\pm$ 0.05 & 46.6 $\pm$ 0.4 \\
\midrule
\multicolumn{9}{l}{\textit{TartanAir}~\cite{wang2020tartanair}} \\
\midrule
$5 \times 5$ $\vec\omega$ & 1.383 $\pm$ 0.066 & 64.2 $\pm$ 0.9 & 12.4 $\pm$ 0.6 & 62.2 $\pm$ 0.8 & 3.12 $\pm$ 0.15 & 41.5 $\pm$ 1.0 & 3.26 $\pm$ 0.16 & 40.3 $\pm$ 1.0 \\
$5 \times 5$ $\vec\omega$, free $\vec{c}, a$ & 1.387 $\pm$ 0.068 & 64.1 $\pm$ 0.9 & 12.4 $\pm$ 0.6 & 62.1 $\pm$ 0.8 & 3.02 $\pm$ 0.13 & 42.2 $\pm$ 0.8 & 3.33 $\pm$ 0.18 & 39.8 $\pm$ 1.2 \\
$5 \times 5$ $\vec\omega$, no dense loss & 1.334 $\pm$ 0.022 & 65.1 $\pm$ 1.0 & 11.9 $\pm$ 0.1 & 63.0 $\pm$ 1.0 & 3.00 $\pm$ 0.05 & 42.2 $\pm$ 0.5 & 3.13 $\pm$ 0.06 & 41.1 $\pm$ 0.5 \\
$5 \times 5$ $\vec\omega$, $\vec\omega$ aux & 1.337 $\pm$ 0.062 & 65.1 $\pm$ 1.4 & 12.0 $\pm$ 0.6 & 63.2 $\pm$ 1.3 & 3.01 $\pm$ 0.14 & 42.4 $\pm$ 1.4 & 3.15 $\pm$ 0.15 & 41.3 $\pm$ 1.4 \\
$7 \times 7$ $\vec\omega$ & 1.364 $\pm$ 0.045 & 64.2 $\pm$ 0.9 & 12.2 $\pm$ 0.4 & 62.2 $\pm$ 0.8 & 3.07 $\pm$ 0.10 & 41.7 $\pm$ 0.7 & 3.21 $\pm$ 0.11 & 40.6 $\pm$ 0.7 \\
$5 \times 5$ $\vec\theta$ & 1.305 $\pm$ 0.037 & 65.4 $\pm$ 0.7 & 11.7 $\pm$ 0.3 & 63.4 $\pm$ 0.6 & 2.94 $\pm$ 0.08 & 42.8 $\pm$ 0.5 & 3.07 $\pm$ 0.08 & 41.6 $\pm$ 0.5 \\
$5 \times 5$ $\vec\theta$, no dense loss & 1.333 $\pm$ 0.081 & 65.2 $\pm$ 1.2 & 11.9 $\pm$ 0.8 & 63.1 $\pm$ 1.2 & 3.00 $\pm$ 0.18 & 42.2 $\pm$ 1.4 & 3.14 $\pm$ 0.20 & 41.0 $\pm$ 1.4 \\
Radial profile & \textbf{1.291} $\pm$ 0.021 & \textbf{65.8} $\pm$ 0.4 & \textbf{11.5} $\pm$ 0.2 & \textbf{63.8} $\pm$ 0.4 & \textbf{2.91} $\pm$ 0.04 & \textbf{42.9} $\pm$ 0.4 & \textbf{3.04} $\pm$ 0.04 & \textbf{41.8} $\pm$ 0.3 \\
Direct regression & 1.344 $\pm$ 0.044 & 64.7 $\pm$ 0.7 & 11.9 $\pm$ 0.4 & 62.7 $\pm$ 0.6 & 3.03 $\pm$ 0.11 & 42.2 $\pm$ 0.7 & 3.16 $\pm$ 0.10 & 41.1 $\pm$ 0.7 \\
\midrule
\multicolumn{9}{l}{\textit{MegaDepth}~\cite{li2018megadepth}} \\
\midrule
$5 \times 5$ $\vec\omega$ & 0.640 $\pm$ 0.016 & 86.8 $\pm$ 0.5 & 22.2 $\pm$ 0.5 & 49.3 $\pm$ 0.8 & 1.53 $\pm$ 0.06 & 64.7 $\pm$ 0.9 & 1.64 $\pm$ 0.03 & 61.8 $\pm$ 0.9 \\
$5 \times 5$ $\vec\omega$, free $\vec{c}, a$ & 0.642 $\pm$ 0.015 & 86.8 $\pm$ 0.5 & 22.3 $\pm$ 0.4 & 49.0 $\pm$ 0.8 & 1.53 $\pm$ 0.07 & 64.6 $\pm$ 1.0 & 1.64 $\pm$ 0.02 & 61.8 $\pm$ 0.8 \\
$5 \times 5$ $\vec\omega$, no dense loss & 0.665 $\pm$ 0.018 & 86.1 $\pm$ 0.3 & 23.1 $\pm$ 0.5 & 48.2 $\pm$ 0.7 & 1.59 $\pm$ 0.06 & 63.4 $\pm$ 0.4 & 1.71 $\pm$ 0.05 & 60.5 $\pm$ 0.7 \\
$5 \times 5$ $\vec\omega$, $\vec\omega$ aux & 0.655 $\pm$ 0.030 & 86.6 $\pm$ 0.6 & 22.7 $\pm$ 0.8 & 48.8 $\pm$ 0.9 & 1.56 $\pm$ 0.06 & 64.1 $\pm$ 1.1 & 1.69 $\pm$ 0.07 & 61.3 $\pm$ 1.2 \\
$7 \times 7$ $\vec\omega$ & 0.645 $\pm$ 0.010 & 86.7 $\pm$ 0.3 & 22.7 $\pm$ 0.2 & 48.9 $\pm$ 0.2 & 1.54 $\pm$ 0.02 & 64.3 $\pm$ 0.4 & 1.67 $\pm$ 0.04 & 61.4 $\pm$ 0.4 \\
$5 \times 5$ $\vec\theta$ & \textbf{0.627} $\pm$ 0.013 & 86.9 $\pm$ 0.5 & 22.0 $\pm$ 0.5 & \textbf{49.7} $\pm$ 0.8 & \textbf{1.49} $\pm$ 0.04 & \textbf{65.1} $\pm$ 0.8 & \textbf{1.62} $\pm$ 0.04 & \textbf{62.3} $\pm$ 0.8 \\
$5 \times 5$ $\vec\theta$, no dense loss & 0.665 $\pm$ 0.020 & 86.4 $\pm$ 0.6 & 23.0 $\pm$ 0.5 & 48.3 $\pm$ 0.6 & 1.58 $\pm$ 0.05 & 63.6 $\pm$ 0.7 & 1.70 $\pm$ 0.05 & 60.8 $\pm$ 0.8 \\
Radial profile & 0.656 $\pm$ 0.032 & 86.2 $\pm$ 0.3 & 22.7 $\pm$ 1.0 & 48.8 $\pm$ 1.1 & 1.55 $\pm$ 0.07 & 63.7 $\pm$ 1.0 & 1.70 $\pm$ 0.08 & 60.9 $\pm$ 1.2 \\
Direct regression & 0.630 $\pm$ 0.018 & \textbf{87.0} $\pm$ 0.6 & \textbf{21.9} $\pm$ 0.7 & \textbf{49.7} $\pm$ 0.8 & 1.52 $\pm$ 0.07 & 64.8 $\pm$ 1.0 & 1.65 $\pm$ 0.06 & 61.7 $\pm$ 1.0 \\
\midrule
\multicolumn{9}{l}{\textit{LaMAR}~\cite{sarlin2022lamar}} \\
\midrule
$5 \times 5$ $\vec\omega$ & 0.705 $\pm$ 0.034 & 84.4 $\pm$ 0.4 & 23.9 $\pm$ 1.0 & 47.3 $\pm$ 1.1 & 1.87 $\pm$ 0.09 & 57.5 $\pm$ 1.0 & 1.61 $\pm$ 0.08 & 62.4 $\pm$ 0.9 \\
$5 \times 5$ $\vec\omega$, free $\vec{c}, a$ & 0.707 $\pm$ 0.034 & 84.4 $\pm$ 0.4 & 23.9 $\pm$ 1.0 & 47.0 $\pm$ 1.1 & 1.88 $\pm$ 0.10 & 57.2 $\pm$ 1.2 & 1.58 $\pm$ 0.07 & 62.8 $\pm$ 0.8 \\
$5 \times 5$ $\vec\omega$, no dense loss & 0.728 $\pm$ 0.022 & 83.4 $\pm$ 0.5 & 24.8 $\pm$ 0.7 & 45.7 $\pm$ 0.5 & 1.93 $\pm$ 0.06 & 55.9 $\pm$ 0.5 & 1.66 $\pm$ 0.05 & 60.8 $\pm$ 0.4 \\
$5 \times 5$ $\vec\omega$, $\vec\omega$ aux & 0.693 $\pm$ 0.033 & 84.7 $\pm$ 0.7 & 23.6 $\pm$ 1.0 & 47.4 $\pm$ 1.3 & 1.84 $\pm$ 0.08 & 57.6 $\pm$ 1.2 & 1.58 $\pm$ 0.08 & 62.5 $\pm$ 1.1 \\
$7 \times 7$ $\vec\omega$ & 0.698 $\pm$ 0.025 & 84.3 $\pm$ 0.1 & 23.7 $\pm$ 0.8 & 47.2 $\pm$ 0.5 & 1.85 $\pm$ 0.06 & 57.5 $\pm$ 0.3 & 1.59 $\pm$ 0.06 & 62.3 $\pm$ 0.2 \\
$5 \times 5$ $\vec\theta$ & \textbf{0.670} $\pm$ 0.030 & \textbf{84.8} $\pm$ 0.2 & \textbf{22.8} $\pm$ 1.0 & \textbf{48.0} $\pm$ 0.8 & \textbf{1.77} $\pm$ 0.08 & \textbf{58.3} $\pm$ 0.8 & \textbf{1.53} $\pm$ 0.06 & \textbf{63.1} $\pm$ 0.8 \\
$5 \times 5$ $\vec\theta$, no dense loss & 0.748 $\pm$ 0.029 & 83.0 $\pm$ 0.4 & 25.2 $\pm$ 1.0 & 45.4 $\pm$ 0.9 & 1.98 $\pm$ 0.08 & 55.4 $\pm$ 0.8 & 1.70 $\pm$ 0.06 & 60.3 $\pm$ 0.8 \\
Radial profile & 0.677 $\pm$ 0.019 & 84.7 $\pm$ 0.5 & 23.0 $\pm$ 0.6 & \textbf{48.0} $\pm$ 0.7 & 1.79 $\pm$ 0.05 & \textbf{58.3} $\pm$ 0.8 & 1.54 $\pm$ 0.04 & \textbf{63.1} $\pm$ 0.7 \\
Direct regression & 0.696 $\pm$ 0.024 & 84.7 $\pm$ 0.3 & 23.6 $\pm$ 0.8 & 47.6 $\pm$ 0.7 & 1.86 $\pm$ 0.05 & 57.9 $\pm$ 0.6 & 1.58 $\pm$ 0.06 & 62.9 $\pm$ 0.6 \\
\bottomrule
\end{tabular}}
\label{tab:ablation_pinhole}
\end{table*}

%% file: tab/ablation_distorted.tex
\begin{table*}[t]
\caption{\textbf{Representation ablation on distorted benchmarks.} Median error and mAA per benchmark of the variants of Tab.~\ref{tab:ablations}, as mean and standard deviation over five seeds. Bold: best mean. Each variant fits the camera model of Sec.~\ref{sec:metrics}, except the direct regression, which has no fit.}
\centering
\resizebox{0.92\textwidth}{!}{%
\begin{tabular}{l cc cc cc cc}
\toprule
 & \multicolumn{2}{c}{AE ($^\circ$)} & \multicolumn{2}{c}{RE (px)} & \multicolumn{2}{c}{vFoV ($^\circ$)} & \multicolumn{2}{c}{hFoV ($^\circ$)} \\
\cmidrule(lr){2-3} \cmidrule(lr){4-5} \cmidrule(lr){6-7} \cmidrule(lr){8-9}
Method & Median $\downarrow$ & mAA ($5^\circ$) $\uparrow$ & Median $\downarrow$ & mAA (50\,px) $\uparrow$ & Median $\downarrow$ & mAA ($5^\circ$) $\uparrow$ & Median $\downarrow$ & mAA ($5^\circ$) $\uparrow$ \\
\midrule
\multicolumn{9}{l}{\textit{MegaDepth-radial}~\cite{li2018megadepth}} \\
\midrule
$5 \times 5$ $\vec\omega$ & \textbf{0.586} $\pm$ 0.006 & \textbf{87.3} $\pm$ 0.3 & \textbf{19.8} $\pm$ 0.1 & 51.2 $\pm$ 0.4 & \textbf{1.43} $\pm$ 0.02 & \textbf{65.8} $\pm$ 0.5 & \textbf{1.52} $\pm$ 0.02 & \textbf{63.3} $\pm$ 0.6 \\
$5 \times 5$ $\vec\omega$, free $\vec{c}, a$ & 0.588 $\pm$ 0.006 & \textbf{87.3} $\pm$ 0.3 & \textbf{19.8} $\pm$ 0.1 & 50.9 $\pm$ 0.4 & 1.44 $\pm$ 0.03 & 65.6 $\pm$ 0.6 & \textbf{1.52} $\pm$ 0.02 & \textbf{63.3} $\pm$ 0.5 \\
$5 \times 5$ $\vec\omega$, no dense loss & 0.609 $\pm$ 0.016 & 86.8 $\pm$ 0.3 & 20.5 $\pm$ 0.6 & 50.1 $\pm$ 0.5 & 1.49 $\pm$ 0.05 & 64.6 $\pm$ 0.6 & 1.60 $\pm$ 0.05 & 61.9 $\pm$ 0.5 \\
$5 \times 5$ $\vec\omega$, $\vec\omega$ aux & 0.604 $\pm$ 0.023 & 87.1 $\pm$ 0.4 & 20.4 $\pm$ 0.8 & 50.6 $\pm$ 0.5 & 1.47 $\pm$ 0.04 & 65.1 $\pm$ 0.5 & 1.57 $\pm$ 0.05 & 62.6 $\pm$ 0.5 \\
$7 \times 7$ $\vec\omega$ & 0.595 $\pm$ 0.013 & 87.2 $\pm$ 0.2 & 20.2 $\pm$ 0.4 & 51.0 $\pm$ 0.3 & \textbf{1.43} $\pm$ 0.03 & 65.5 $\pm$ 0.2 & 1.56 $\pm$ 0.01 & 63.0 $\pm$ 0.3 \\
$5 \times 5$ $\vec\theta$ & 0.590 $\pm$ 0.016 & \textbf{87.3} $\pm$ 0.4 & 20.0 $\pm$ 0.6 & \textbf{51.3} $\pm$ 0.6 & 1.44 $\pm$ 0.03 & 65.7 $\pm$ 0.7 & 1.56 $\pm$ 0.04 & 63.1 $\pm$ 0.7 \\
$5 \times 5$ $\vec\theta$, no dense loss & 0.600 $\pm$ 0.011 & 87.1 $\pm$ 0.4 & 20.1 $\pm$ 0.3 & 50.9 $\pm$ 0.4 & 1.46 $\pm$ 0.03 & 65.2 $\pm$ 0.6 & 1.57 $\pm$ 0.02 & 62.7 $\pm$ 0.5 \\
Radial profile & 0.613 $\pm$ 0.015 & 87.0 $\pm$ 0.3 & 20.7 $\pm$ 0.7 & 50.6 $\pm$ 0.4 & 1.49 $\pm$ 0.03 & 65.0 $\pm$ 0.5 & 1.59 $\pm$ 0.04 & 62.5 $\pm$ 0.5 \\
Direct regression & 0.608 $\pm$ 0.009 & 87.0 $\pm$ 0.4 & 20.7 $\pm$ 0.2 & 50.5 $\pm$ 0.2 & 1.47 $\pm$ 0.02 & 64.9 $\pm$ 0.3 & 1.57 $\pm$ 0.02 & 62.4 $\pm$ 0.4 \\
\midrule
\multicolumn{9}{l}{\textit{ScanNet++}~\cite{yeshwanth2023scannetpp}} \\
\midrule
$5 \times 5$ $\vec\omega$ & 1.010 $\pm$ 0.015 & 80.2 $\pm$ 1.6 & \textbf{12.8} $\pm$ 0.1 & 67.0 $\pm$ 1.1 & \textbf{1.95} $\pm$ 0.07 & \textbf{56.9} $\pm$ 1.1 & \textbf{2.30} $\pm$ 0.05 & 51.8 $\pm$ 0.6 \\
$5 \times 5$ $\vec\omega$, free $\vec{c}, a$ & 1.058 $\pm$ 0.018 & 79.6 $\pm$ 1.1 & 13.5 $\pm$ 0.4 & 66.2 $\pm$ 0.8 & 1.97 $\pm$ 0.04 & \textbf{56.9} $\pm$ 0.8 & \textbf{2.30} $\pm$ 0.07 & \textbf{52.0} $\pm$ 0.8 \\
$5 \times 5$ $\vec\omega$, no dense loss & 1.116 $\pm$ 0.078 & 76.4 $\pm$ 3.3 & 14.2 $\pm$ 1.0 & 63.7 $\pm$ 2.6 & 2.24 $\pm$ 0.12 & 52.0 $\pm$ 2.5 & 2.39 $\pm$ 0.13 & 49.8 $\pm$ 1.7 \\
$5 \times 5$ $\vec\omega$, $\vec\omega$ aux & 1.041 $\pm$ 0.016 & 79.7 $\pm$ 1.0 & 13.3 $\pm$ 0.3 & 66.5 $\pm$ 0.7 & 2.00 $\pm$ 0.05 & 56.4 $\pm$ 0.7 & 2.36 $\pm$ 0.11 & 50.5 $\pm$ 1.2 \\
$7 \times 7$ $\vec\omega$ & \textbf{1.009} $\pm$ 0.014 & 80.7 $\pm$ 0.8 & \textbf{12.8} $\pm$ 0.2 & \textbf{67.3} $\pm$ 0.7 & 1.98 $\pm$ 0.03 & 56.8 $\pm$ 1.1 & 2.31 $\pm$ 0.07 & 51.6 $\pm$ 0.7 \\
$5 \times 5$ $\vec\theta$ & 1.023 $\pm$ 0.031 & 80.9 $\pm$ 1.1 & 13.2 $\pm$ 0.5 & 67.1 $\pm$ 1.3 & 1.99 $\pm$ 0.10 & 56.7 $\pm$ 1.3 & 2.51 $\pm$ 0.22 & 49.1 $\pm$ 3.2 \\
$5 \times 5$ $\vec\theta$, no dense loss & 1.070 $\pm$ 0.043 & 79.7 $\pm$ 1.3 & 13.7 $\pm$ 0.7 & 66.0 $\pm$ 1.4 & 2.16 $\pm$ 0.10 & 53.6 $\pm$ 1.4 & 2.45 $\pm$ 0.19 & 49.7 $\pm$ 2.7 \\
Radial profile & 1.038 $\pm$ 0.020 & 81.0 $\pm$ 0.7 & 13.4 $\pm$ 0.3 & 67.0 $\pm$ 0.9 & 2.03 $\pm$ 0.07 & 56.2 $\pm$ 1.1 & 2.40 $\pm$ 0.13 & 50.5 $\pm$ 1.7 \\
Direct regression & 1.035 $\pm$ 0.020 & \textbf{81.2} $\pm$ 0.4 & 13.6 $\pm$ 0.3 & 66.7 $\pm$ 0.6 & 2.00 $\pm$ 0.07 & \textbf{56.9} $\pm$ 1.1 & 2.40 $\pm$ 0.15 & 50.1 $\pm$ 2.2 \\
\midrule
\multicolumn{9}{l}{\textit{Mono}~\cite{engel2016photometrically}} \\
\midrule
$5 \times 5$ $\vec\omega$ & 1.358 $\pm$ 0.028 & \textbf{68.2} $\pm$ 0.9 & 18.1 $\pm$ 0.4 & \textbf{53.4} $\pm$ 0.6 & 2.94 $\pm$ 0.07 & 42.8 $\pm$ 1.1 & 3.21 $\pm$ 0.10 & 40.1 $\pm$ 1.1 \\
$5 \times 5$ $\vec\omega$, free $\vec{c}, a$ & 1.361 $\pm$ 0.028 & 68.1 $\pm$ 0.8 & 18.2 $\pm$ 0.4 & 53.3 $\pm$ 0.6 & 3.00 $\pm$ 0.07 & 42.3 $\pm$ 1.3 & 3.20 $\pm$ 0.13 & 40.2 $\pm$ 1.1 \\
$5 \times 5$ $\vec\omega$, no dense loss & 1.463 $\pm$ 0.070 & 66.2 $\pm$ 1.2 & 19.4 $\pm$ 1.0 & 51.5 $\pm$ 1.2 & 3.20 $\pm$ 0.18 & 40.3 $\pm$ 2.2 & 3.44 $\pm$ 0.17 & 38.1 $\pm$ 1.7 \\
$5 \times 5$ $\vec\omega$, $\vec\omega$ aux & 1.385 $\pm$ 0.059 & 68.0 $\pm$ 1.0 & 18.4 $\pm$ 0.8 & 53.2 $\pm$ 0.8 & 3.01 $\pm$ 0.15 & 42.4 $\pm$ 1.5 & 3.22 $\pm$ 0.15 & 40.0 $\pm$ 1.5 \\
$7 \times 7$ $\vec\omega$ & 1.385 $\pm$ 0.037 & 67.5 $\pm$ 1.0 & 18.5 $\pm$ 0.6 & 52.6 $\pm$ 1.0 & 3.00 $\pm$ 0.08 & 42.3 $\pm$ 0.8 & 3.29 $\pm$ 0.09 & 39.6 $\pm$ 0.8 \\
$5 \times 5$ $\vec\theta$ & \textbf{1.338} $\pm$ 0.091 & 68.1 $\pm$ 1.1 & \textbf{17.8} $\pm$ 1.1 & 53.1 $\pm$ 1.0 & \textbf{2.88} $\pm$ 0.24 & \textbf{43.3} $\pm$ 1.6 & \textbf{3.18} $\pm$ 0.17 & \textbf{40.3} $\pm$ 1.4 \\
$5 \times 5$ $\vec\theta$, no dense loss & 1.405 $\pm$ 0.018 & 66.8 $\pm$ 0.9 & 18.7 $\pm$ 0.3 & 52.1 $\pm$ 0.7 & 3.07 $\pm$ 0.04 & 41.4 $\pm$ 0.6 & 3.25 $\pm$ 0.15 & 39.5 $\pm$ 1.1 \\
Radial profile & 1.382 $\pm$ 0.051 & 67.3 $\pm$ 1.1 & 18.5 $\pm$ 0.7 & 52.3 $\pm$ 1.0 & 2.99 $\pm$ 0.12 & 42.4 $\pm$ 1.3 & 3.30 $\pm$ 0.16 & 39.5 $\pm$ 1.2 \\
Direct regression & 1.391 $\pm$ 0.077 & 67.6 $\pm$ 1.3 & 18.5 $\pm$ 1.1 & 52.8 $\pm$ 1.3 & 3.01 $\pm$ 0.18 & 42.5 $\pm$ 1.8 & 3.21 $\pm$ 0.20 & \textbf{40.3} $\pm$ 1.9 \\
\midrule
\multicolumn{9}{l}{\textit{Cathedral}~\cite{kocur2026minimal}} \\
\midrule
$5 \times 5$ $\vec\omega$ & 0.812 $\pm$ 0.053 & 83.7 $\pm$ 0.8 & 9.3 $\pm$ 0.5 & \textbf{76.5} $\pm$ 0.5 & 1.15 $\pm$ 0.10 & 73.7 $\pm$ 1.6 & 1.51 $\pm$ 0.10 & 62.8 $\pm$ 2.5 \\
$5 \times 5$ $\vec\omega$, free $\vec{c}, a$ & 0.826 $\pm$ 0.052 & 83.6 $\pm$ 0.9 & 9.4 $\pm$ 0.6 & \textbf{76.5} $\pm$ 0.5 & 1.14 $\pm$ 0.05 & 73.9 $\pm$ 0.9 & 1.54 $\pm$ 0.13 & 62.2 $\pm$ 2.9 \\
$5 \times 5$ $\vec\omega$, no dense loss & 0.845 $\pm$ 0.036 & 82.8 $\pm$ 1.3 & 9.8 $\pm$ 0.6 & 75.7 $\pm$ 1.1 & 1.21 $\pm$ 0.11 & 72.0 $\pm$ 2.8 & 1.57 $\pm$ 0.10 & 60.6 $\pm$ 2.3 \\
$5 \times 5$ $\vec\omega$, $\vec\omega$ aux & \textbf{0.794} $\pm$ 0.034 & \textbf{84.0} $\pm$ 0.8 & \textbf{9.0} $\pm$ 0.5 & \textbf{76.5} $\pm$ 0.7 & \textbf{1.11} $\pm$ 0.05 & \textbf{74.2} $\pm$ 1.1 & \textbf{1.46} $\pm$ 0.11 & \textbf{63.9} $\pm$ 2.2 \\
$7 \times 7$ $\vec\omega$ & 0.825 $\pm$ 0.063 & 83.2 $\pm$ 1.3 & 9.4 $\pm$ 0.8 & 76.2 $\pm$ 1.0 & 1.18 $\pm$ 0.15 & 72.9 $\pm$ 2.8 & 1.51 $\pm$ 0.10 & 62.6 $\pm$ 2.3 \\
$5 \times 5$ $\vec\theta$ & 0.876 $\pm$ 0.017 & 82.1 $\pm$ 0.7 & 10.0 $\pm$ 0.1 & 75.3 $\pm$ 0.3 & 1.28 $\pm$ 0.06 & 70.3 $\pm$ 1.2 & 1.64 $\pm$ 0.05 & 59.3 $\pm$ 0.7 \\
$5 \times 5$ $\vec\theta$, no dense loss & 0.866 $\pm$ 0.055 & 82.8 $\pm$ 1.6 & 9.9 $\pm$ 0.7 & 75.7 $\pm$ 0.9 & 1.28 $\pm$ 0.11 & 71.1 $\pm$ 2.3 & 1.60 $\pm$ 0.12 & 60.8 $\pm$ 3.0 \\
Radial profile & 0.841 $\pm$ 0.044 & 82.6 $\pm$ 1.2 & 9.6 $\pm$ 0.5 & 75.8 $\pm$ 0.7 & 1.28 $\pm$ 0.04 & 70.8 $\pm$ 1.5 & 1.53 $\pm$ 0.14 & 61.6 $\pm$ 2.8 \\
Direct regression & 0.862 $\pm$ 0.061 & 82.7 $\pm$ 1.0 & 9.7 $\pm$ 0.5 & 75.8 $\pm$ 0.7 & 1.31 $\pm$ 0.11 & 70.3 $\pm$ 1.9 & 1.63 $\pm$ 0.10 & 60.9 $\pm$ 1.5 \\
\midrule
\multicolumn{9}{l}{\textit{InFlux}~\cite{liang2025influx}} \\
\midrule
$5 \times 5$ $\vec\omega$ & \textbf{1.981} $\pm$ 0.106 & \textbf{57.9} $\pm$ 1.4 & 117.3 $\pm$ 3.8 & \textbf{11.5} $\pm$ 0.6 & \textbf{3.99} $\pm$ 0.22 & 32.5 $\pm$ 1.3 & 6.17 $\pm$ 0.32 & 21.2 $\pm$ 0.8 \\
$5 \times 5$ $\vec\omega$, free $\vec{c}, a$ & 1.984 $\pm$ 0.109 & 57.8 $\pm$ 1.4 & 117.0 $\pm$ 3.5 & 11.2 $\pm$ 0.6 & 4.05 $\pm$ 0.20 & 32.2 $\pm$ 1.4 & \textbf{6.15} $\pm$ 0.32 & 21.3 $\pm$ 0.8 \\
$5 \times 5$ $\vec\omega$, no dense loss & 2.058 $\pm$ 0.114 & 57.2 $\pm$ 1.7 & \textbf{116.3} $\pm$ 3.8 & 11.2 $\pm$ 1.0 & 4.15 $\pm$ 0.24 & 31.6 $\pm$ 2.0 & 6.42 $\pm$ 0.34 & 20.4 $\pm$ 1.8 \\
$5 \times 5$ $\vec\omega$, $\vec\omega$ aux & 2.126 $\pm$ 0.064 & 56.3 $\pm$ 1.8 & 118.7 $\pm$ 6.0 & 11.1 $\pm$ 0.6 & 4.30 $\pm$ 0.12 & 30.8 $\pm$ 1.4 & 6.64 $\pm$ 0.17 & 20.2 $\pm$ 1.2 \\
$7 \times 7$ $\vec\omega$ & 2.143 $\pm$ 0.068 & 56.1 $\pm$ 2.1 & 122.0 $\pm$ 3.8 & 10.6 $\pm$ 1.0 & 4.33 $\pm$ 0.14 & 30.6 $\pm$ 2.0 & 6.65 $\pm$ 0.24 & 19.9 $\pm$ 1.7 \\
$5 \times 5$ $\vec\theta$ & 2.146 $\pm$ 0.185 & 55.7 $\pm$ 2.8 & 122.5 $\pm$ 7.1 & 10.6 $\pm$ 1.2 & 4.34 $\pm$ 0.36 & 30.6 $\pm$ 2.6 & 6.68 $\pm$ 0.59 & 19.9 $\pm$ 1.9 \\
$5 \times 5$ $\vec\theta$, no dense loss & 2.025 $\pm$ 0.169 & 57.5 $\pm$ 2.4 & 117.6 $\pm$ 9.0 & \textbf{11.5} $\pm$ 1.2 & 4.09 $\pm$ 0.35 & \textbf{32.7} $\pm$ 2.0 & 6.27 $\pm$ 0.51 & \textbf{21.5} $\pm$ 1.3 \\
Radial profile & 2.157 $\pm$ 0.101 & 55.8 $\pm$ 1.2 & 122.0 $\pm$ 2.8 & 10.9 $\pm$ 1.0 & 4.38 $\pm$ 0.22 & 30.4 $\pm$ 1.7 & 6.71 $\pm$ 0.29 & 19.8 $\pm$ 1.7 \\
Direct regression & 2.267 $\pm$ 0.142 & 54.2 $\pm$ 3.1 & 125.9 $\pm$ 5.4 & 10.4 $\pm$ 1.6 & 4.57 $\pm$ 0.28 & 28.6 $\pm$ 2.8 & 7.07 $\pm$ 0.44 & 18.1 $\pm$ 2.7 \\
\midrule
\multicolumn{9}{l}{\textit{InFlux++}~\cite{liang2026influxpp}} \\
\midrule
$5 \times 5$ $\vec\omega$ & 2.176 $\pm$ 0.057 & 51.2 $\pm$ 0.5 & 84.8 $\pm$ 1.3 & 12.5 $\pm$ 0.5 & 4.21 $\pm$ 0.12 & \textbf{33.4} $\pm$ 0.8 & 6.30 $\pm$ 0.15 & 24.1 $\pm$ 0.8 \\
$5 \times 5$ $\vec\omega$, free $\vec{c}, a$ & 2.179 $\pm$ 0.051 & 51.1 $\pm$ 0.4 & 85.3 $\pm$ 1.2 & 12.3 $\pm$ 0.5 & 4.19 $\pm$ 0.13 & \textbf{33.4} $\pm$ 0.8 & 6.28 $\pm$ 0.15 & 24.1 $\pm$ 0.8 \\
$5 \times 5$ $\vec\omega$, no dense loss & 2.152 $\pm$ 0.048 & 51.5 $\pm$ 1.1 & 84.1 $\pm$ 1.5 & 12.1 $\pm$ 0.4 & 4.20 $\pm$ 0.11 & 33.2 $\pm$ 0.8 & 6.23 $\pm$ 0.20 & 24.1 $\pm$ 0.8 \\
$5 \times 5$ $\vec\omega$, $\vec\omega$ aux & 2.190 $\pm$ 0.035 & 50.9 $\pm$ 1.1 & 85.2 $\pm$ 2.9 & 12.2 $\pm$ 0.5 & 4.22 $\pm$ 0.07 & 33.0 $\pm$ 1.0 & 6.32 $\pm$ 0.15 & 23.7 $\pm$ 1.0 \\
$7 \times 7$ $\vec\omega$ & 2.184 $\pm$ 0.059 & 50.8 $\pm$ 1.0 & 85.8 $\pm$ 2.9 & 12.5 $\pm$ 0.5 & 4.23 $\pm$ 0.12 & 32.9 $\pm$ 0.7 & 6.38 $\pm$ 0.19 & 23.9 $\pm$ 0.9 \\
$5 \times 5$ $\vec\theta$ & 2.162 $\pm$ 0.029 & \textbf{51.6} $\pm$ 0.6 & \textbf{82.1} $\pm$ 1.0 & 12.4 $\pm$ 0.3 & \textbf{4.18} $\pm$ 0.08 & \textbf{33.4} $\pm$ 0.5 & 6.31 $\pm$ 0.13 & 23.9 $\pm$ 0.6 \\
$5 \times 5$ $\vec\theta$, no dense loss & \textbf{2.144} $\pm$ 0.108 & 51.4 $\pm$ 1.3 & 85.1 $\pm$ 4.0 & 12.4 $\pm$ 0.5 & 4.19 $\pm$ 0.26 & \textbf{33.4} $\pm$ 1.5 & \textbf{6.20} $\pm$ 0.34 & \textbf{24.3} $\pm$ 1.3 \\
Radial profile & 2.173 $\pm$ 0.068 & 51.1 $\pm$ 1.0 & 82.6 $\pm$ 2.4 & 12.4 $\pm$ 0.7 & 4.24 $\pm$ 0.13 & 33.0 $\pm$ 1.2 & 6.27 $\pm$ 0.24 & 23.7 $\pm$ 1.3 \\
Direct regression & 2.200 $\pm$ 0.090 & 50.9 $\pm$ 1.5 & 84.3 $\pm$ 3.3 & \textbf{12.7} $\pm$ 0.5 & 4.27 $\pm$ 0.15 & 32.9 $\pm$ 1.2 & 6.36 $\pm$ 0.27 & 24.0 $\pm$ 1.0 \\
\bottomrule
\end{tabular}}
\label{tab:ablation_distorted}
\end{table*}

%% file: tab/extended_pinhole.tex
\begin{table*}[t]
\caption{\textbf{Extended results on pinhole benchmarks.} Median error and mAA per benchmark, as mean and standard deviation over seeds where there are several. Fit: the camera model the intrinsics are fitted in (Tab.~\ref{tab:cams}); GeoCalib uses its pinhole weights. Bold: best mean. $^{*}$: our GPU implementation of AnyCalib's fit. ViT-S / B / L: the network's backbone. Foundation models and UniK3D are left out of benchmarks whose dataset is in their training data.}
\centering
\resizebox{\textwidth}{!}{%
\begin{tabular}{l l cc cc cc cc}
\toprule
 & & \multicolumn{2}{c}{AE ($^\circ$)} & \multicolumn{2}{c}{RE (px)} & \multicolumn{2}{c}{vFoV ($^\circ$)} & \multicolumn{2}{c}{hFoV ($^\circ$)} \\
\cmidrule(lr){3-4} \cmidrule(lr){5-6} \cmidrule(lr){7-8} \cmidrule(lr){9-10}
Method & Fit & Median $\downarrow$ & mAA ($5^\circ$) $\uparrow$ & Median $\downarrow$ & mAA (50\,px) $\uparrow$ & Median $\downarrow$ & mAA ($5^\circ$) $\uparrow$ & Median $\downarrow$ & mAA ($5^\circ$) $\uparrow$ \\
\midrule
\multicolumn{10}{l}{\textit{Stanford2D3D}~\cite{armeni2017joint}} \\
\midrule
GeoCalib$_\text{pinhole}$ & Pinhole & 1.436 & 65.2 & 15.2 & 60.0 & 3.21 & 39.9 & 3.56 & 37.2 \\
MoGe~\cite{wang2025moge} & Pinhole & 1.921 & 54.6 & 21.6 & 49.4 & 4.34 & 33.7 & 4.98 & 30.3 \\
MoGe-2~\cite{wang2025moge2} & Pinhole & 1.838 & 58.9 & 18.7 & 55.8 & 4.15 & 31.7 & 4.64 & 29.3 \\
MoGe-3~\cite{kong2026moge3} & Pinhole & 2.528 & 48.2 & 23.9 & 47.5 & 5.67 & 19.9 & 6.12 & 18.4 \\
SurGe~\cite{knaebel2026surge} & Pinhole & 1.796 & 57.4 & 20.3 & 51.9 & 4.06 & 34.5 & 4.68 & 30.9 \\
UniDepth~\cite{piccinelli2024unidepth} & Pinhole & 6.555 & 19.1 & 73.6 & 16.2 & 14.66 & 8.5 & 16.72 & 7.7 \\
UniDepthV2~\cite{piccinelli2025unidepthv2} & Pinhole & 2.937 & 41.8 & 30.9 & 38.5 & 6.59 & 20.8 & 7.11 & 19.2 \\
Depth Anything 3~\cite{lin2025da3} & Pinhole & 2.018 & 52.3 & 23.0 & 46.7 & 4.47 & 34.5 & 5.35 & 31.2 \\
MapAnything~\cite{keetha2026mapanything} & Pinhole & 2.324 & 49.5 & 24.4 & 45.5 & 5.04 & 27.2 & 5.51 & 25.0 \\
VGGT-$\Omega$~\cite{wang2026vggtomega} & Pinhole & 1.723 & 56.9 & 19.2 & 51.0 & 3.87 & 36.9 & 4.50 & 33.2 \\
UniK3D$^{*}$ & Pinhole & 1.788 & 59.8 & 18.5 & 54.5 & 3.86 & 35.3 & 4.48 & 31.0 \\
AnyCalib$_\text{pinhole}$ & Pinhole & 1.129 & 70.5 & 12.1 & 64.4 & 2.55 & 46.9 & 2.88 & 43.3 \\
AnyCalib$_\text{gen}$ & Pinhole & 1.297 & 67.6 & 13.9 & 61.6 & 2.95 & 43.6 & 3.24 & 40.4 \\
AnyCalib$_\text{ours}^{*}$ ViT-S & Pinhole & 1.720 $\pm$ 0.039 & 59.6 $\pm$ 0.2 & 18.6 $\pm$ 0.7 & 54.3 $\pm$ 0.4 & 3.86 $\pm$ 0.08 & 35.0 $\pm$ 0.7 & 4.37 $\pm$ 0.13 & 31.8 $\pm$ 0.7 \\
AnyCalib$_\text{ours}^{*}$ ViT-L & Pinhole & \textbf{0.957} $\pm$ 0.025 & \textbf{75.3} $\pm$ 0.6 & \textbf{10.1} $\pm$ 0.2 & \textbf{69.0} $\pm$ 0.7 & \textbf{2.14} $\pm$ 0.05 & \textbf{52.4} $\pm$ 0.6 & \textbf{2.40} $\pm$ 0.04 & \textbf{48.9} $\pm$ 0.5 \\
Ours ViT-S & Pinhole & 1.719 & 59.5 & 18.6 & 54.4 & 3.85 & 35.2 & 4.28 & 31.9 \\
Ours ViT-B & Pinhole & 1.337 & 65.6 & 14.4 & 60.0 & 3.01 & 41.7 & 3.40 & 38.4 \\
Ours ViT-L & Pinhole & 0.997 $\pm$ 0.009 & 72.8 $\pm$ 0.3 & 10.9 $\pm$ 0.1 & 66.5 $\pm$ 0.4 & 2.24 $\pm$ 0.02 & 50.5 $\pm$ 0.2 & 2.55 $\pm$ 0.04 & 47.1 $\pm$ 0.2 \\
\midrule
\multicolumn{10}{l}{\textit{TartanAir}~\cite{wang2020tartanair}} \\
\midrule
GeoCalib$_\text{pinhole}$ & Pinhole & 2.182 & 53.1 & 19.2 & 53.3 & 4.92 & 30.5 & 5.12 & 29.6 \\
UniDepth~\cite{piccinelli2024unidepth} & Pinhole & 13.993 & 3.1 & 204.1 & 3.6 & 32.15 & 1.6 & 35.10 & 1.5 \\
AnyCalib$_\text{pinhole}$ & Pinhole & 1.645 & 61.1 & 15.0 & 59.2 & 3.62 & 36.4 & 3.93 & 34.7 \\
AnyCalib$_\text{gen}$ & Pinhole & 1.764 & 59.3 & 16.0 & 58.2 & 4.06 & 33.5 & 4.09 & 33.4 \\
AnyCalib$_\text{ours}^{*}$ ViT-S & Pinhole & 1.959 $\pm$ 0.076 & 54.8 $\pm$ 1.2 & 17.5 $\pm$ 0.7 & 54.3 $\pm$ 1.2 & 4.42 $\pm$ 0.18 & 31.6 $\pm$ 1.2 & 4.60 $\pm$ 0.16 & 30.7 $\pm$ 1.2 \\
AnyCalib$_\text{ours}^{*}$ ViT-L & Pinhole & \textbf{1.378} $\pm$ 0.090 & 64.1 $\pm$ 1.7 & \textbf{12.2} $\pm$ 0.9 & 62.1 $\pm$ 1.6 & \textbf{3.11} $\pm$ 0.20 & \textbf{41.7} $\pm$ 1.3 & \textbf{3.22} $\pm$ 0.22 & \textbf{40.6} $\pm$ 1.3 \\
Ours ViT-S & Pinhole & 1.793 & 59.0 & 15.9 & 58.2 & 4.04 & 35.4 & 4.21 & 34.4 \\
Ours ViT-B & Pinhole & 1.728 & 58.6 & 15.5 & 57.5 & 3.90 & 35.5 & 4.06 & 34.4 \\
Ours ViT-L & Pinhole & 1.383 $\pm$ 0.066 & \textbf{64.2} $\pm$ 0.9 & 12.4 $\pm$ 0.6 & \textbf{62.2} $\pm$ 0.8 & 3.12 $\pm$ 0.15 & 41.5 $\pm$ 1.0 & 3.26 $\pm$ 0.16 & 40.3 $\pm$ 1.0 \\
\midrule
\multicolumn{10}{l}{\textit{MegaDepth}~\cite{li2018megadepth}} \\
\midrule
GeoCalib$_\text{pinhole}$ & Pinhole & 1.893 & 54.9 & 58.2 & 25.4 & 4.51 & 31.6 & 4.91 & 30.0 \\
UniDepth~\cite{piccinelli2024unidepth} & Pinhole & 4.515 & 30.4 & 128.1 & 9.7 & 10.82 & 16.2 & 11.10 & 14.9 \\
AnyCalib$_\text{pinhole}$ & Pinhole & 1.306 & 67.0 & 39.5 & 33.2 & 3.14 & 40.8 & 3.36 & 39.3 \\
AnyCalib$_\text{gen}$ & Pinhole & 1.489 & 63.4 & 47.1 & 28.5 & 3.58 & 36.6 & 3.84 & 34.5 \\
AnyCalib$_\text{ours}^{*}$ ViT-S & Pinhole & 1.012 $\pm$ 0.021 & 76.3 $\pm$ 0.2 & 34.0 $\pm$ 0.9 & 36.8 $\pm$ 0.6 & 2.40 $\pm$ 0.06 & 49.7 $\pm$ 0.5 & 2.59 $\pm$ 0.07 & 47.1 $\pm$ 0.6 \\
AnyCalib$_\text{ours}^{*}$ ViT-L & Pinhole & \textbf{0.632} $\pm$ 0.023 & \textbf{86.9} $\pm$ 0.5 & \textbf{22.0} $\pm$ 0.7 & \textbf{49.3} $\pm$ 0.8 & \textbf{1.52} $\pm$ 0.06 & \textbf{64.7} $\pm$ 1.1 & \textbf{1.62} $\pm$ 0.04 & \textbf{62.1} $\pm$ 0.8 \\
Ours ViT-S & Pinhole & 0.951 & 77.9 & 32.2 & 38.3 & 2.26 & 51.1 & 2.45 & 48.7 \\
Ours ViT-B & Pinhole & 0.830 & 81.3 & 27.9 & 42.0 & 1.97 & 56.2 & 2.15 & 53.5 \\
Ours ViT-L & Pinhole & 0.640 $\pm$ 0.016 & 86.8 $\pm$ 0.5 & 22.2 $\pm$ 0.5 & \textbf{49.3} $\pm$ 0.8 & 1.53 $\pm$ 0.06 & \textbf{64.7} $\pm$ 0.9 & 1.64 $\pm$ 0.03 & 61.8 $\pm$ 0.9 \\
\midrule
\multicolumn{10}{l}{\textit{LaMAR}~\cite{sarlin2022lamar}} \\
\midrule
GeoCalib$_\text{pinhole}$ & Pinhole & 1.148 & 70.5 & 39.6 & 32.6 & 3.04 & 41.3 & 2.61 & 45.8 \\
MoGe~\cite{wang2025moge} & Pinhole & 0.740 & 84.1 & 25.6 & 44.6 & 1.97 & 55.3 & 1.69 & 60.6 \\
MoGe-2~\cite{wang2025moge2} & Pinhole & 0.873 & 82.8 & 29.1 & 41.2 & 2.31 & 51.2 & 1.98 & 56.9 \\
MoGe-3~\cite{kong2026moge3} & Pinhole & 0.979 & 81.7 & 32.3 & 38.2 & 2.59 & 48.2 & 2.23 & 54.1 \\
SurGe~\cite{knaebel2026surge} & Pinhole & 0.618 & 86.6 & 21.1 & 50.2 & 1.64 & 60.9 & 1.40 & 65.8 \\
UniDepth~\cite{piccinelli2024unidepth} & Pinhole & 1.142 & 59.1 & 37.0 & 26.4 & 2.47 & 37.1 & 2.89 & 32.4 \\
UniDepthV2~\cite{piccinelli2025unidepthv2} & Pinhole & 0.916 & 83.0 & 30.3 & 38.7 & 2.30 & 52.2 & 2.04 & 57.1 \\
Depth Anything 3~\cite{lin2025da3} & Pinhole & \textbf{0.259} & \textbf{92.3} & \textbf{8.8} & \textbf{69.0} & \textbf{0.68} & \textbf{79.1} & \textbf{0.60} & \textbf{82.0} \\
MapAnything~\cite{keetha2026mapanything} & Pinhole & 1.401 & 74.7 & 44.8 & 20.1 & 3.17 & 40.6 & 2.72 & 48.6 \\
VGGT-$\Omega$~\cite{wang2026vggtomega} & Pinhole & 0.893 & 81.8 & 29.9 & 39.6 & 2.22 & 51.4 & 2.25 & 52.5 \\
UniK3D$^{*}$ & Pinhole & 2.776 & 47.4 & 84.5 & 7.9 & 7.22 & 11.5 & 6.38 & 13.1 \\
AnyCalib$_\text{pinhole}$ & Pinhole & 0.848 & 80.8 & 28.6 & 41.5 & 2.26 & 51.6 & 1.92 & 57.1 \\
AnyCalib$_\text{gen}$ & Pinhole & 1.082 & 75.9 & 35.9 & 34.6 & 2.81 & 44.1 & 2.49 & 48.3 \\
AnyCalib$_\text{ours}^{*}$ ViT-S & Pinhole & 1.142 $\pm$ 0.024 & 72.7 $\pm$ 0.5 & 38.7 $\pm$ 0.8 & 33.0 $\pm$ 0.7 & 3.01 $\pm$ 0.06 & 42.1 $\pm$ 0.7 & 2.60 $\pm$ 0.05 & 46.9 $\pm$ 0.7 \\
AnyCalib$_\text{ours}^{*}$ ViT-L & Pinhole & 0.701 $\pm$ 0.023 & 84.6 $\pm$ 0.3 & 23.9 $\pm$ 0.8 & 46.9 $\pm$ 0.8 & 1.85 $\pm$ 0.06 & 57.7 $\pm$ 0.7 & 1.60 $\pm$ 0.06 & 62.5 $\pm$ 0.8 \\
Ours ViT-S & Pinhole & 1.101 & 73.6 & 37.4 & 34.4 & 2.92 & 43.4 & 2.49 & 48.1 \\
Ours ViT-B & Pinhole & 0.897 & 78.6 & 30.5 & 39.9 & 2.38 & 49.5 & 2.03 & 54.4 \\
Ours ViT-L & Pinhole & 0.705 $\pm$ 0.034 & 84.4 $\pm$ 0.4 & 23.9 $\pm$ 1.0 & 47.3 $\pm$ 1.1 & 1.87 $\pm$ 0.09 & 57.5 $\pm$ 1.0 & 1.61 $\pm$ 0.08 & 62.4 $\pm$ 0.9 \\
\bottomrule
\end{tabular}}
\label{tab:extended_pinhole}
\end{table*}

%% file: tab/extended_distorted.tex
\begin{table*}[t]
\caption{\textbf{Extended results on distorted benchmarks.} Median error and mAA per benchmark, as mean and standard deviation over seeds where there are several. Fit: the camera model the intrinsics are fitted in (Tab.~\ref{tab:cams}); GeoCalib uses its distorted weights. Bold: best mean. $^{*}$: our GPU implementation of AnyCalib's fit. ViT-S / B / L: the network's backbone. UniK3D is left out of benchmarks whose dataset is in its training data.}
\centering
\resizebox{\textwidth}{!}{%
\begin{tabular}{l l cc cc cc cc}
\toprule
 & & \multicolumn{2}{c}{AE ($^\circ$)} & \multicolumn{2}{c}{RE (px)} & \multicolumn{2}{c}{vFoV ($^\circ$)} & \multicolumn{2}{c}{hFoV ($^\circ$)} \\
\cmidrule(lr){3-4} \cmidrule(lr){5-6} \cmidrule(lr){7-8} \cmidrule(lr){9-10}
Method & Fit & Median $\downarrow$ & mAA ($5^\circ$) $\uparrow$ & Median $\downarrow$ & mAA (50\,px) $\uparrow$ & Median $\downarrow$ & mAA ($5^\circ$) $\uparrow$ & Median $\downarrow$ & mAA ($5^\circ$) $\uparrow$ \\
\midrule
\multicolumn{10}{l}{\textit{MegaDepth-radial}~\cite{li2018megadepth}} \\
\midrule
GeoCalib$_\text{dist}$ & BC-1 & 1.924 & 55.0 & 56.4 & 26.4 & 4.56 & 31.8 & 5.06 & 30.5 \\
AnyCalib$_\text{dist}$ & BC-1 & 1.621 & 62.0 & 46.8 & 28.5 & 3.78 & 35.6 & 4.02 & 33.3 \\
AnyCalib$_\text{gen}$ & BC-1 & 1.460 & 63.9 & 44.6 & 29.1 & 3.55 & 37.1 & 3.79 & 34.6 \\
AnyCalib$_\text{ours}^{*}$ ViT-S & BC-1 & 0.999 $\pm$ 0.029 & 76.2 $\pm$ 0.6 & 33.3 $\pm$ 1.6 & 37.2 $\pm$ 1.1 & 2.42 $\pm$ 0.07 & 49.1 $\pm$ 1.1 & 2.62 $\pm$ 0.07 & 46.8 $\pm$ 1.1 \\
AnyCalib$_\text{ours}^{*}$ ViT-L & BC-1 & 0.590 $\pm$ 0.009 & \textbf{87.4} $\pm$ 0.4 & \textbf{19.7} $\pm$ 0.2 & \textbf{51.2} $\pm$ 0.5 & \textbf{1.42} $\pm$ 0.03 & \textbf{65.8} $\pm$ 0.6 & 1.54 $\pm$ 0.02 & 63.1 $\pm$ 0.6 \\
Ours ViT-S & BC-1 & 0.929 & 77.7 & 31.7 & 38.9 & 2.22 & 51.0 & 2.47 & 48.5 \\
Ours ViT-B & BC-1 & 0.803 & 81.7 & 26.7 & 43.3 & 1.92 & 56.3 & 2.04 & 54.3 \\
Ours ViT-L & BC-1 & \textbf{0.586} $\pm$ 0.006 & 87.3 $\pm$ 0.3 & 19.8 $\pm$ 0.1 & \textbf{51.2} $\pm$ 0.4 & 1.43 $\pm$ 0.02 & \textbf{65.8} $\pm$ 0.5 & \textbf{1.52} $\pm$ 0.02 & \textbf{63.3} $\pm$ 0.6 \\
\midrule
\multicolumn{10}{l}{\textit{ScanNet++}~\cite{yeshwanth2023scannetpp}} \\
\midrule
GeoCalib$_\text{dist}$ & Division & 4.985 & 23.7 & 71.6 & 16.0 & 10.27 & 11.8 & 14.24 & 9.8 \\
AnyCalib$_\text{dist}$ & KB-4 & 1.571 & 67.3 & 20.6 & 53.1 & 3.91 & 31.8 & 3.38 & 39.0 \\
AnyCalib$_\text{gen}$ & KB-4 & 1.884 & 60.8 & 24.8 & 46.1 & 4.27 & 29.5 & 5.04 & 23.7 \\
AnyCalib$_\text{ours}^{*}$ ViT-S & KB-4 & 1.814 $\pm$ 0.094 & 60.2 $\pm$ 2.1 & 23.2 $\pm$ 1.1 & 47.4 $\pm$ 1.9 & 4.02 $\pm$ 0.23 & 33.2 $\pm$ 1.7 & 4.33 $\pm$ 0.24 & 31.1 $\pm$ 1.5 \\
AnyCalib$_\text{ours}^{*}$ ViT-L & KB-4 & \textbf{1.002} $\pm$ 0.020 & \textbf{81.7} $\pm$ 0.5 & \textbf{12.7} $\pm$ 0.2 & \textbf{68.2} $\pm$ 0.7 & 1.97 $\pm$ 0.07 & \textbf{57.5} $\pm$ 1.4 & \textbf{2.28} $\pm$ 0.06 & \textbf{52.3} $\pm$ 1.0 \\
Ours ViT-S & KB-4 & 2.038 & 55.7 & 25.9 & 43.9 & 4.14 & 31.9 & 4.33 & 30.6 \\
Ours ViT-B & KB-4 & 1.520 & 66.2 & 19.5 & 53.4 & 3.14 & 39.9 & 3.51 & 35.8 \\
Ours ViT-L & KB-4 & 1.010 $\pm$ 0.015 & 80.2 $\pm$ 1.6 & 12.8 $\pm$ 0.1 & 67.0 $\pm$ 1.1 & \textbf{1.95} $\pm$ 0.07 & 56.9 $\pm$ 1.1 & 2.30 $\pm$ 0.05 & 51.8 $\pm$ 0.6 \\
\midrule
\multicolumn{10}{l}{\textit{Mono}~\cite{engel2016photometrically}} \\
\midrule
GeoCalib$_\text{dist}$ & Division & 3.932 & 33.0 & 56.6 & 22.5 & 9.04 & 16.3 & 10.63 & 14.0 \\
UniK3D$^{*}$ & UCM & 6.963 & 14.0 & 161.4 & 2.1 & 15.76 & 5.9 & 23.12 & 2.0 \\
AnyCalib$_\text{dist}$ & UCM & 1.606 & 62.4 & 21.7 & 47.9 & 3.60 & 37.4 & 3.94 & 34.2 \\
AnyCalib$_\text{gen}$ & UCM & 1.641 & 61.3 & 22.5 & 46.3 & 3.66 & 36.8 & 3.95 & 34.3 \\
AnyCalib$_\text{ours}^{*}$ ViT-S & UCM & 2.008 $\pm$ 0.042 & 54.4 $\pm$ 0.7 & 26.9 $\pm$ 0.7 & 41.0 $\pm$ 0.7 & 4.58 $\pm$ 0.11 & 29.9 $\pm$ 0.7 & 4.96 $\pm$ 0.17 & 28.2 $\pm$ 0.6 \\
AnyCalib$_\text{ours}^{*}$ ViT-L & UCM & 1.481 $\pm$ 0.044 & 65.2 $\pm$ 1.1 & 19.5 $\pm$ 0.6 & 50.9 $\pm$ 1.1 & 3.26 $\pm$ 0.12 & 40.1 $\pm$ 1.0 & 3.48 $\pm$ 0.10 & 38.0 $\pm$ 0.9 \\
Ours ViT-S & UCM & 1.992 & 55.7 & 26.4 & 42.1 & 4.54 & 31.3 & 5.03 & 28.4 \\
Ours ViT-B & UCM & 1.595 & 62.2 & 21.1 & 48.0 & 3.56 & 36.8 & 3.86 & 34.5 \\
Ours ViT-L & UCM & \textbf{1.358} $\pm$ 0.028 & \textbf{68.2} $\pm$ 0.9 & \textbf{18.1} $\pm$ 0.4 & \textbf{53.4} $\pm$ 0.6 & \textbf{2.94} $\pm$ 0.07 & \textbf{42.8} $\pm$ 1.1 & \textbf{3.21} $\pm$ 0.10 & \textbf{40.1} $\pm$ 1.1 \\
\midrule
\multicolumn{10}{l}{\textit{Cathedral}~\cite{kocur2026minimal}} \\
\midrule
GeoCalib$_\text{dist}$ & Division & 1.054 & 76.7 & 11.7 & 70.1 & 1.82 & 59.3 & 2.13 & 51.7 \\
UniK3D$^{*}$ & Division & 2.165 & 53.8 & 27.1 & 39.8 & 2.96 & 44.4 & 5.47 & 22.4 \\
AnyCalib$_\text{dist}$ & Division & 2.019 & 60.2 & 22.9 & 51.9 & 4.30 & 30.1 & 4.51 & 28.8 \\
AnyCalib$_\text{gen}$ & Division & 1.853 & 64.2 & 20.7 & 56.3 & 3.85 & 33.4 & 4.53 & 29.8 \\
AnyCalib$_\text{ours}^{*}$ ViT-S & Division & 1.109 $\pm$ 0.030 & 77.3 $\pm$ 0.6 & 12.0 $\pm$ 0.6 & 70.0 $\pm$ 0.6 & 1.79 $\pm$ 0.07 & 59.9 $\pm$ 1.3 & 2.39 $\pm$ 0.08 & 49.2 $\pm$ 1.6 \\
AnyCalib$_\text{ours}^{*}$ ViT-L & Division & 0.877 $\pm$ 0.078 & 82.3 $\pm$ 1.1 & 9.8 $\pm$ 0.8 & 75.3 $\pm$ 0.8 & 1.43 $\pm$ 0.14 & 68.2 $\pm$ 2.6 & 1.57 $\pm$ 0.23 & 62.4 $\pm$ 3.6 \\
Ours ViT-S & Division & 1.009 & 80.2 & 11.4 & 72.9 & 1.48 & 65.5 & 2.13 & 52.9 \\
Ours ViT-B & Division & 0.937 & 81.2 & 10.7 & 73.4 & 1.39 & 68.3 & 1.79 & 57.3 \\
Ours ViT-L & Division & \textbf{0.812} $\pm$ 0.053 & \textbf{83.7} $\pm$ 0.8 & \textbf{9.3} $\pm$ 0.5 & \textbf{76.5} $\pm$ 0.5 & \textbf{1.15} $\pm$ 0.10 & \textbf{73.7} $\pm$ 1.6 & \textbf{1.51} $\pm$ 0.10 & \textbf{62.8} $\pm$ 2.5 \\
\midrule
\multicolumn{10}{l}{\textit{InFlux}~\cite{liang2025influx}} \\
\midrule
GeoCalib$_\text{dist}$ & BC-1 & 6.803 & 15.7 & 276.0 & 3.8 & 13.97 & 7.0 & 20.54 & 5.1 \\
UniK3D$^{*}$ & BC-2 & 3.641 & 34.1 & 183.5 & 4.3 & 7.30 & 15.6 & 11.17 & 9.7 \\
AnyCalib$_\text{dist}$ & BC-2 & 4.062 & 29.6 & 209.9 & 5.6 & 8.23 & 12.9 & 12.54 & 8.0 \\
AnyCalib$_\text{gen}$ & BC-2 & 3.886 & 32.7 & 205.7 & 7.2 & 7.80 & 15.5 & 12.11 & 10.1 \\
AnyCalib$_\text{ours}^{*}$ ViT-S & BC-2 & 3.416 $\pm$ 0.137 & 38.9 $\pm$ 0.8 & 180.1 $\pm$ 2.9 & 8.1 $\pm$ 0.3 & 6.96 $\pm$ 0.25 & 21.6 $\pm$ 0.3 & 10.53 $\pm$ 0.42 & 14.6 $\pm$ 0.5 \\
AnyCalib$_\text{ours}^{*}$ ViT-L & BC-2 & 2.061 $\pm$ 0.196 & \textbf{58.3} $\pm$ 3.2 & 117.7 $\pm$ 10.2 & 11.4 $\pm$ 1.7 & 4.17 $\pm$ 0.43 & 32.0 $\pm$ 3.8 & 6.38 $\pm$ 0.55 & \textbf{21.3} $\pm$ 2.8 \\
Ours ViT-S & BC-2 & 3.090 & 41.9 & 168.5 & 8.4 & 6.28 & 22.8 & 9.49 & 15.3 \\
Ours ViT-B & BC-2 & 2.749 & 46.2 & 150.1 & 9.5 & 5.57 & 25.1 & 8.52 & 17.9 \\
Ours ViT-L & BC-2 & \textbf{1.981} $\pm$ 0.106 & 57.9 $\pm$ 1.4 & \textbf{117.3} $\pm$ 3.8 & \textbf{11.5} $\pm$ 0.6 & \textbf{3.99} $\pm$ 0.22 & \textbf{32.5} $\pm$ 1.3 & \textbf{6.17} $\pm$ 0.32 & 21.2 $\pm$ 0.8 \\
\midrule
\multicolumn{10}{l}{\textit{InFlux++}~\cite{liang2026influxpp}} \\
\midrule
GeoCalib$_\text{dist}$ & BC-1 & 4.921 & 26.7 & 176.0 & 6.2 & 10.04 & 15.1 & 14.59 & 11.2 \\
UniK3D$^{*}$ & BC-2 & 5.427 & 21.6 & 185.7 & 3.0 & 10.97 & 11.8 & 15.88 & 8.6 \\
AnyCalib$_\text{dist}$ & BC-2 & 3.493 & 34.2 & 125.2 & 8.5 & 6.91 & 20.5 & 10.32 & 13.9 \\
AnyCalib$_\text{gen}$ & BC-2 & 3.396 & 35.5 & 121.1 & 9.3 & 6.70 & 22.0 & 10.00 & 15.5 \\
AnyCalib$_\text{ours}^{*}$ ViT-S & BC-2 & 3.149 $\pm$ 0.075 & 38.9 $\pm$ 0.7 & 126.9 $\pm$ 3.5 & 7.6 $\pm$ 0.3 & 6.33 $\pm$ 0.15 & 22.3 $\pm$ 0.6 & 9.29 $\pm$ 0.20 & 15.4 $\pm$ 0.6 \\
AnyCalib$_\text{ours}^{*}$ ViT-L & BC-2 & \textbf{2.074} $\pm$ 0.050 & \textbf{52.9} $\pm$ 0.6 & \textbf{80.2} $\pm$ 2.1 & 12.4 $\pm$ 0.4 & \textbf{4.05} $\pm$ 0.12 & \textbf{34.1} $\pm$ 0.9 & \textbf{5.93} $\pm$ 0.20 & \textbf{25.3} $\pm$ 1.1 \\
Ours ViT-S & BC-2 & 2.901 & 41.1 & 120.7 & 7.9 & 5.79 & 23.8 & 8.71 & 16.0 \\
Ours ViT-B & BC-2 & 2.625 & 44.5 & 111.5 & 9.8 & 5.20 & 27.4 & 7.67 & 19.4 \\
Ours ViT-L & BC-2 & 2.176 $\pm$ 0.057 & 51.2 $\pm$ 0.5 & 84.8 $\pm$ 1.3 & \textbf{12.5} $\pm$ 0.5 & 4.21 $\pm$ 0.12 & 33.4 $\pm$ 0.8 & 6.30 $\pm$ 0.15 & 24.1 $\pm$ 0.8 \\
\bottomrule
\end{tabular}}
\label{tab:extended_distorted}
\end{table*}

%% file: fig/pose_equal.tex
% generated by rdnet/plot_calib_net_tikz.py from calib_net_cathedral_eq_ray.csv
\begin{tikzpicture}
\definecolor{cOursL}{HTML}{1F77B4}
\definecolor{cOursC}{HTML}{FF7F0E}
\definecolor{cAny}{HTML}{E377C2}
\definecolor{cAnyO}{HTML}{17BECF}
\definecolor{cGeo}{HTML}{7F7F7F}
\definecolor{cUni}{HTML}{BCBD22}
\definecolor{cSix}{HTML}{2CA02C}
\definecolor{cSeven}{HTML}{D62728}
\definecolor{cNine}{HTML}{9467BD}
\definecolor{cTen}{HTML}{8C564B}
\begin{axis}[
  scale only axis, width=0.85\linewidth, height=0.52\linewidth,
  xmode=log, log basis x=10,
  xmin=28.3, xmax=512.8, ymin=45, ymax=85,
  xtick={30,50,100,200,300,500}, xticklabels={30,50,100,200,300,500}, minor xtick={},
  ytick={45,55,65,75,85}, yticklabels={45,55,65,75,85},
  xlabel={Mean runtime per pair (ms)}, ylabel={Pose mAA at $10^\circ$ (\%)},
  label style={font=\small}, tick label style={font=\footnotesize},
  xlabel style={yshift=2pt}, ylabel style={yshift=-3pt},
  axis lines=left, axis line style={-, line width=0.4pt}, tick style={black, line width=0.4pt}, tick align=outside,
  major tick length=2pt, grid=major, grid style={black!12, line width=0.3pt},
  clip mode=individual, name=ax,
]
\addplot[cNine, line width=0.8pt, mark=*, mark size=0.8pt] coordinates {(44.1,48.75) (53.6,56.22) (68.2,62.46) (83.2,65.38) (105.3,67.45) (158.3,69.26) (236.9,70.40)};
\addplot[cSix, line width=0.8pt, mark=star, mark size=0.8pt] coordinates {(52.1,71.22) (62.7,73.22) (83.1,74.85) (108.2,75.60) (150.8,76.08) (264.5,76.56) (443.5,76.90)};
\addplot[cUni, solid, line width=0.8pt, mark=triangle*, mark size=0.6pt, mark options={solid}] coordinates {(157.0,70.77) (166.3,72.09) (181.7,73.09) (198.0,73.57) (222.3,73.84) (280.1,74.06) (364.3,74.11)};
\addplot[cUni, dashed, line width=0.8pt, mark=triangle*, mark size=0.6pt, mark options={solid}] coordinates {(124.0,61.09) (125.4,61.98) (127.9,62.93) (131.3,63.17) (137.3,63.19) (153.9,63.40) (180.6,63.41)};
\addplot[cGeo, solid, line width=0.8pt, mark=pentagon*, mark size=0.6pt, mark options={solid}] coordinates {(264.9,76.09) (273.8,77.27) (288.9,78.61) (305.1,79.02) (329.7,79.37) (388.2,79.65) (474.8,79.83)};
\addplot[cGeo, dashed, line width=0.8pt, mark=pentagon*, mark size=0.6pt, mark options={solid}] coordinates {(235.5,77.92) (236.6,79.09) (239.1,80.10) (242.5,80.45) (248.7,80.76) (266.2,81.08) (294.7,81.22)};
\addplot[cAny, solid, line width=0.8pt, mark=diamond*, mark size=0.6pt, mark options={solid}] coordinates {(104.0,75.70) (113.2,77.16) (129.0,78.36) (145.6,78.92) (170.8,79.18) (231.2,79.44) (320.1,79.51)};
\addplot[cAny, dashed, line width=0.8pt, mark=diamond*, mark size=0.6pt, mark options={solid}] coordinates {(73.8,77.22) (74.9,78.46) (77.4,79.36) (80.9,79.84) (87.2,80.22) (104.8,80.47) (133.6,80.62)};
\addplot[cAnyO, solid, line width=0.8pt, mark=diamond, mark size=0.6pt, mark options={solid}] coordinates {(76.6,76.36) (85.4,77.83) (100.5,78.93) (116.9,79.43) (141.5,79.81) (200.7,80.12) (288.5,80.20)};
\addplot[cAnyO, dashed, line width=0.8pt, mark=diamond, mark size=0.6pt, mark options={solid}] coordinates {(46.8,79.19) (47.9,80.42) (50.3,81.31) (53.8,81.88) (59.9,82.21) (77.3,82.51) (105.7,82.65)};
\addplot[cOursC, solid, line width=0.8pt, mark=square*, mark size=0.6pt, mark options={solid}] coordinates {(60.7,76.31) (69.4,77.72) (84.2,78.88) (100.2,79.44) (124.7,79.79) (183.1,80.05) (270.3,80.15)};
\addplot[cOursC, dashed, line width=0.8pt, mark=square*, mark size=0.6pt, mark options={solid}] coordinates {(30.7,79.21) (31.8,80.37) (34.2,81.36) (37.6,81.94) (43.7,82.40) (61.1,82.62) (89.3,82.73)};
\addplot[cOursL, solid, line width=0.8pt, mark=*, mark size=0.6pt, mark options={solid}] coordinates {(67.8,76.34) (76.5,77.74) (91.6,78.90) (107.7,79.42) (132.2,79.79) (191.0,79.99) (278.3,80.16)};
\addplot[cOursL, dashed, line width=0.8pt, mark=*, mark size=0.6pt, mark options={solid}] coordinates {(37.8,79.49) (38.8,80.62) (41.3,81.69) (44.6,82.22) (50.8,82.53) (68.2,82.80) (96.5,82.92)};
\end{axis}
\node[anchor=south, inner sep=0pt, yshift=3pt] at (ax.outer north) {\footnotesize\begin{tabular}{@{}c@{}}
 \mbox{\tikz[baseline=-0.6ex]\draw[cOursL, line width=0.8pt, mark=*, mark size=0.8pt, mark indices={2}] plot coordinates {(0,0) (0.2,0) (0.4,0)};\,Ours ViT-L}\hspace{6pt}\mbox{\tikz[baseline=-0.6ex]\draw[cOursC, line width=0.8pt, mark=square*, mark size=0.8pt, mark indices={2}] plot coordinates {(0,0) (0.2,0) (0.4,0)};\,Ours ViT-B}\hspace{6pt}\mbox{\tikz[baseline=-0.6ex]\draw[cAnyO, line width=0.8pt, mark=diamond, mark size=0.8pt, mark indices={2}] plot coordinates {(0,0) (0.2,0) (0.4,0)};\,AnyCalib$_\text{ours}$} \\
 \mbox{\tikz[baseline=-0.6ex]\draw[cAny, line width=0.8pt, mark=diamond*, mark size=0.8pt, mark indices={2}] plot coordinates {(0,0) (0.2,0) (0.4,0)};\,AnyCalib$_\text{dist}$~\cite{tirado2025anycalib}}\hspace{6pt}\mbox{\tikz[baseline=-0.6ex]\draw[cGeo, line width=0.8pt, mark=pentagon*, mark size=0.8pt, mark indices={2}] plot coordinates {(0,0) (0.2,0) (0.4,0)};\,GeoCalib$_\text{dist}$~\cite{veicht2024geocalib}}\hspace{6pt}\mbox{\tikz[baseline=-0.6ex]\draw[cUni, line width=0.8pt, mark=triangle*, mark size=0.8pt, mark indices={2}] plot coordinates {(0,0) (0.2,0) (0.4,0)};\,UniK3D~\cite{piccinelli2025unik3d}} \\
 \mbox{\tikz[baseline=-0.6ex]\draw[cSix, line width=0.8pt, mark=star, mark size=0.8pt, mark indices={2}] plot coordinates {(0,0) (0.2,0) (0.4,0)};\,6pt + sampling~\cite{kocur2026minimal}}\hspace{6pt}\mbox{\tikz[baseline=-0.6ex]\draw[cNine, line width=0.8pt, mark=*, mark size=0.8pt, mark indices={2}] plot coordinates {(0,0) (0.2,0) (0.4,0)};\,9pt~\cite{fitzgibbon2001division,kocur2026minimal}}
\end{tabular}};
\end{tikzpicture}